\documentclass[preprint,12pt]{elsarticle}

\usepackage{graphicx} 
\usepackage{booktabs} 
\usepackage{amssymb} 
\usepackage{amsmath} 
\usepackage{multirow} 
\usepackage{colortbl} 

\usepackage{microtype} 
\usepackage{url} 
\usepackage{xcolor} 
\usepackage{wasysym} 

\usepackage{hyperref} 
\hypersetup{
 colorlinks = true, 
 linkcolor = linkcolor, 
 filecolor = magenta, 
 citecolor = dark-green,
 urlcolor = cyan 
}
\usepackage[capitalise]{cleveref} 

\definecolor{linkcolor}{rgb}{0.956, 0.298, 0.235} 
\colorlet{dark-green}{green!70!black} 

\journal{Neurocomputing}

\begin{document}

\begin{frontmatter}

\title{SU-YOLO: Spiking Neural Network for Efficient Underwater Object Detection}

\author[1,2]{Chenyang~Li\fnref{equal}}
\ead{chenyang@ctgu.edu.cn}

\author[3,2]{Wenxuan~Liu\fnref{equal}}
\ead{lwxfight@163.com}

\author[1]{Guoqiang~Gong\corref{cor1}}
\ead{guoqiang_gong@163.com}

\author[1]{Xiaobo~Ding}
\ead{84479265@qq.com}

\author[2,4]{Xian~Zhong}
\ead{zhongx@whut.edu.cn}

\address[1]{College of Computer and Information Technology, China Three Gorges University, Yichang, 443002, China}

\address[2]{Hubei Key Laboratory of Transportation Internet of Things, School of Computer Science and Artificial Intelligence, Wuhan University of Technology, Wuhan, 430070, China}

\address[3]{State Key Laboratory for Multimedia Information Processing, School of Computer Science, Peking University, Beijing, 100091, China}


\address[4]{State Key Laboratory of Maritime Technology and Safety, Wuhan University of Technology, Wuhan, 430063, China}

\cortext[cor1]{Corresponding author.}
\fntext[equal]{Contributed equally to this work.}

\begin{abstract}
Underwater object detection is critical for oceanic research and industrial safety inspections. However, the complex optical environment and the limited resources of underwater equipment pose significant challenges to achieving high accuracy and low power consumption. To address these issues, we propose Spiking Underwater YOLO (SU-YOLO), a Spiking Neural Network (SNN) model. Leveraging the lightweight and energy-efficient properties of SNNs, SU-YOLO incorporates a novel spike-based underwater image denoising method based solely on integer addition, which enhances the quality of feature maps with minimal computational overhead. In addition, we introduce Separated Batch Normalization (SeBN), a technique that normalizes feature maps independently across multiple time steps and is optimized for integration with residual structures to capture the temporal dynamics of SNNs more effectively. The redesigned spiking residual blocks integrate the Cross Stage Partial Network (CSPNet) with the YOLO architecture to mitigate spike degradation and enhance the model's feature extraction capabilities. Experimental results on \textsc{URPC2019} underwater dataset demonstrate that SU-YOLO achieves mAP of 78.8\% with 6.97M parameters and an energy consumption of 2.98 mJ, surpassing mainstream SNN models in both detection accuracy and computational efficiency. These results underscore the potential of SNNs for engineering applications. The code is available in https://github.com/lwxfight/snn-underwater.
\end{abstract}


\end{frontmatter}

\section{Introduction}

Underwater exploration has diverse applications, including aquatic life detection, facility safety inspections, and salvage operations. With advancements in remotely operated vehicles (ROVs) and autonomous underwater vehicles (AUVs), unmanned exploration methods are increasingly replacing traditional manual approaches, in which object detection plays a critical role. However, the complex underwater optical environment, characterized by low image brightness and high noise levels~\cite{Ju2025UnderwaterSN}, renders underwater object detection significantly more challenging than general object detection. In addition, the size and power constraints of ROVs and AUVs necessitate lightweight detection models to minimize resource consumption, further complicating the task. These factors make underwater object detection a demanding research area in computer vision.

Several algorithms optimized for underwater object detection have been proposed, such as YOLOv9s-UI~\cite{Pan2024OptimizationAA}, which demonstrate accuracy improvements over baseline models. However, incorporating attention mechanisms in these models often increases computational complexity, rendering them unsuitable for resource-constrained devices like AUVs. Lightweight models~\cite{Wang2020YOLONU, Wang2022ULOAU} have been introduced to address this issue by reducing parameter counts and computational demands. Unfortunately, these simplifications typically result in lower detection accuracy compared to state-of-the-art general object detection models.

Spiking neural networks (SNNs) offer a promising approach to achieving both low computation and high accuracy. As third-generation neural networks~\cite{Maass1996NetworksOS, mm/ZhongHLHDY024, tamd/YouZLWHYH24}, SNNs are inspired by the neuronal communication mechanisms in the brain. Unlike conventional artificial neural networks (ANNs), SNNs transmit information via discrete spikes rather than continuous values, which significantly reduces computational cost and energy consumption. Recent studies have explored applying SNNs to object detection, including Spiking-YOLO~\cite{Kim2019SpikingYOLOSN}, EMS-YOLO~\cite{Su2023DeepDS}, and SpikeYOLO~\cite{Luo2024IntegerValuedTA}. However, these works primarily focus on general object detection and lack optimizations for underwater conditions, such as image preprocessing and noise suppression, thereby limiting their performance in underwater environments.

To address these limitations, we propose SU-YOLO, an SNN-based underwater object detection model built on the ``you only look once'' (YOLO) framework~\cite{Redmon2015YouOL}. Given that YOLOv9~\cite{Wang2024YOLOv9LW} currently stands as a stable and high-performance object detection framework, SU-YOLO adopts its glean network as the baseline. First, to accommodate the energy constraints of underwater devices, we simplify the network structure and convert it into a spiking architecture to achieve minimal power consumption. During this process, we found that the conventional ResNet~\cite{He2015DeepRL} structure commonly used in SNNs leads to spike degradation even in shallow networks; consequently, we redesigned the lightweight SU-Blocks based on the CSPNet~\cite{Wang2019CSPNetAN} structure as the fundamental unit. Second, recognizing that existing SNNs lack effective methods for image noise removal, we developed a novel spiking underwater image denoising module. This module processes feature maps rather than the original image, removing both positive and negative noise to improve detection accuracy. In addition, existing batch normalization (BN) methods either ignore the temporal dynamics of SNNs~\cite{Wu2018DirectTF, Zheng2020GoingDW}, thereby losing potential information, or disregard the residual structure in networks~\cite{Kim2020RevisitingBN, Duan2022TemporalEB}, leading to excessive neuron charging and activation. To overcome these issues, we propose an optimized normalization method.

\begin{figure}
	\centering
	\includegraphics[width = 0.6\linewidth]{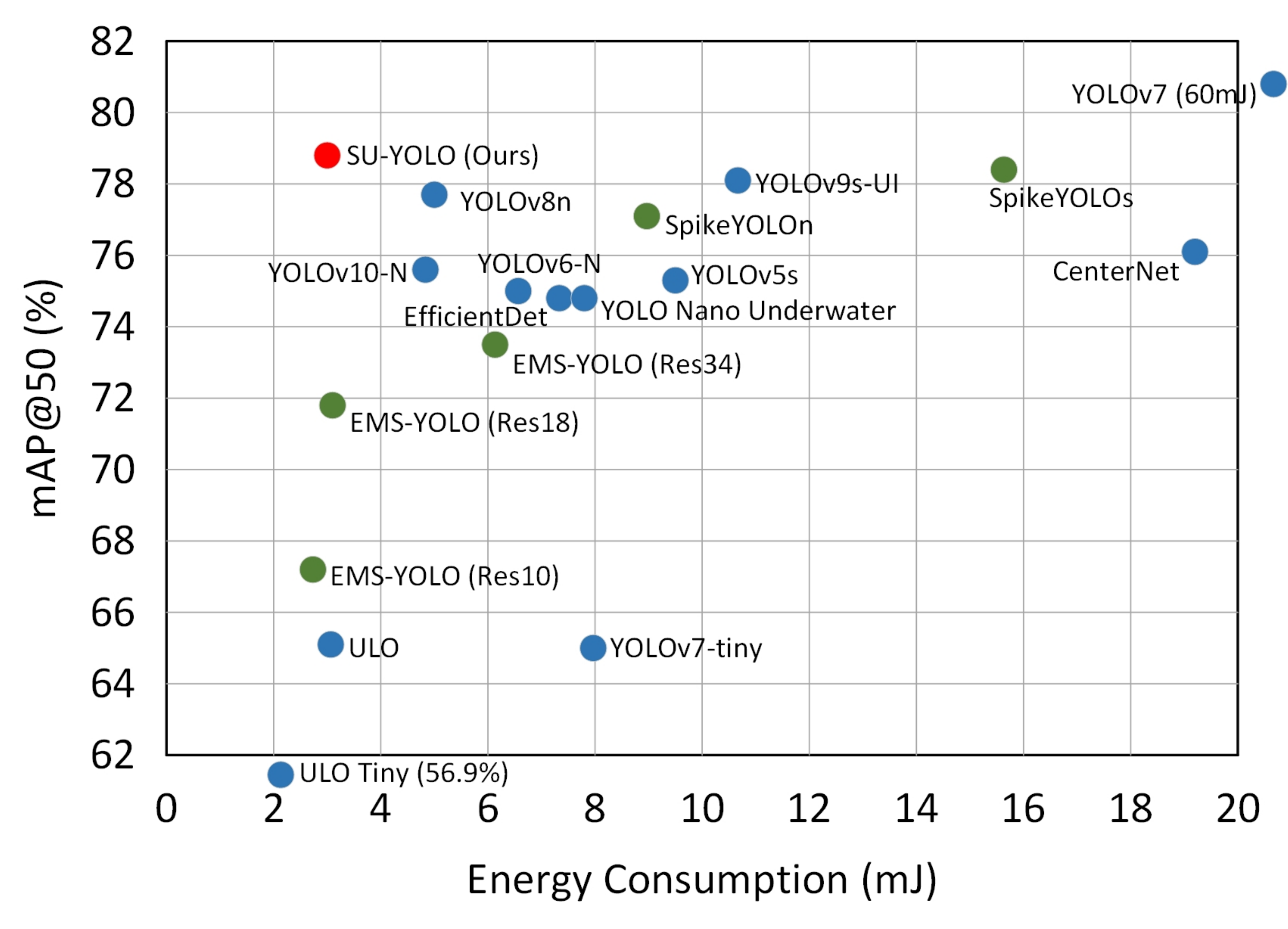}
	\caption{\textbf{Comparison of detection performance and energy consumption across various models on \textsc{URPC2019}.} The blue dots represent ANN models, while the green and red dots indicate SNN models.}
	\label{Fig:1}
\end{figure}

To the best of our knowledge, the proposed SU-YOLO is the first SNN specifically designed for underwater object detection. As shown in \cref{Fig:1}, our approach significantly enhances SNN performance in underwater object detection, achieving an mAP$_{0.5}$ of 78.8\% on \textsc{URPC2019} with 6.97M parameters and an energy consumption of 2.98 mJ, exceeding the performance of existing lightweight SNN and several ANN models.

The main contributions of this work are fourfold:

\begin{itemize}
	\item We develop a lightweight SNN model based on the YOLO framework that supports direct training while effectively mitigating spike degradation. The model outperforms existing SNNs and several ANNs in underwater object detection, achieving optimal accuracy with only four time steps.

	\item We design a novel spiking-compatible underwater image denoising method that integrates seamlessly into SNN architectures. This method requires only integer addition calculations, significantly improving feature map quality with minimal computational overhead.

	\item We propose an improved batch normalization method for SNNs, termed SeBN, which concatenates feature maps across time steps along the channel dimension for BN operations. SeBN is optimized for residual structures and outperforms classical BN and alternative methods, such as tdBN.

	\item Experiments on \textsc{URPC2019} and \textsc{UDD} underwater image datasets validate the feasibility and superior performance of SNNs for underwater object detection.

\end{itemize}

\section{Related Work}

\subsection{SNN Object Detection}

Object detection is a fundamental problem in computer vision. Since the introduction of R-CNN~\cite{Ren2015FasterRT}, ANNs have dominated deep learning–based object detection methods. The development of Spiking-YOLO~\cite{Kim2019SpikingYOLOSN} marked the first successful application of SNNs to object detection. Spiking-YOLO employs an ANN-SNN conversion method~\cite{Rueckauer2017ConversionOC}, representing activation values through spike firing rates by converting a pre-trained ANN into an SNN. However, this approach requires more than 1000 time steps to accurately represent the original values, which diminishes the computational efficiency of SNNs. Methods such as FSHNN~\cite{Chakraborty2021AFS} and Spike Calibration~\cite{Li2022SpikeCF} mitigate this issue by reducing the required time steps to 300 through optimized network structures. More recently, Fast-SNN~\cite{Hu2023FastSNNFS} introduces a signed IF neuron model and a layer-wise fine-tuning mechanism to address quantization and accumulation errors in ANN-SNN conversion, further compressing the time steps to 3. However, the use of negative spike computation in Fast-SNN compromises the binary nature of spikes, resulting in a non-standard spiking network structure.

An alternative to conversion is to directly train SNNs~\cite{Tavanaei2018DeepLI}, though this approach poses challenges due to the non-differentiability of the spike firing function. Lee \textit{et al.}~\cite{Lee2016TrainingDS} addressed this by treating the membrane potential as a differentiable parameter, which allows for gradient calculation and the development of deep, trainable SNNs. Subsequent surrogate gradient methods~\cite{Neftci2019SurrogateGL}, using functions such as the sigmoid and arctangent, have further facilitated SNN training. Notable examples include the spatiotemporal backpropagation algorithm (STBP)~\cite{Wu2017SpatioTemporalBF} and the spike-based backpropagation algorithm~\cite{Lee2019EnablingSB}.

Advancements in training methods have spurred research on SNN-based object detection. EMS-YOLO~\cite{Su2023DeepDS} employs a fully spiking ResNet~\cite{He2015DeepRL}, addressing issues in prior non-spike-based computations observed in MS-ResNet~\cite{Hu2021AdvancingSN} and SEW-ResNet~\cite{Fang2021DeepRL}, and demonstrates robust performance in general object detection. Similarly, SpikeYOLO~\cite{Luo2024IntegerValuedTA} introduces the I-LIF spiking neuron to convert non-spike-based computations into spike-based computations, thereby improving detection accuracy. However, these approaches are primarily designed for general object detection and perform suboptimally in underwater scenarios due to challenges such as blurry objects and severe noise.

\subsection{Underwater Object Detection}

Underwater object detection is particularly challenging because of the complex underwater environment and the performance limitations of underwater equipment. Several studies have attempted to overcome these challenges. For example, Song \textit{et al.}~\cite{song2020research} developed an algorithm based on Mask R-CNN~\cite{He2017MaskR} that employs multi-scale retinal enhancement and transfer learning to improve detection accuracy. These two-stage object detection methods, however, typically incur high computational costs.

In contrast, single-stage detection algorithms, such as the YOLO series, feature simpler architectures and lower computational requirements, making them more suitable for resource-constrained underwater devices. Zhang \textit{et al.}~\cite{Zhang2021LightweightUO} proposed a lightweight underwater detection network based on YOLOv4~\cite{Bochkovskiy2020YOLOv4OS}, which incorporates attention-based feature fusion to achieve accuracy comparable to the original model with fewer parameters. Further improvements were made by Zhang \textit{et al.}~\cite{Zhang2023AnIY} on YOLOv5 using newly designed blocks and MLLE image enhancement, surpassing the performance of YOLOv7~\cite{Wang2022YOLOv7TB} and YOLOv8~\cite{Varghese2024YOLOv8AN}. TC-YOLO~\cite{Liu2023UnderwaterOD}, an extension of YOLOv5s, integrates the Transformer~\cite{Vaswani2017AttentionIA} self-attention mechanism, significantly enhancing feature extraction for underwater objects. Despite these advances, the incorporation of attention mechanisms increases computational load, limiting their applicability in resource-constrained settings. Lightweight models such as ULO~\cite{Wang2022ULOAU} reduce parameter counts and computational costs (3.94M parameters and 3.42 GFLOPs) but still achieve an mAP of only 65.13\% on \textsc{URPC}, which lags behind mainstream object detection models. These limitations motivate our proposal of SU-YOLO, an SNN-based underwater object detection model.

\subsection{Underwater Image Denoising}

Image denoising is critical in underwater object detection. Conventional denoising methods, including mean filtering, Gaussian filtering, and non-local means (NLM), often blur noise and object boundaries, resulting in a loss of fine details. To address these shortcomings, Li \textit{et al.}~\cite{Li2021EnhancingUI} proposed an adaptive color and contrast enhancement (ACCE) framework combined with denoising, which leverages frequency-domain decomposition and variational optimization to enhance visual quality. You \textit{et al.}~\cite{You2023ResearchOI} employed a wavelet transform–based approach to denoise images prior to edge detection, using multi-resolution analysis to better preserve edge details. Although these methods provide improvements, they often struggle to remove complex noise without significantly blurring the image. Recent deep learning approaches have also been proposed. For example, Tian \textit{et al.}~\cite{Tian2023ACT} introduced the cross Transformer denoising network (CTNet) that combines Transformer and CNN for image denoising, significant advantages in images with complex noise. Ju \textit{et al.}~\cite{Ju2024TowardsMS} proposed a patch-based denoising diffusion model to address image blurring and color distortion, which effectively overcomes underwater suspended particles and light absorption. However, these methods typically involve intensive floating-point computations or require running separate models, making them less compatible with SNN architectures. Therefore, we have designed a dedicated lightweight underwater image denoising method tailored for SNNs.

\subsection{Normalization}

BN~\cite{Ioffe2015BatchNA} is widely used in ANNs to mitigate internal covariate shift by normalizing input statistics, thereby preventing gradient vanishing or explosion during training. However, for SNNs, which include an additional temporal dimension, conventional BN is less effective. Several improved normalization techniques have been proposed. NeuNorm~\cite{Wu2018DirectTF} normalizes input features along the channel dimension before updating membrane potentials, yet it does not fully resolve the gradient vanishing problem. Temporal dimension BN (tdBN)~\cite{Zheng2020GoingDW} normalizes inputs across all time steps, thus accounting for both temporal and spatial dimensions. The SpikingJelly~\cite{Fang2023SpikingJellyAO} framework also applies BN by flattening the time dimension into the batch dimension.

Although these methods are promising, our experiments indicate that independently normalizing each time step, rather than merging features across time steps, yields better performance. Approaches such as BNTT~\cite{Kim2020RevisitingBN} and TEBN~\cite{Duan2022TemporalEB} follow this design but fail to consider local parallel structures, such as shortcut connections, or the need for batch-scale fusion in SNNs. To address these issues, we propose SeBN, an optimized normalization method, which is detailed in subsequent sections.

\begin{figure}	
	\centering
	\includegraphics[width = \linewidth]{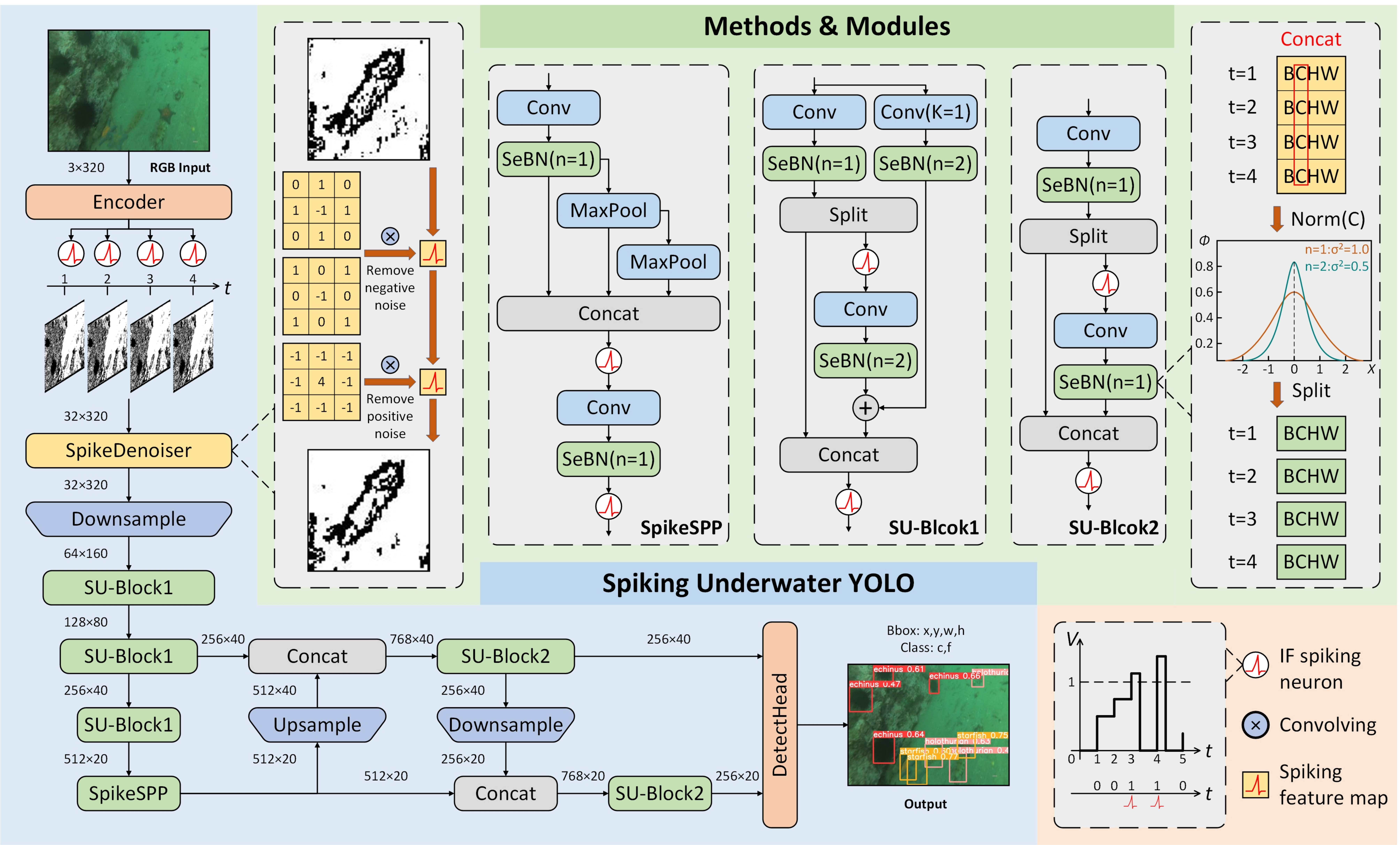}
	\caption{\textbf{Diagram of SU-YOLO and the structure of its modules.} In the main structural diagram, narrower graphics represent feature maps with reduced dimensions.}
	\label{Fig:2}
\end{figure} 

\section{Proposed Method}

\subsection{Spiking Underwater YOLO}

Our model is based on the YOLO framework and comprises the SU-Block1, SU-Block2, SpikeSPP, Encoder, SpikeDenoiser, and DetectHead modules, as illustrated in \cref{Fig:2}. These components are carefully designed and optimized to enhance detection performance in underwater environments. During the forward pass, normalized RGB image data (scaled to the range [0,1]) is fed into the network as floating-point values and then converted to spike outputs via integrate-and-fire (IF) neurons in the Encoder module. The SpikeDenoiser then performs image denoising and downsampling. Subsequently, the resulting feature map is processed by the backbone, which consists of three SU-Block1 residual blocks with increasing channel dimensions. Multi-scale features are extracted using the SpikeSPP module, followed by feature fusion in the Neck, and the DetectHead produces the final predictions.

In SU-YOLO, all spiking neurons are implemented as IF neurons. Unlike Leaky IF (LIF) neurons, IF neurons do not exhibit membrane voltage decay over time. Their simpler mathematical formulation and lower computational cost reduce overall power consumption. The IF neuron model is described by:
\begin{equation}\label{eqn-1}
	V_i^l(t) = \left(1 - \Theta_i^l(t) \right) \left(V_i^l(t-1) + \sum_j \left(\Theta_j^{l-1}(t) W_{i,j}^l + B_{i,j}^l \right) \right),
\end{equation}
where $V_i^l(t)$ denotes the membrane voltage of the $i$-th neuron in layer $l$ at time $t$, while $W$ and $B$ represent the weights and biases of the corresponding convolutional layer. The spike indicator $\Theta$ is defined as:
\begin{equation}\label{eqn-2}
	\Theta_i^l(t) = H \left(V_i^l(t) - V_\mathrm{th} \right),
\end{equation}
with $V_\mathrm{th}$ as the threshold voltage and $H(x)$ being the Heaviside step function:
\begin{equation}\label{eqn-3}
	H(x) = 
	\begin{cases} 
	1, & \text{if } x \geq 0, \\
	0, & \text{if } x < 0.
	\end{cases}
\end{equation}
When the accumulated membrane voltage exceeds the threshold, the neuron emits a spike and resets to zero potential, awaiting the next charge.

\subsection{SpikeDenoiser}

Noise severely degrades detection performance in underwater object detection. Existing SNN-based methods~\cite{Kim2019SpikingYOLOSN,Su2023DeepDS,Luo2024IntegerValuedTA} target general object detection and lack specialized denoising techniques, rendering them vulnerable to noise interference in underwater environments. To overcome this limitation while preserving the fully spiking nature of the network, we propose a denoising method that operates directly on the feature maps. In SNNs, the binary nature of feature maps causes underwater salt-and-pepper noise to appear as isolated pixels with inverted binary values compared to their neighbors. To effectively extract and correct these isolated points, we employ three convolution filters with the following kernel definitions:
\begin{equation}\label{eqn-4}
	K_1 = \begin{bmatrix} 
	0 & 1 & 0 \\
	1 & -1 & 1 \\
	0 & 1 & 0 
	\end{bmatrix},
\end{equation}
\begin{equation}\label{eqn-5}
	K_2 = \begin{bmatrix} 
	1 & 0 & 1 \\
	0 & -1 & 0 \\
	1 & 0 & 1 
	\end{bmatrix},
\end{equation}
\begin{equation}\label{eqn-6}
	K_3 = \begin{bmatrix} 
	-1 & -1 & -1 \\
	-1 & 4 & -1 \\
	-1 & -1 & -1 
	\end{bmatrix}.
\end{equation}
The input feature map, encoded with spikes, is sequentially convolved with $K_1$, $K_2$, and $K_3$. If the convolution result equals 4, the current pixel is inverted. This process is mathematically expressed as:
\begin{equation}\label{eqn-7}
	I_\mathrm{conv}(x,y) = \sum_{i=-1}^{1} \sum_{j=-1}^{1} I_\mathrm{in}(x+i, y+j) \times K_n(i+1, j+1),
\end{equation}
\begin{equation}\label{eqn-8}
	I_\mathrm{out}(x,y) = 
	\begin{cases} 
	1 - I_\mathrm{in}(x,y), & \text{if } I_\mathrm{conv}(x,y) = 4, \\
	I_\mathrm{in}(x,y), & \text{if } I_\mathrm{conv}(x,y) < 4,
	\end{cases}
\end{equation}
where $I_\mathrm{in}(x,y)$ is the pixel value at coordinates $(x,y)$, $I_\mathrm{conv}(x,y)$ is the convolution result, and $I_\mathrm{out}(x,y)$ is the output. Here, $K_n$ ($n = 1, 2, 3$) denotes the corresponding convolution kernel.

\begin{figure}
	\centering
	\includegraphics[width = \linewidth]{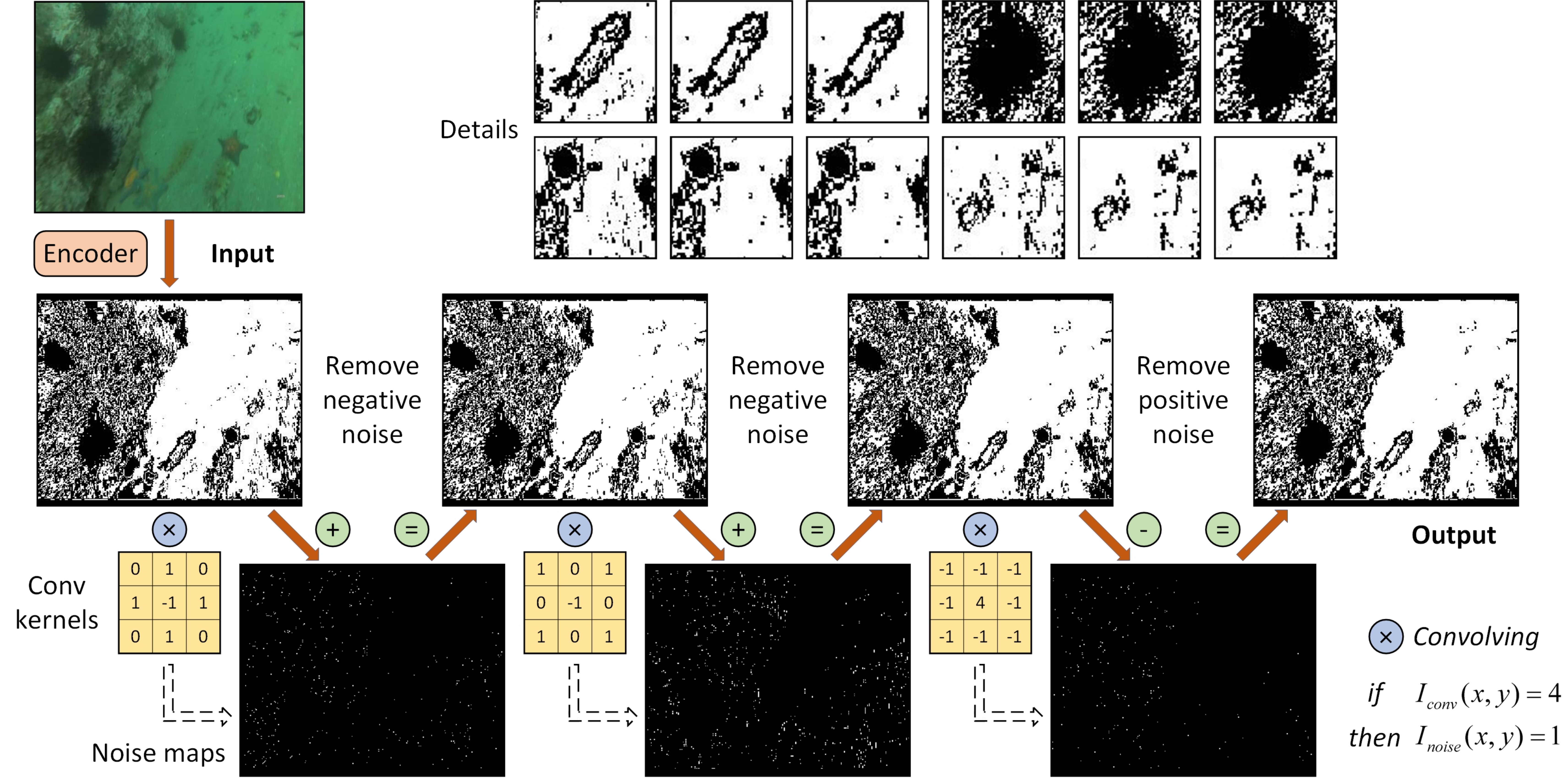}
	\caption{\textbf{Schematic diagram illustrating the processing flow and effects of the SpikeDenoiser module.}}
	\label{Fig:3}
\end{figure} 
 
As illustrated in \cref{Fig:3}, the method primarily fills missing pixels (Negative Noise) and removes isolated pixels (Positive Noise). Kernels $K_1$ and $K_2$ identify and fill pixels where the center is 0 and the surrounding pixels are 1, while $K_3$ targets and removes isolated pixels where the center is 1 and the surrounding pixels are 0. Experimental results indicate that an aggressive approach for filling Negative Noise, combined with a conservative strategy for removing Positive Noise, yields optimal denoising performance. Notably, the SpikeDenoiser operates only during inference and incurs minimal computational cost, less than 0.1 GSOPs and 0.01 mJ for a $320 \times 320$ image, while improving mAP by 3.5\% on \textsc{UDD}. This method maintains the fully spiking nature of the network since both the input and output feature maps remain binary, and the operations rely solely on integer additions, reducing energy consumption by up to $9\times$ and $37\times$ compared to floating-point addition and multiplication, respectively~\cite{Horowitz201411CE}. Its compatibility with various SNN architectures makes it a versatile solution for underwater image denoising.

\begin{figure}	
	\centering
	\includegraphics[width = 0.8\linewidth]{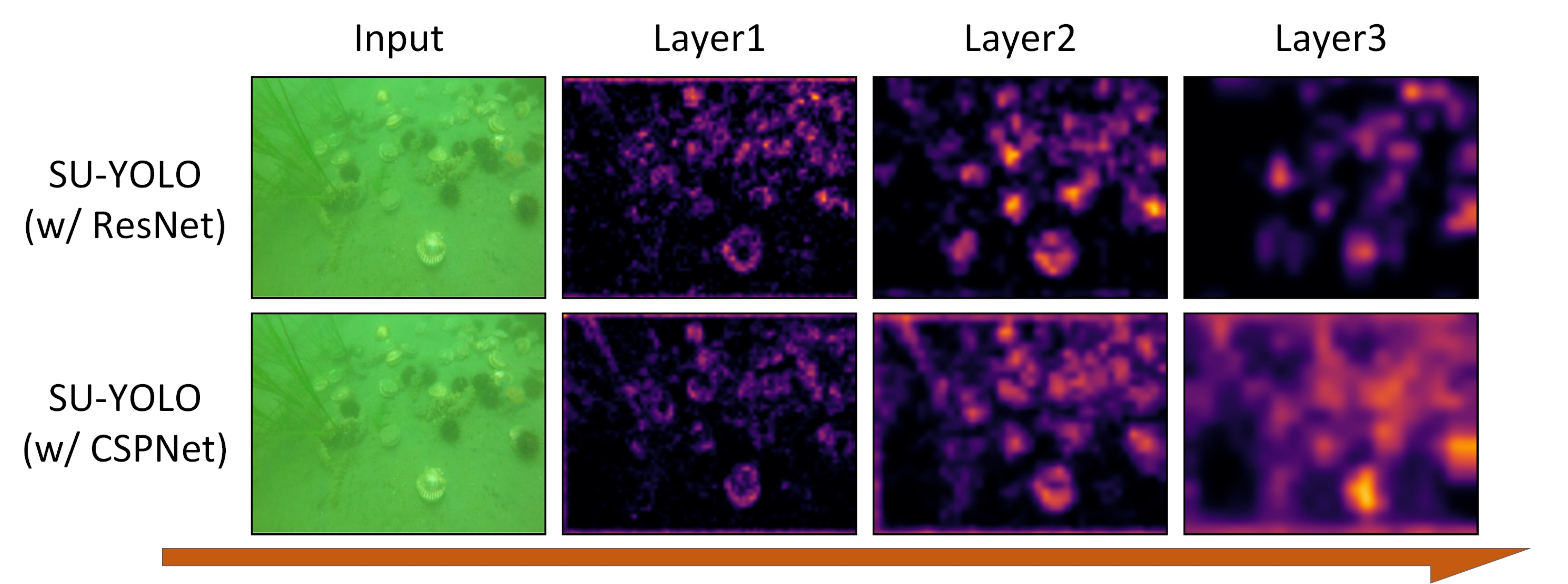}
	\caption{\textbf{Feature maps extracted from backbone layers when using ResNet or CSPNet in SNN.}}
	\label{Fig:4}
\end{figure} 

\subsection{SU-Block}

SU-Block1 and SU-Block2 serve as the fundamental residual blocks in our model. Underwater object detection typically requires shallow, computationally efficient networks due to device performance constraints. Although ResNet is commonly employed in existing SNNs~\cite{Su2023DeepDS, Hu2021AdvancingSN,Fang2021DeepRL}, the inherent low quality of underwater images limits the feature extraction capabilities of shallow ResNet architectures, leading to reduced detection accuracy. In contrast, CSPNet (Cross Stage Partial Network) provides enhanced feature representation even under challenging imaging conditions and with shallower network depths, thereby maintaining high accuracy while reducing computational costs. Moreover, we observed that a simple ResNet structure suffers from spike degradation in SNNs (see \cref{Fig:4}), as the firing rate decreases with increasing network depth, which can result in the loss of critical object features. CSPNet, however, effectively mitigates this degradation by maintaining robust neuron activity throughout the backbone. Therefore, we designed the SU-Block based on the CSPNet architecture.

SU-Block1, employed in the backbone for feature extraction, downsampling, and channel expansion, integrates CSPNet and shortcut connections (see \cref{Fig:2}). It is defined as:
\begin{equation}\label{eqn-9}
	X_1^l, X_2^l = \mathrm{Split} \left(\mathrm{Conv} \left(X_\mathrm{out}^{l-1} \right) \right),
\end{equation}
\begin{equation}\label{eqn-10}
	X_\mathrm{out}^l = \mathrm{IF} \left(\mathrm{Concat} \left(X_1^l, \mathrm{Conv} \left(\mathrm{IF} \left(X_2^l \right) \right) + \mathrm{Conv} \left(X_\mathrm{out}^{l-1} \right) \right) \right),
\end{equation}
where $X_\mathrm{out}^l$ denotes the output of the $l$-th layer, $\mathrm{IF}(\cdot)$ represents the IF neuron activation, $\mathrm{Conv}(\cdot)$ denotes a convolutional layer, $\mathrm{Split}(\cdot)$ splits the input equally along the channel dimension, and $\mathrm{Concat}(\cdot)$ concatenates inputs along the channel dimension.

Following a $3 \times 3$ convolution, the feature map is split into two groups. One group remains unchanged while the other undergoes an additional $3 \times 3$ convolution for further feature extraction. The two groups are then concatenated, activated, and output. To compensate for the absence of a residual path in the second group, a shortcut connection is introduced to preserve gradient flow. SU-Block2, used in the Neck for feature fusion, follows a similar design but without downsampling or additional residual connections, thereby enhancing parameter efficiency. It is defined as:
\begin{equation}\label{eqn-11}
	X_1^l, X_2^l = \mathrm{Split} \left(\mathrm{Conv} \left(X_\mathrm{out}^{l-1} \right) \right),
\end{equation}
\begin{equation}\label{eqn-12}
	X_\mathrm{out}^l = \mathrm{IF} \left(\mathrm{Concat} \left(X_1^l, \mathrm{Conv} \left(\mathrm{IF} \left(X_2^l \right) \right) \right) \right).
\end{equation}

\begin{figure}	
	\centering
	\includegraphics[width = 0.8\linewidth]{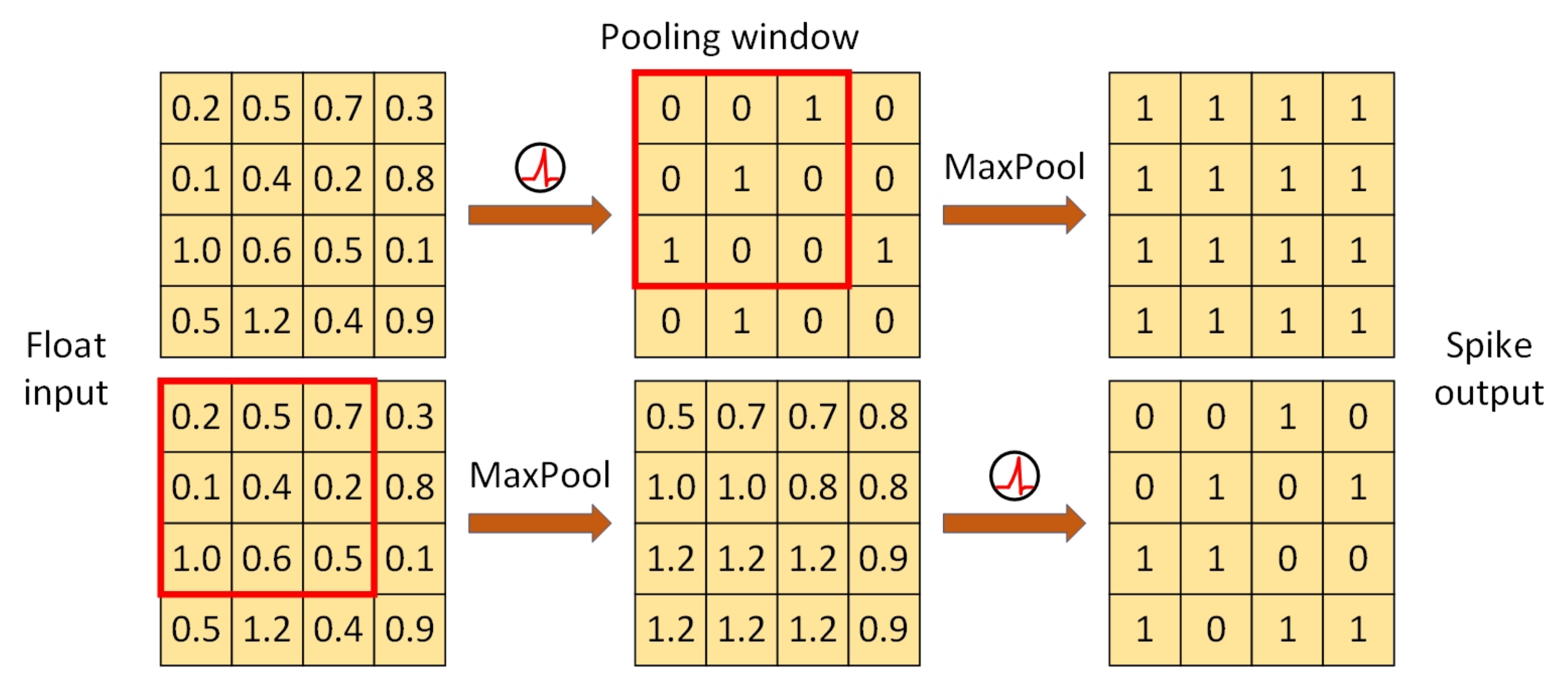}
	\caption{\textbf{Impact of pooling and activation sequence on output.}}
	\label{Fig:5}
\end{figure} 

\subsection{SpikeSPP}

The SpikeSPP module is a simplified version of spatial pyramid pooling (SPP)~\cite{He2014SpatialPP}, as depicted in \cref{Fig:5}. Its operations are defined by:
\begin{equation}\label{eqn-13}
	X_1^l = \mathrm{Conv} \left(X_\mathrm{out}^{l-1} \right),
\end{equation}
\begin{equation}\label{eqn-14}
	X_2^l = \mathrm{MaxPool} \left(X_1^l \right),
\end{equation}
\begin{equation}\label{eqn-15}
	X_3^l = \mathrm{MaxPool} \left(X_2^l \right),
\end{equation}
\begin{equation}\label{eqn-16}
	X_\mathrm{out}^l = \mathrm{IF} \left(\mathrm{Conv} \left(\mathrm{IF} \left(\mathrm{Concat} \left(X_1^l, X_2^l, X_3^l \right) \right) \right) \right),
\end{equation}
where $\mathrm{MaxPool}(\cdot)$ represents max-pooling, and $\mathrm{Concat}(\cdot)$ concatenates along the channel dimension. Although SPP provides multi-level pooling for extracting multi-scale features, pooling after activation can lead to significant information loss in SNNs due to the binary nature of spikes. In underwater object detection, where salt-and-pepper noise is prevalent, pooling prior to activation better preserves information. Experimental results further indicate that removing the third max-pooling layer reduces computational load without sacrificing performance. The resulting features are then passed to the IF neurons, which generate spikes with appropriate firing rates.

\subsection{Encoder and DetectHead}

The Encoder module converts floating-point input images into spikes. It applies preliminary feature extraction and normalization using conventional convolution and BN. The resulting feature map is duplicated $T$ times and passed through spike neurons, yielding binary spike outputs that complete the image encoding process.

The DetectHead adopts the decoupled head design from YOLOv9~\cite{Wang2024YOLOv9LW}, modified to handle spike-based inputs and outputs. Separate convolutional layers compute class probabilities and bounding box coordinates. Multi-level feature fusion is achieved by accepting inputs from two SU-Block2 modules, thereby capturing richer features. In the final spiking neuron of the DetectHead, the membrane potential threshold is set to infinity, allowing it to accumulate all floating-point values over each time step, which ensures more accurate predictions. Moreover, the parameter $n$ in SeBN is set to the number of time steps $T$, which reduces the range of single-step outputs and stabilizes the accumulated membrane potential, thereby improving gradient stability. The final network output is obtained by reading the membrane potential of these neurons.

\subsection{Separated Batch Normalization}
\label{subsec:SeBN}

In SNNs, each spiking neuron charges and fires independently at each time step, meaning that presynaptic inputs may exhibit different distribution characteristics over time. Normalizing feature maps across all time steps~\cite{Zheng2020GoingDW} can obscure these differences by forcing them into a single distribution. By assigning distinct normalization parameters for each time step, the network preserves these temporal variations, thereby enhancing its information capacity and expressive power. Additionally, residual network structures require careful adjustment of data distributions prior to summation~\cite{Kim2020RevisitingBN} to prevent neuron overcharging and excessive spike firing. To address these issues, we propose Separated Batch Normalization (SeBN), which normalizes feature maps independently across time steps while optimizing for integration with residual structures. SeBN is defined as:
\begin{equation}\label{eqn-17}
	y_{t,k} = \gamma_{t,k} \left(\frac{x_{t,k} - \mu_{t,k}}{\sqrt{n \left(\sigma_{t,k}^2 + \varepsilon \right)}} \right) + \beta_{t,k},
\end{equation}
where $x_{t,k}$ is the feature map of the $k$-th channel at time step $t$, $n$ is a positive integer representing the number of SeBN modules connected after the current module, and $\varepsilon$ is a small constant to prevent division by zero. The parameters $\gamma_{t,k}$ and $\beta_{t,k}$ are trainable, and $y_{t,k}$ is the normalized output. The mean $\mu_{t,k}$ and variance $\sigma_{t,k}^2$ are computed as:
\begin{equation}\label{eqn-18}
	\mu_{t,k} = \frac{1}{N} \sum_{i=1}^N (x_{t,k})_i,
\end{equation}
\begin{equation}\label{eqn-19}
	\sigma_{t,k}^2 = \frac{1}{N} \sum_{i=1}^N \left((x_{t,k})_i - \mu_{t,k} \right)^2,
\end{equation}
with $N$ representing the batch size.

\begin{figure}
	\centering
	\includegraphics[width = \linewidth]{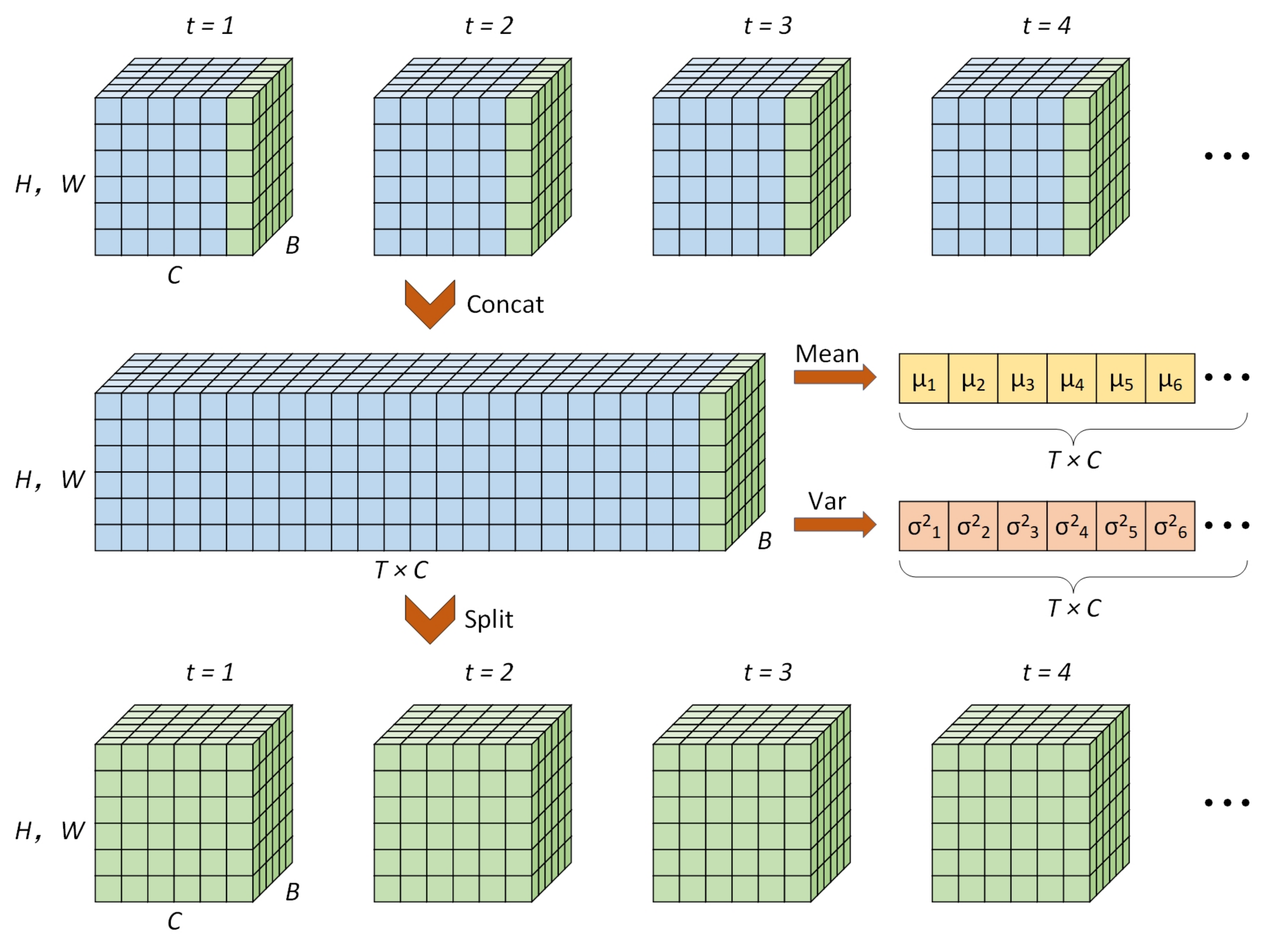}
	\caption{\textbf{Calculation and fusion methods of SeBN.} In the SeBN computation diagram, cubes represent the feature maps at each time step, where $B$ denotes the batch size, $C$ indicates the number of channels, $H$ and $W$ are the height and width of the feature maps, and $T$ represents the total number of time steps. The symbols $\mu_k$ and $\sigma_k^2$ denote the mean and variance of the $k$-th feature map.}
	\label{Fig:6}
\end{figure} 
 
As shown in \cref{Fig:6}, feature maps from all $T$ time steps are concatenated along the channel dimension and normalized using SeBN. The normalized feature maps are then evenly split and redistributed across the $T$ time steps. In this normalization, the factor $n$ adjusts the distribution of the input data from $N(0,1)$ to $N \left(0,\frac{1}{n} \right)$, ensuring that the sum in residual structures maintains a distribution of $N(0,1)$ and preventing excessive neuron activation. During training, running means and variances for each batch across $T \times C$ are updated via a moving average to compute $\mu$ and $\sigma_{\mathrm{run},t,k}^2$, which are then used during inference.

\begin{figure}	
	\centering
	\includegraphics[width = 0.8\linewidth]{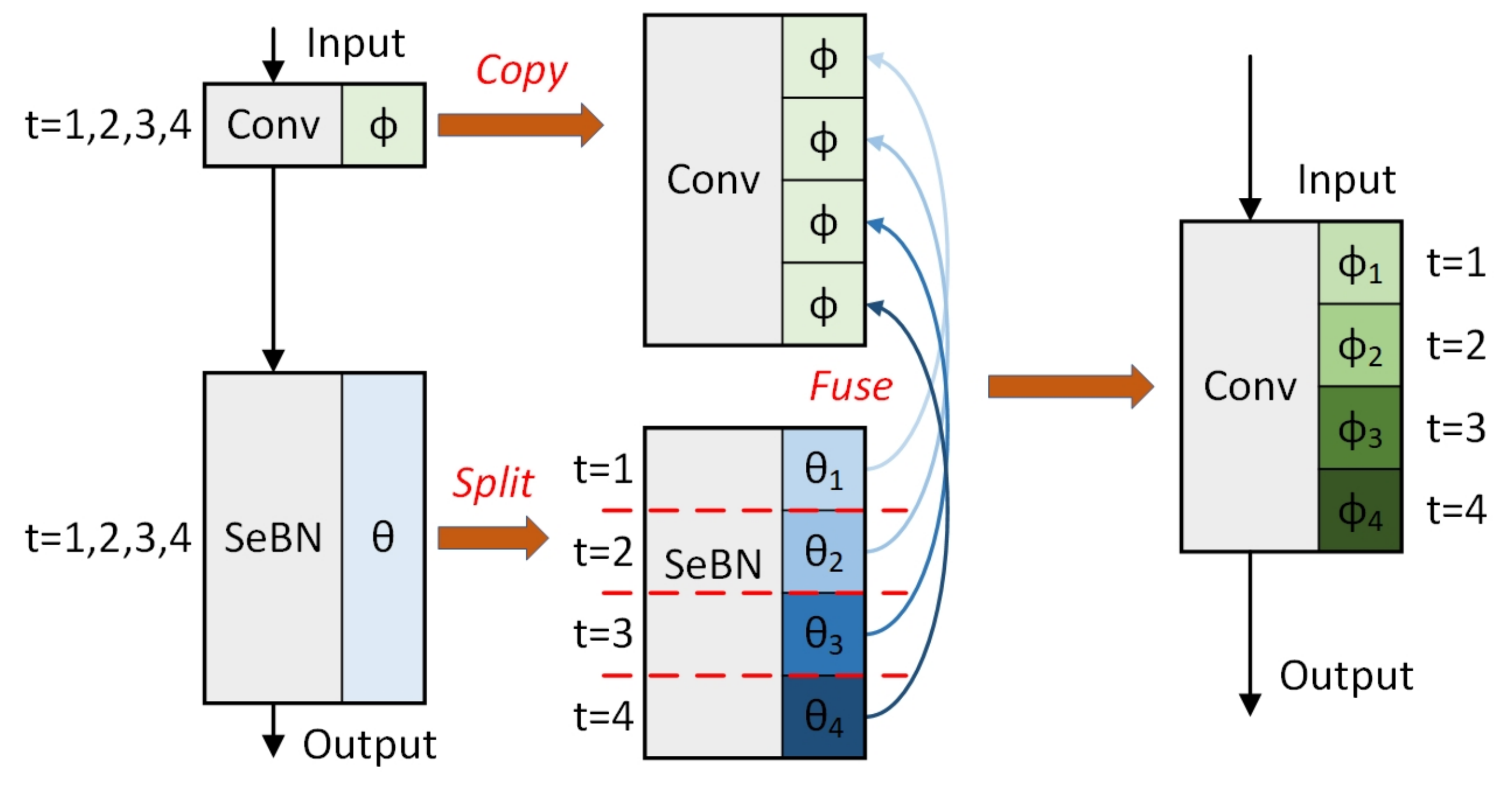}
	\caption{\textbf{Process of batch-scale fusion of SeBN.} Here, $\theta_t$ represents all parameters of the SeBN layer at time step $t$, $\phi$ represents parameters shared by the convolutional layer across all time steps, and $\phi_t$ denotes parameters of the convolutional layer at time step $t$.}
	\label{Fig:7}
\end{figure} 

To maintain the fully spiking nature of the SNN during inference, the normalization layer is removed, and a batch-scale fusion method is introduced for SeBN. As shown in \cref{Fig:7}, the weights and biases $\phi$ shared across all time steps in the original convolutional layer are duplicated for each time step. The parameters for each time step, $\theta_t$, are fused with $\phi$ to generate $T$ new sets of weights and biases, denoted by $\phi_t$. This fusion is performed as follows:
\begin{equation}\label{eqn-20}
	W_{t,k}' = \gamma_{t,k} \left(\frac{W_k}{\sqrt{n \left(\sigma_{\mathrm{run},t,k}^2 + \varepsilon \right)}} \right),
\end{equation}
\begin{equation}\label{eqn-21}
	B_{t,k}' = \gamma_{t,k} \left(\frac{B_k - \mu_{\mathrm{run},t,k}}{\sqrt{n \left(\sigma_{\mathrm{run},t,k}^2 + \varepsilon \right)}} \right) + \beta_{t,k},
\end{equation}
where $W_k$ and $B_k$ are the weights and biases for the $k$-th output channel in the original convolutional layer, and $\mu_{\mathrm{run},t,k}$ and $\sigma_{\mathrm{run},t,k}^2$ are the running mean and variance computed using an exponential moving average (EMA) over the training set. The fused weights and biases, $W_{t,k}'$ and $B_{t,k}'$, are then used for each time step during inference, eliminating the need for the SeBN layer and thereby reducing computational load and accelerating inference.

In the fused convolutional layer, each time step utilizes its own dedicated parameters. Although this increases the model's file size, the computational load and inference time remain unchanged. Given that storage is a relatively inexpensive resource, this trade-off is acceptable considering the significant accuracy improvements achieved by SeBN.

\section{Experimental Results}
\subsection{Datasets and Implementation Details}

\subsubsection{Datasets}

To evaluate our model's performance in underwater object detection, we conduct experiments on \textsc{URPC2019} and \textsc{UDD} datasets. \textsc{URPC2019} dataset~\cite{urpc2019-nrbk1_dataset}, released during the 2019 Underwater Robot Picking Contest, contains 4,707 labeled underwater images spanning four object classes: scallops, starfish, echinus, and holothurian. It is partitioned into a training set with 3,767 images, a validation set with 695 images, and a test set with 245 images. \textsc{UDD} dataset~\cite{Liu2020AND} comprises 2,227 labeled underwater images featuring three classes (scallops, holothurian, and echinus), and is divided into a training set of 1,827 images and a validation set of 400 images.

For experiments involving SeBN, we also utilize \textsc{Pascal VOC 2012} dataset~\cite{Everingham2014ThePV}, which consists of 11,540 labeled images across 20 object classes. This dataset is split into 5,717 training images and 5,823 validation images with object detection annotations.

\subsubsection{Implementation Details}

All experiments are performed on a single NVIDIA RTX 2080Ti (22GB) GPU, paired with an AMD Ryzen 5600 CPU and 32GB of RAM, running Ubuntu 22.04 LTS with Python 3.8 and PyTorch 2.0.0. For training, all neurons in our model are implemented as IF neurons using the SpikingJelly~\cite{Fang2023SpikingJellyAO} framework with a reset potential $V_{\mathrm{rst}} = 0$ and a threshold potential $V_{\mathrm{th}} = 1.0$. We employ the SGD optimizer with a learning rate of 0.001, a batch size of 16, and an input image size of $320 \times 320$. The model is trained for 300 epochs.


\subsubsection{Computational Cost}

In conventional ANNs, computational cost is measured in FLOPs, which represent the number of floating-point operations (additions or multiplications). However, since SNNs communicate via discrete spikes, calculating FLOPs is less straightforward. Instead, we use synaptic operations (SOPs) as the metric for computational cost. SOPs account for the convolution operations in each layer and the spike emissions from preceding layers, while also considering the floating-point additions involved in neuron charging. The formulas for FLOPs and SOPs are defined as follows:
\begin{equation}\label{eqn-22}
	\mathrm{FLOPs} = \sum_{i=1}^{N} \left(C_{\mathrm{in},i} \times k_{h,i} \times k_{w,i} \times C_{\mathrm{out},i} \times H_{\mathrm{out},i} \times W_{\mathrm{out},i}\right),
\end{equation}
\begin{equation}\label{eqn-23}
\begin{aligned}
	\mathrm{SOPs} = \sum_{t=1}^{T} \sum_{i=1}^{N} \Bigl(\left(f_{i-1,t} \times C_{\mathrm{in},i} \times k_{h,i} \times k_{w,i} + 1\right) \\
	\times C_{\mathrm{out},i} \times H_{\mathrm{out},i} \times W_{\mathrm{out},i}\Bigr),
\end{aligned}
\end{equation}
where $T$ is the total number of time steps, $N$ is the total number of network layers, $f_{i-1,t}$ denotes the firing rate of neurons in the $(i-1)$-th layer at time $t$, $C_{\mathrm{in},i}$ and $C_{\mathrm{out},i}$ are the numbers of input and output channels of the $i$-th convolutional layer, $k_{w,i}$ and $k_{h,i}$ are the kernel dimensions, and $W_{\mathrm{out},i}$ and $H_{\mathrm{out},i}$ are the output feature map dimensions.

Note that while FLOPs account for both multiplications and additions, SOPs only account for additions, thereby leading to lower energy consumption for the same operation count. For fair comparison, we also estimate energy consumption. Given that in convolutional neural networks the ratio of multiplications to additions is approximately 1:1, we approximate the number of multiply-accumulate operations (MACs) as FLOPs divided by 2. According to estimates in~\cite{Horowitz201411CE}, on a typical 45nm chip, a MAC operation consumes about 4.6 pJ and an addition consumes about 0.9 pJ.

\begin{figure}	
	\centering
	\includegraphics[width = 0.8\linewidth]{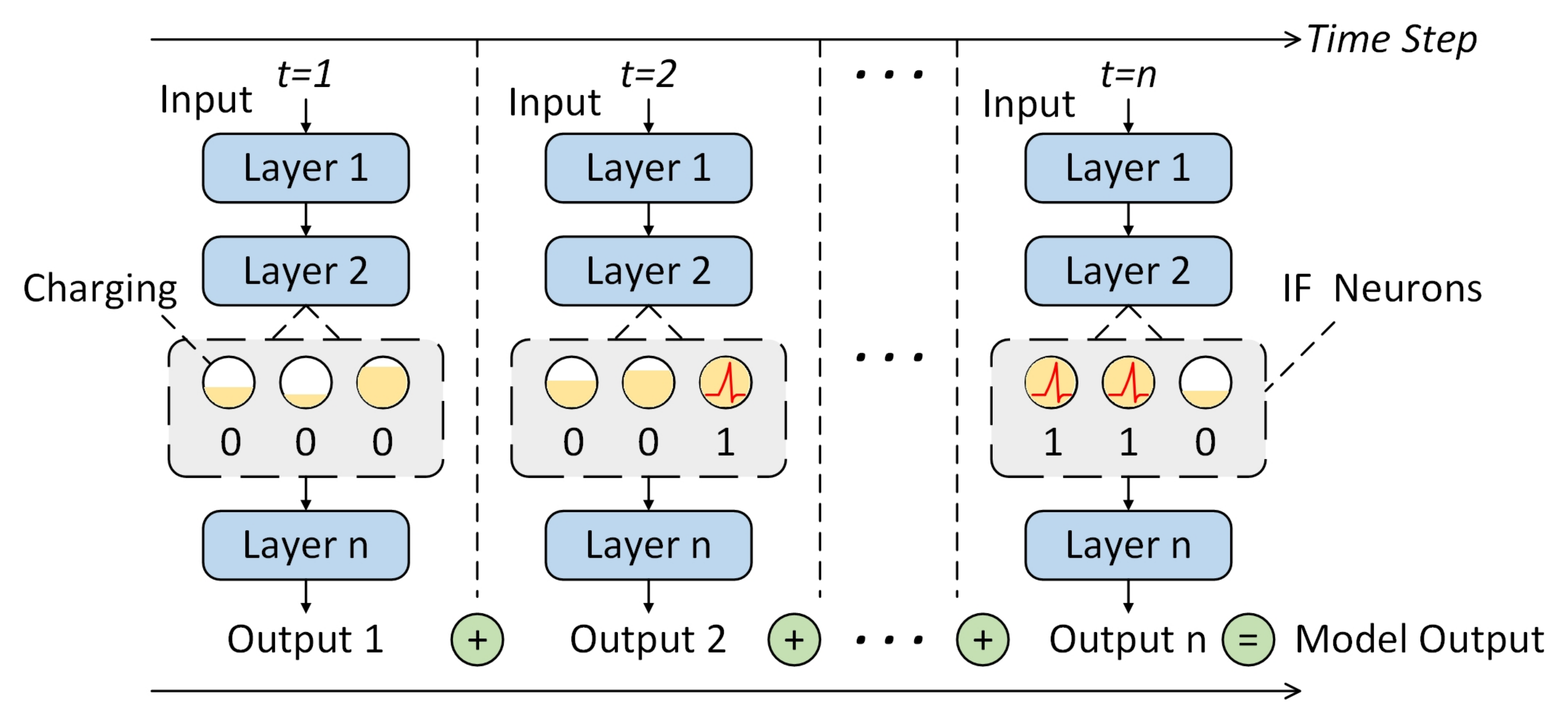}
	\caption{\textbf{Running process of SU-YOLO varies with time steps.}}
	\label{Fig:8}
\end{figure} 

\subsubsection{Time Step}
In SNNs, time steps serve as discrete units of temporal processing, during which neuronal states (\textit{e.g.}, membrane potential and spike generation) and synaptic dynamics are updated. As illustrated in \cref{Fig:8}, more time steps result in increased spike emissions, thereby enhancing the network's expressiveness. However, this also increases computational overhead and reduces inference speed. Consequently, for equivalent detection accuracy, methods that require fewer time steps are more advantageous in terms of energy efficiency and real-time performance.

\begin{table}[!t]
	\centering
	\setlength{\tabcolsep}{2.5pt}
	\scriptsize
	\begin{tabular}{l|ccc|cc|cc|c}
	\toprule[1.1pt]
	\multirow{2}[2]{*}{Model} & \multirow{2}[2]{*}{\begin{tabular}{c} Params \\ (M) \end{tabular}} & \multirow{2}[2]{*}{\begin{tabular}{c} GFLOPs/ \\ GSOPs \end{tabular}} & \multirow{2}[2]{*}{\begin{tabular}{c} Time \\ (Step) \end{tabular}} & \multicolumn{2}{c|}{mAP$_{0.5}$} & \multicolumn{2}{c|}{mAP$_{0.5:0.95}$} & \multirow{2}[2]{*}{\begin{tabular}{c} Energy \\ (mJ) \end{tabular}} \\
	\cmidrule(lr){5-6} \cmidrule(lr){7-8}
	& & & & \textsc{URPC} & \textsc{UDD} & \textsc{URPC} & \textsc{UDD} & \\
	\midrule
	\multicolumn{9}{c}{ANN} \\ 
	\midrule
	CenterNet (Res50)~\cite{Duan2019CenterNetKT} & 23.51 & 8.35 & - & 0.761 & - & - & - & 19.21 \\
	EfficientDet-D3~\cite{Tan2019EfficientDetSA} & 12.00 & 3.19 & - & 0.748 & - & - & - & 7.34 \\
	YOLOv5s~\cite{glenn_jocher_2021_5563715} & 7.20 & 4.13 & - & 0.753 & - & 0.420 & - & 9.50 \\
	YOLOv6-N~\cite{Li2023YOLOv6VA} & 4.70 & 2.85 & - & 0.750 & 0.537 & 0.410 & 0.228 & 6.56 \\
	YOLOv7-tiny~\cite{Wang2022YOLOv7TB} & 6.23 & 3.46 & - & 0.650 & 0.461 & 0.311 & 0.196 & 7.96 \\
	YOLOv7~\cite{Wang2022YOLOv7TB} & 36.49 & 25.88 & - & {0.808} & 0.533 & 0.434 & 0.236 & 59.52 \\
        YOLOv8n~\cite{Varghese2024YOLOv8AN} & 3.20 & 2.18 & - & 0.777 & 0.576 & 0.436 & 0.267 & 5.01 \\
	YOLOv10-N~\cite{Wang2024YOLOv10RE} & 2.71 & 2.10 & - & 0.756 & 0.542 & 0.425 & 0.251 & 4.83 \\
	ULO Tiny~\cite{Wang2022ULOAU} & 3.26 & 0.93 & - & 0.569 & - & - & - & {2.14} \\
	ULO~\cite{Wang2022ULOAU} & 3.76 & 1.34 & - & 0.651 & - & - & - & 3.08 \\
	YOLO Nano Underwater~\cite{Wang2020YOLONU} & 4.33 & 3.39 & - & 0.748 & - & - & - & 7.80 \\
	YOLOv9s-UI~\cite{Pan2024OptimizationAA} & 4.11 & 4.63 & - & 0.781 & - & {0.487} & - & 10.65 \\
	\midrule
	\multicolumn{9}{c}{ANN-SNN} \\ 
	\midrule
	Spiking-YOLO~\cite{Kim2019SpikingYOLOSN} $\ast$ & 8.68 & - & 3500 & 0.537 & 0.389 & 0.224 & 0.138 & - \\
	Fast-SNN (Dark19)~\cite{Hu2023FastSNNFS} & 20.80 & - & 15 & 0.604 & 0.454 & 0.250 & 0.163 & - \\
	\midrule
	\multicolumn{9}{c}{SNN} \\ 
	\midrule
	EMS-YOLO (Res10)~\cite{Su2023DeepDS} & 6.20 & 0.45 + 1.88 & 4 & 0.672 & 0.506 & 0.301 & 0.194 & \textbf{2.73} \\
	EMS-YOLO (Res18)~\cite{Su2023DeepDS} & 9.34 & 0.45 + 2.28 & 4 & 0.718 & 0.529 & 0.349 & 0.211 & 3.09 \\
	EMS-YOLO (Res34)~\cite{Su2023DeepDS} & 14.40 & 0.45 + 5.66 & 4 & 0.735 & 0.557 & 0.369 & 0.224 & 6.13 \\
	SpikeYOLOn~\cite{Luo2024IntegerValuedTA} & 12.64 & 0.08 + 11.05 & $4 \times 1$ & 0.735 & 0.499 & 0.392 & 0.220 & 10.13 \\
	SpikeYOLOn~\cite{Luo2024IntegerValuedTA} & 12.64 & 0.08 + 6.20 & $1 \times 4$ & 0.771 & 0.532 & 0.429 & 0.242 & 5.76 \\
	SpikeYOLOs~\cite{Luo2024IntegerValuedTA} & 22.05 & 0.11 + 16.61 & $4 \times 1$ & 0.761 & 0.529 & 0.418 & 0.244 & 15.20 \\
	SpikeYOLOs~\cite{Luo2024IntegerValuedTA} & 22.05 & 0.11 + 9.32 & $1 \times 4$ & 0.784 & 0.557 & \textbf{0.441} & 0.252 & 8.64 \\
	\rowcolor{gray!20}
	SU-YOLO (Ours) & 6.97 & 0.16 + 2.90 & 4 & \textbf{0.788} & \textbf{0.582} & 0.429 & \textbf{0.266} & 2.98 \\
	\bottomrule[1.1pt]
	\end{tabular}
	\caption{\textbf{Experimental results on \textsc{URPC2019} and \textsc{UDD}.} mAP$_{0.5}$ denotes mean average precision at an IoU threshold of 0.5, and mAP$_{0.5:0.95}$ spans IoU thresholds from 0.5 to 0.95. For SNNs, computational load includes FLOPs (from encoding) and SOPs. SpikeYOLO's Time is calculated as time steps $\times$ virtual time steps. mAP values marked with $\ast$ are from the ANN before conversion. The ANN-SNN conversion at $T = 3500$ is assumed lossless.}
	\label{tab:1}
\end{table}

\subsection{Comparison Experiments}

We compared our model with existing methods on \textsc{URPC2019} and \textsc{UDD}, with results summarized in \cref{tab:1}. Our directly trained SNN outperforms models obtained via ANN-SNN conversion while using fewer time steps. Compared to existing directly trained SNNs, our model achieves higher mAP with fewer parameters. Moreover, its performance is comparable to that of lightweight ANNs.

EMS-YOLO~\cite{Su2023DeepDS} currently serves as a benchmark for SNN-based object detection. On \textsc{URPC2019}, the lightest version of EMS-YOLO, based on ResNet10, achieves an mAP$_{0.5}$ of 0.672 with an energy consumption of 2.73 mJ, while a larger version using ResNet34 attains an mAP$_{0.5}$ of 0.735. SpikeYOLO~\cite{Luo2024IntegerValuedTA} further improves performance, reaching an mAP$_{0.5}$ of 0.784 with energy consumption of 8.64 mJ when using one time step and four virtual time steps. Our method surpasses these results, achieving an mAP$_{0.5}$ of 0.788 with only 6.97M parameters and a computational load of 2.90 GSOPs, while consuming merely 2.98 mJ, significantly outperforming other SNN approaches in underwater object detection.

We also compared our model with several ANNs. Although our mAP is slightly lower than that of advanced models like YOLOv7~\cite{Wang2022YOLOv7TB}, our energy consumption is only 5\% of YOLOv7's. When compared to lightweight ANNs with similar energy profiles, our model demonstrates superior performance. Additional experiments on \textsc{UDD} confirm that our model consistently outperforms alternative methods.

\subsection{Ablation Experiments}

A series of ablation experiments were performed using the same parameter settings as the comparative experiments to verify the effectiveness of our proposed modules.

\begin{table}
	\centering
	\setlength{\tabcolsep}{9pt}
	\scriptsize
	\begin{tabular}{ccccc|c|cc|cc}
	\toprule[1.1pt]
	\multirow{2}[2]{*}{BL} & \multirow{2}[2]{*}{SD} & \multirow{2}[2]{*}{SS} & \multirow{2}[2]{*}{SB2} & \multirow{2}[2]{*}{SB1} & \multirow{2}[2]{*}{\begin{tabular}{c} Params \\ (M) \end{tabular}} & \multicolumn{2}{c|}{mAP$_{0.5}$} & \multicolumn{2}{c}{mAP$_{0.5:0.95}$} \\
	\cmidrule(lr){7-8} \cmidrule(lr){9-10}
	& & & & & & \textsc{URPC} & \textsc{UDD} & \textsc{URPC} & \textsc{UDD} \\
	\midrule
	$\CIRCLE$ & $\Circle$ & $\Circle$ & $\Circle$ & $\Circle$ & 10.11 & 0.743 & 0.509 & 0.396 & 0.222 \\
	$\CIRCLE$ & $\CIRCLE$ & $\Circle$ & $\Circle$ & $\Circle$ & 10.11 & 0.752 & 0.523 & 0.398 & 0.232 \\
	$\CIRCLE$ & $\Circle$ & $\CIRCLE$ & $\Circle$ & $\Circle$ & 9.98 & 0.755 & 0.525 & 0.403 & 0.229 \\
	$\CIRCLE$ & $\Circle$ & $\CIRCLE$ & $\CIRCLE$ & $\Circle$ & 8.96 & 0.760 & 0.545 & 0.401 & 0.248 \\
	$\CIRCLE$ & $\Circle$ & $\CIRCLE$ & $\CIRCLE$ & $\CIRCLE$ & 6.97 & 0.781 & 0.567 & 0.422 & 0.257 \\
	\rowcolor{gray!20}
	$\CIRCLE$ & $\CIRCLE$ & $\CIRCLE$ & $\CIRCLE$ & $\CIRCLE$ & 6.97 & \textbf{0.788} & \textbf{0.582} & \textbf{0.429} & \textbf{0.266} \\
	\bottomrule[1.1pt]
	\end{tabular}
	\caption{\textbf{Ablation experiments on \textsc{URPC2019} and \textsc{UDD} for each module in SU-YOLO.} BL, SD, SS, SB2, and SB1 represent Baseline, SpikeDenoiser, SpikeSPP, SU-Block2, and SU-Block1, respectively.}
	\label{tab:2}
\end{table}

\subsubsection{Effectiveness of Modules in SU-YOLO}

We first evaluated the impact of our improved modules by replacing SU-Block1 and SU-Block2 with the basic residual block RepNCSPELAN4 from YOLOv9, substituting SPPELAN for SpikeSPP, and replacing all activation functions with spiking neurons to maintain a fully spiking network. The overall network architecture, including the number and size of feature maps, was kept consistent across all configurations to serve as a baseline. As shown in \cref{tab:2}, each module incrementally reduced the parameter count and improved accuracy. With all improvements applied, mAP$_{0.5}$ increased by 4.5\% and mAP$_{0.5:0.95}$ by 3.3\%. Furthermore, introducing the SpikeDenoiser module independently resulted in a 0.9\% mAP$_{0.5}$ gain over the baseline.

\begin{table}
	\centering
 	\setlength{\tabcolsep}{10pt}
	\scriptsize
	\begin{tabular}{l|cc|cc|c}
	\toprule[1.1pt]
	\multirow{2}[2]{*}{Method} & \multicolumn{2}{c|}{mAP$_{0.5}$} & \multicolumn{2}{c|}{mAP$_{0.5:0.95}$} & \multirow{2}[2]{*}{\begin{tabular}{c} Energy \\ (mJ) \end{tabular}} \\
	\cmidrule(lr){2-3} \cmidrule(lr){4-5}
	& \textsc{URPC} & \textsc{UDD} & \textsc{URPC} & \textsc{UDD} & \\
	\midrule
	- & 0.781 & 0.567 & 0.422 & 0.257 & 2.9700\\
	Mean Filter~\cite{Shao2021AnIA} & 0.780 & 0.557 & 0.414 & 0.248 & \textbf{2.9732}\\
	Gaussian Filter~\cite{Yu2019MemristorBL} & 0.783 & 0.573 & 0.415 & 0.261 & 2.9821\\
	\rowcolor{gray!20}
	SpikeDenoiser & \textbf{0.788} & \textbf{0.582} & \textbf{0.429} & \textbf{0.266} & 2.9751 \\
	\bottomrule[1.1pt]
	\end{tabular}
	\caption{\textbf{Performance and energy consumption when using different denoising methods in SU-YOLO.}}
	\label{tab:3}
\end{table}

\subsubsection{Effectiveness of SpikeDenoiser}

To evaluate computational cost, we compared conventional image denoising methods with our proposed SpikeDenoiser in \cref{tab:3}. For fairness, both the Mean Filter~\cite{Shao2021AnIA} and Gaussian Filter~\cite{Yu2019MemristorBL} were applied to the raw images before network input to preserve the network's binary nature. Experimental results indicate that the Mean Filter degrades mAP, likely due to its strong blurring effect, while the Gaussian Filter slightly improves accuracy by better preserving edge information. In contrast, SpikeDenoiser yields significant mAP improvements on both datasets with minimal additional energy consumption.

\begin{figure}	
	\centering
	\includegraphics[width = 0.8\linewidth]{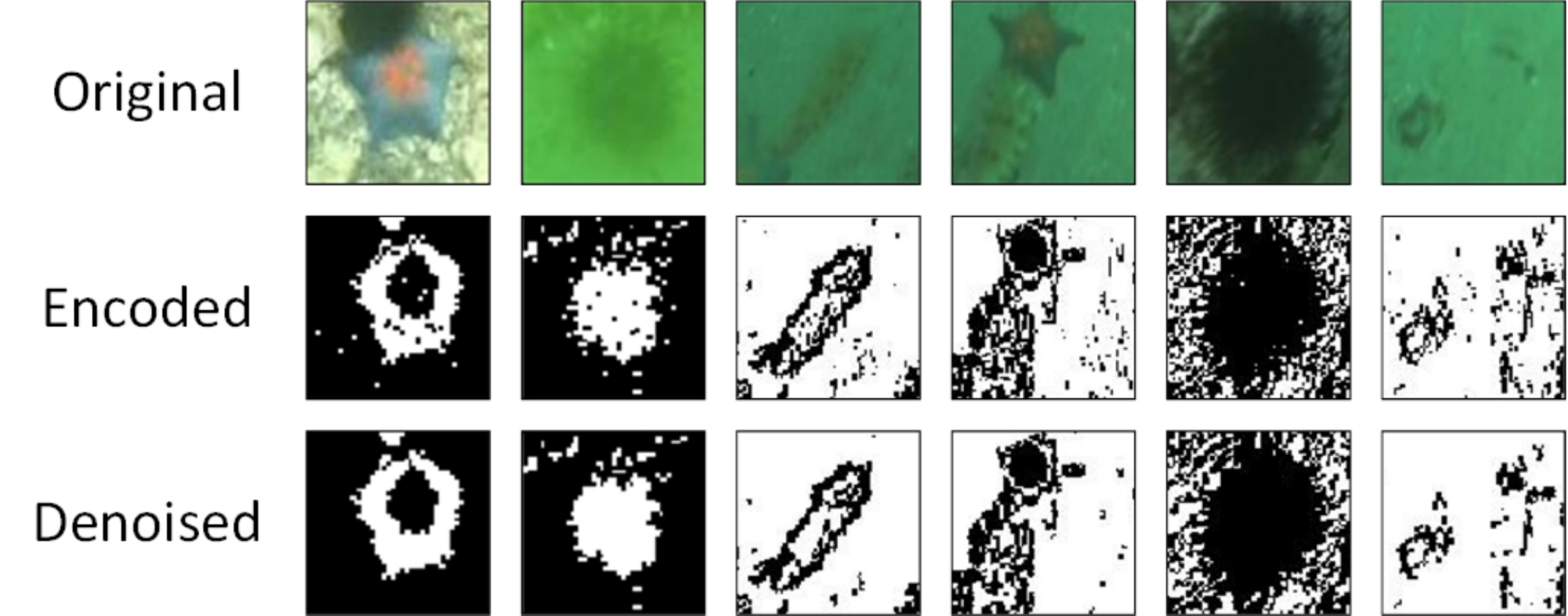}
	\caption{\textbf{Denoising effect of the SpikeDenoiser.}}
	\label{Fig:9}
\end{figure}

Visualization results in \cref{Fig:9} further demonstrate that SpikeDenoiser effectively removes noise from feature maps, delivering clearer inputs to the backbone network.

\begin{table}
	\centering
	\setlength{\tabcolsep}{10pt}
	\scriptsize
	\begin{tabular}{ll|cc|cc}
	\toprule[1.1pt]
	\multirow{2}{*}{Model} & \multirow{2}{*}{Method} & \multicolumn{2}{c|}{\textsc{URPC2019}} & \multicolumn{2}{c}{\textsc{Pascal VOC 2012}} \\
	\cmidrule(lr){3-4}\cmidrule(lr){5-6}
	& & mAP$_{0.5}$ & mAP$_{0.5:0.95}$ & mAP$_{0.5}$ & mAP$_{0.5:0.95}$ \\
	\midrule
	\multirow{4}{*}{EMS-YOLO (Res18)} 
	& BN~\cite{Ioffe2015BatchNA} & 0.698 & 0.334 & 0.514 & 0.310 \\
	& tdBN~\cite{Zheng2020GoingDW} & 0.713 & 0.350 & 0.522 & 0.324 \\
	& TEBN~\cite{Duan2022TemporalEB} & 0.705 & 0.335 & 0.519 & 0.316 \\
	& SeBN & \textbf{0.726} & \textbf{0.362} & \textbf{0.527} & \textbf{0.324} \\
	\midrule
	\multirow{4}{*}{SU-YOLO (Ours)} 
	& BN~\cite{Ioffe2015BatchNA} & 0.735 & 0.387 & 0.438 & 0.266 \\
	& tdBN~\cite{Zheng2020GoingDW} & 0.776 & 0.423 & 0.536 & 0.346 \\
	& TEBN~\cite{Duan2022TemporalEB} & 0.776 & 0.427 & 0.532 & 0.347 \\
	& \cellcolor{gray!20}SeBN & \cellcolor{gray!20}\textbf{0.788} & \cellcolor{gray!20}\textbf{0.429} & \cellcolor{gray!20}\textbf{0.537} & \cellcolor{gray!20}\textbf{0.350} \\
	\bottomrule[1.1pt]
	\end{tabular}
	\caption{\textbf{Detection results for BN, tdBN, TEBN, and SeBN on \textsc{URPC2019} and \textsc{Pascal VOC 2012}.}} 
	\label{tab:4}
\end{table}

\subsubsection{Effectiveness of SeBN}

We compared our novel SeBN with conventional BN methods, as summarized in \cref{tab:4}. Models using standard BN perform relatively poorly, while tdBN and TEBN yield moderate improvements. Notably, in the EMS-YOLO model, where residual blocks are heavily utilized, TEBN underperforms tdBN, likely due to its inadequate handling of element-wise summation in residual structures. In contrast, in SU-YOLO, where fewer such residual blocks are present, TEBN and tdBN exhibit similar performance; replacing both with SeBN further improves detection accuracy. We also evaluated SeBN on EMS-YOLO and on \textsc{Pascal VOC 2012} for general object detection, where SeBN consistently achieved the highest mAP. These results indicate that SeBN has broad applicability and generalizes well to other datasets and SNN architectures.

\begin{figure}
	\centering
	\begin{tabular}{ccc}
	\includegraphics[width = 0.3\linewidth]{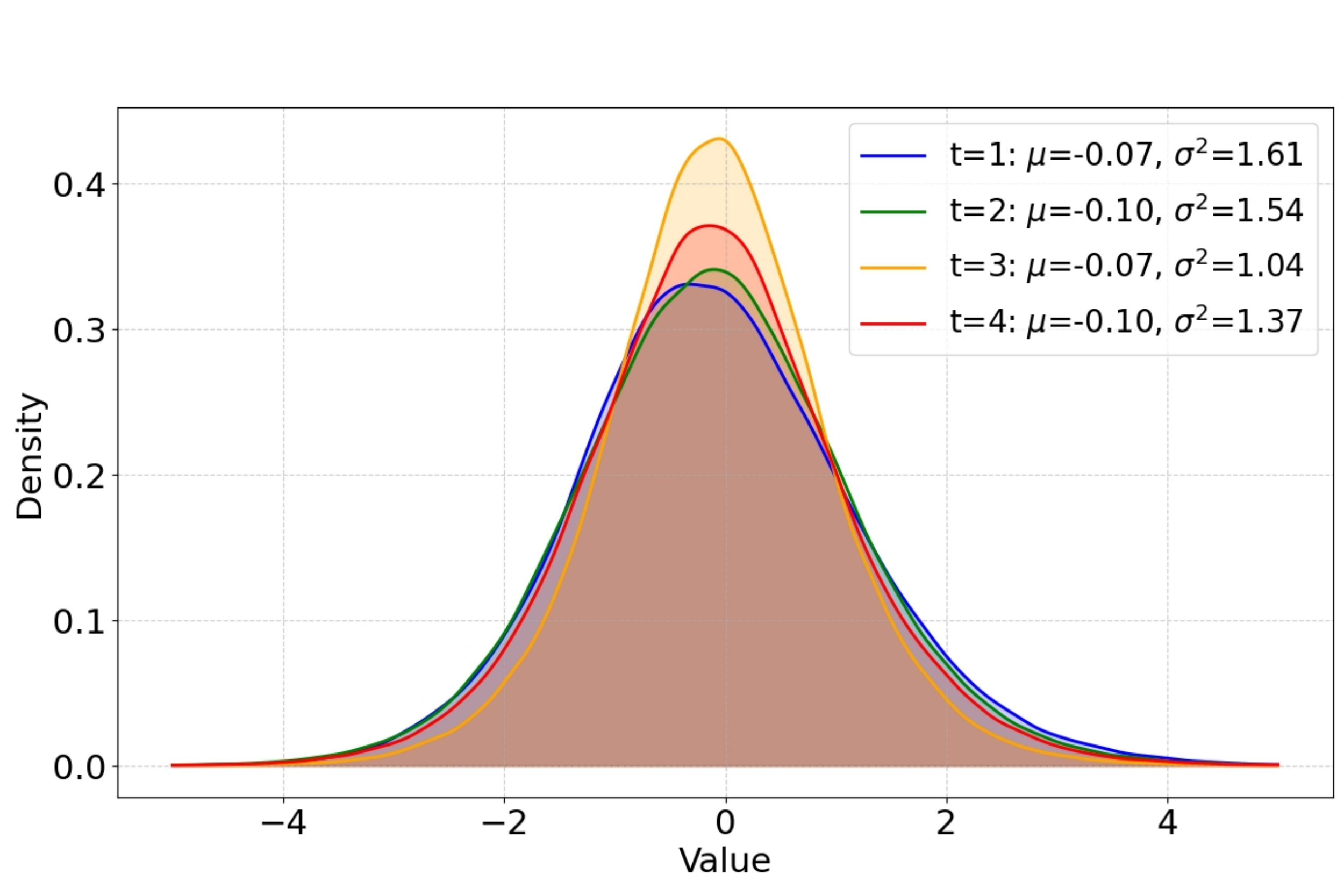} &
	\includegraphics[width = 0.3\linewidth]{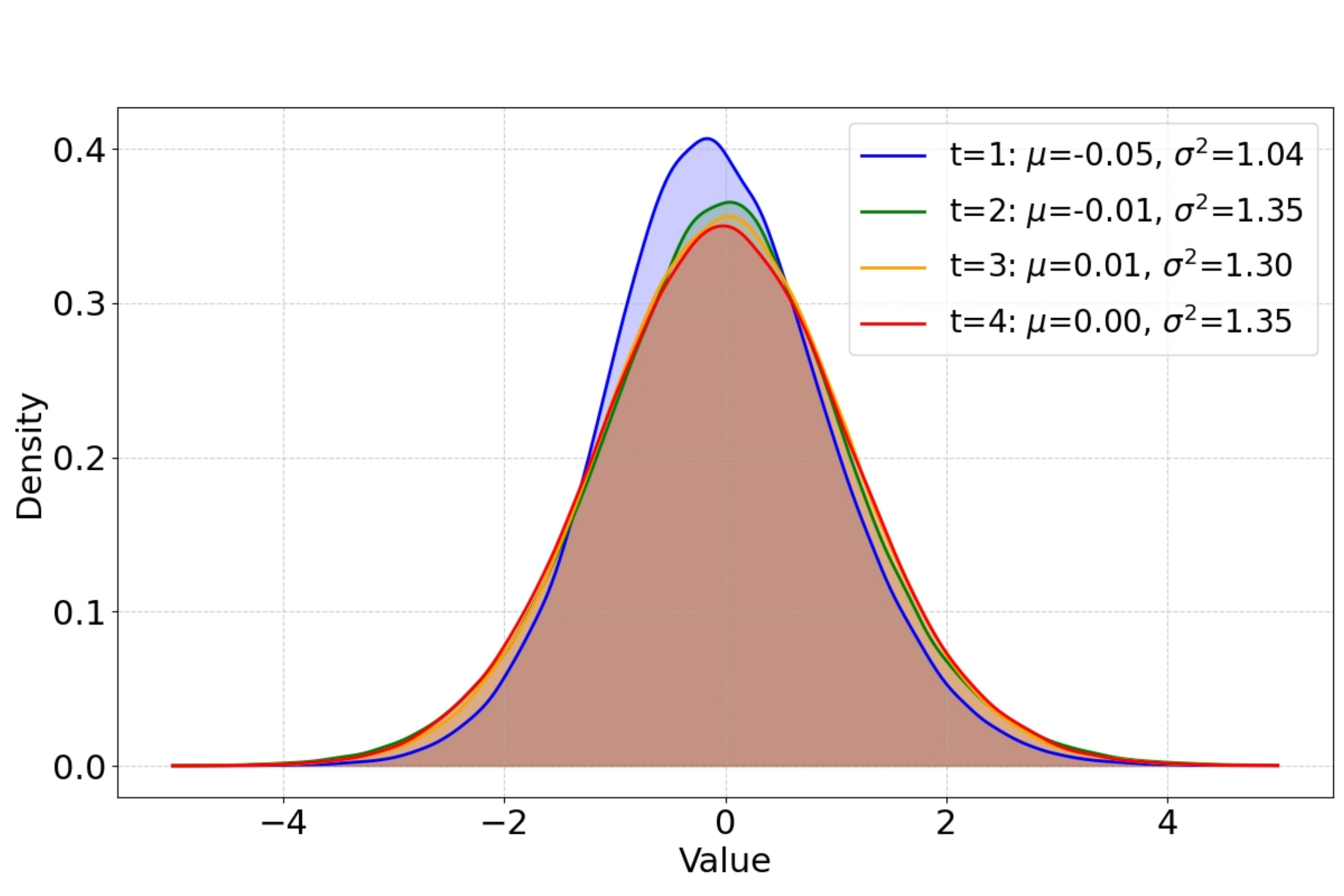} &
	\includegraphics[width = 0.3\linewidth]{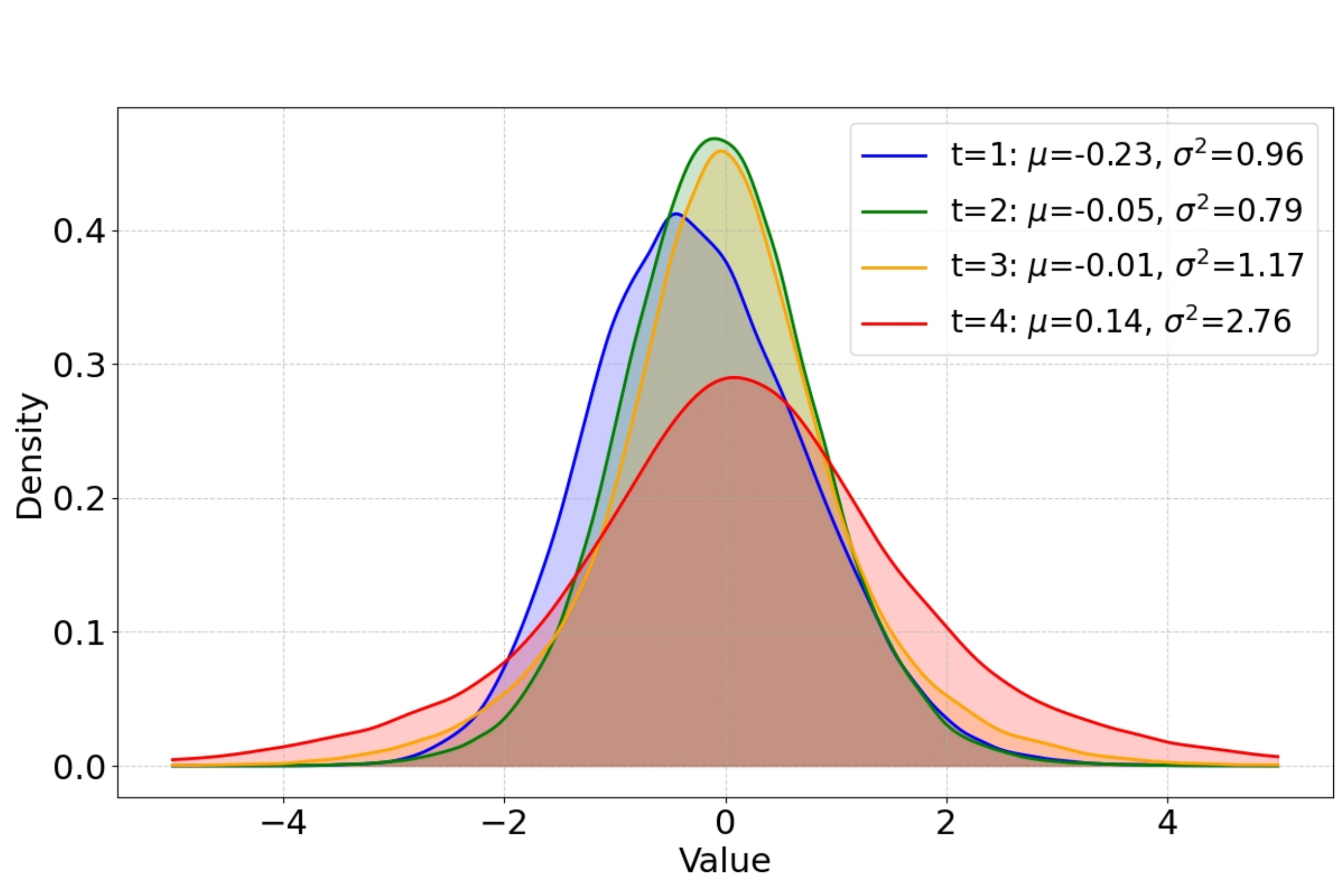} \\
	\footnotesize{(a) BN} &
	\footnotesize{(b) tdBN} &
	\footnotesize{(c) TEBN}
	\end{tabular}
	\begin{tabular}{cc}
	\includegraphics[width = 0.3\linewidth]{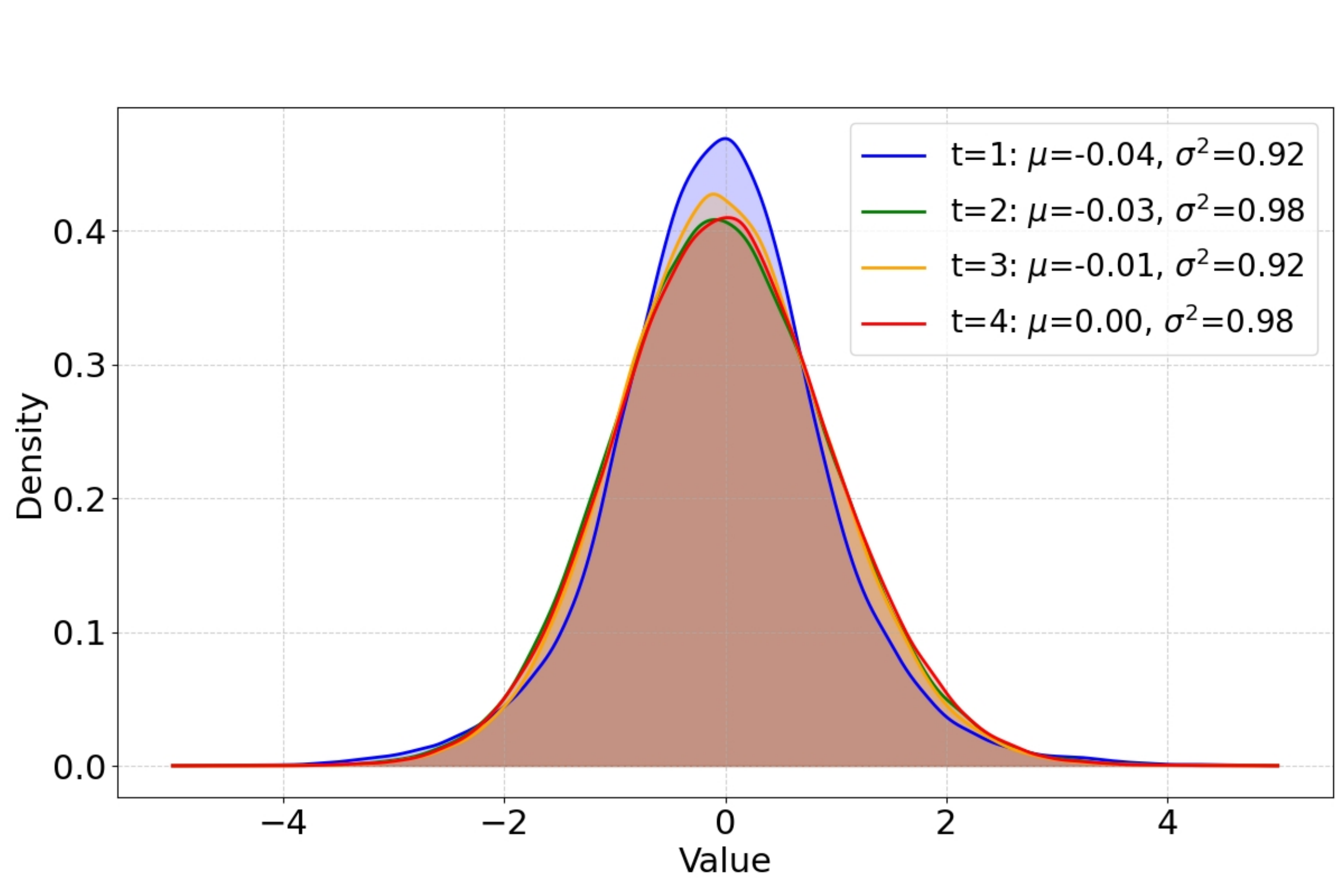} &
	\includegraphics[width = 0.3\linewidth]{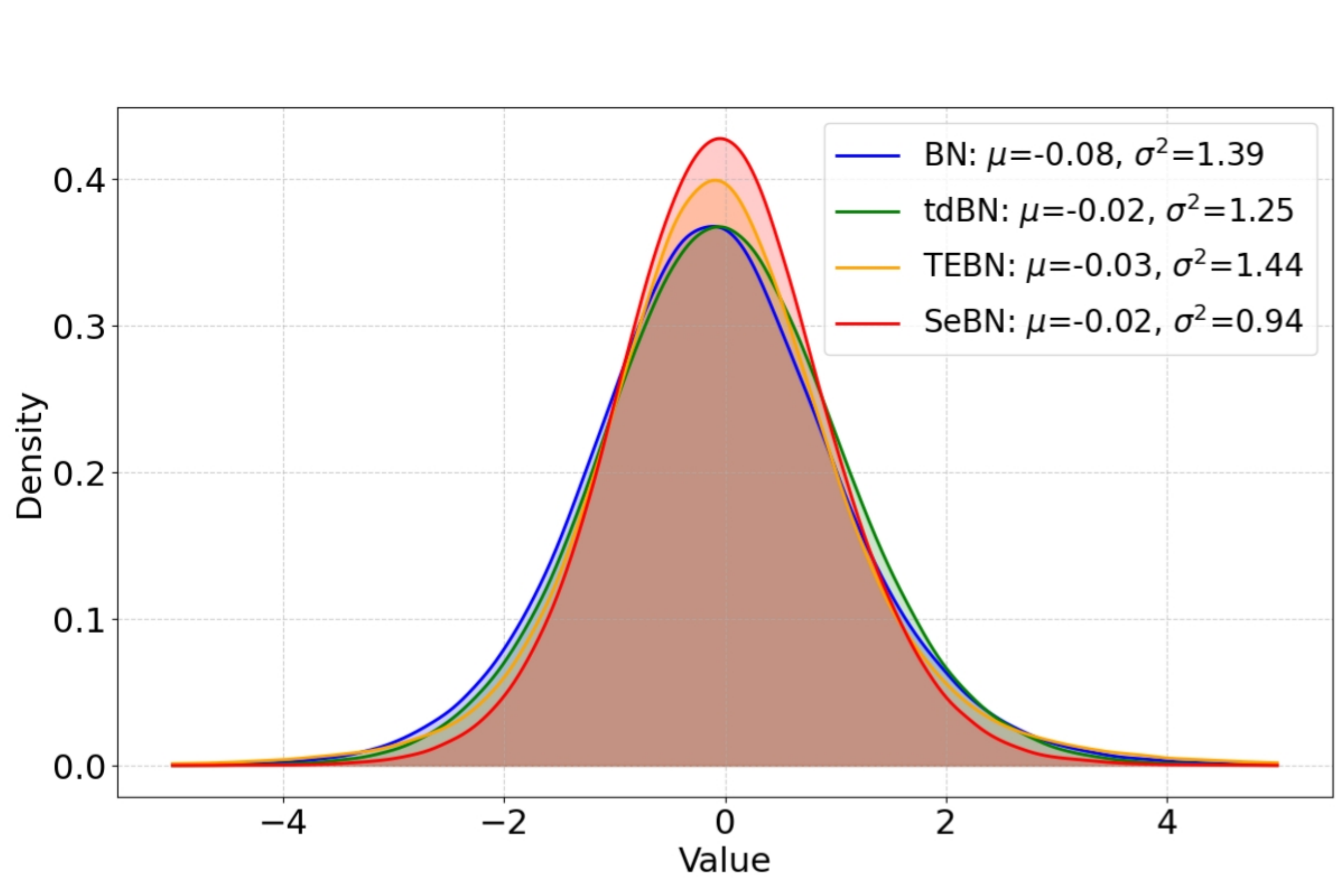} \\
	\footnotesize{(d) SeBN} & 
	\footnotesize{(e) Overall}
	\end{tabular}
	\caption{\textbf{Data distribution after a element summation residual block with different BN methods on \textsc{URPC2019}.} $\mu$ and $\sigma^2$ are the mean and variance of data.}
	\label{Fig:10}
\end{figure}

To illustrate the differences between SeBN and other normalization methods, we visualized the data distributions in \cref{Fig:10}. The distributions were derived from the outputs after the final residual block in layer 3 of SU-YOLO. The results reveal significant instability in the BN and TEBN distributions, with marked fluctuations in mean and variance across time steps. In particular, TEBN exhibits a variance as high as 2.76 at $t=1$, likely due to unoptimized residual addition that amplifies the output range. Although tdBN shows smaller fluctuations, its variance significantly deviates from the desired standard normal distribution. In contrast, SeBN maintains a stable distribution that closely approximates $N(0,1)$ at each time step. An overall comparison, averaged across time steps, further confirms the superior stability and performance of SeBN.

\begin{table}
	\centering
 	\setlength{\tabcolsep}{7pt}
	\scriptsize
	\begin{tabular}{l|c|cc|cc|ccc}
	\toprule[1.1pt]
	\multirow{2}[2]{*}{\begin{tabular}{c} Residual \\ Blocks \end{tabular}} & \multirow{2}[2]{*}{\begin{tabular}{c} Params \\ (M) \end{tabular}} & \multicolumn{2}{c|}{mAP$_{0.5}$} & \multicolumn{2}{c|}{mAP$_{0.5:0.95}$} & \multicolumn{3}{c}{Firing Rate} \\
	\cmidrule(lr){3-4} \cmidrule(lr){5-6} \cmidrule(lr){7-9}
	& & \textsc{URPC} & \textsc{UDD} & \textsc{URPC} & \textsc{UDD} & Layer1 & Layer2 & Layer3 \\
	\midrule
	ANN-Res18 & 9.20 & 0.754 & 0.530 & 0.406 & 0.244 & - & - & - \\
	EMS-Res18 & 9.23 & 0.722 & 0.506 & 0.370 & 0.234 & 0.152 & 0.119 & 0.115 \\
	\rowcolor{gray!20}
	SU-Block & \textbf{6.97} & \textbf{0.788} & \textbf{0.582} & \textbf{0.429} & \textbf{0.266} & 0.127 & 0.154 & 0.188 \\
	\bottomrule[1.1pt]
	\end{tabular}%
	\caption{\textbf{Experimental results of replacing different residual blocks in SU-YOLO.}}
	\label{tab:5}%
\end{table}

\subsubsection{Residual Block Replacement}

Residual blocks are the core of feature extraction and significantly impact detection accuracy. To assess the performance of our designed SU-Block, we experimented with different residual blocks while keeping the overall network structure constant. We calculated the average spike firing rate of various residual blocks on \textsc{URPC2019}. As shown in \cref{tab:5}, SU-Blocks employing the CSPNet structure achieved firing rates of 12.7\%, 15.4\%, and 18.8\%, which are higher than the firing rates of EMS-Blocks using the ResNet structure (15.2\%, 11.9\%, and 11.5\%). This observation aligns with our visualization analysis: SU-Blocks effectively mitigate spike degradation, releasing more spikes and thereby enhancing feature extraction. Moreover, \cref{tab:5} demonstrates that SU-YOLO, using SU-Blocks, achieves an outstanding mAP$_{0.5}$ of 78.8\%, significantly higher than that obtained using EMS-Blocks or the residual blocks employed in the ANN version.

\begin{figure}
	\centering
	\tabcolsep = 1pt
	\begin{tabular}{cccc}
	\includegraphics[width = 0.24\linewidth]{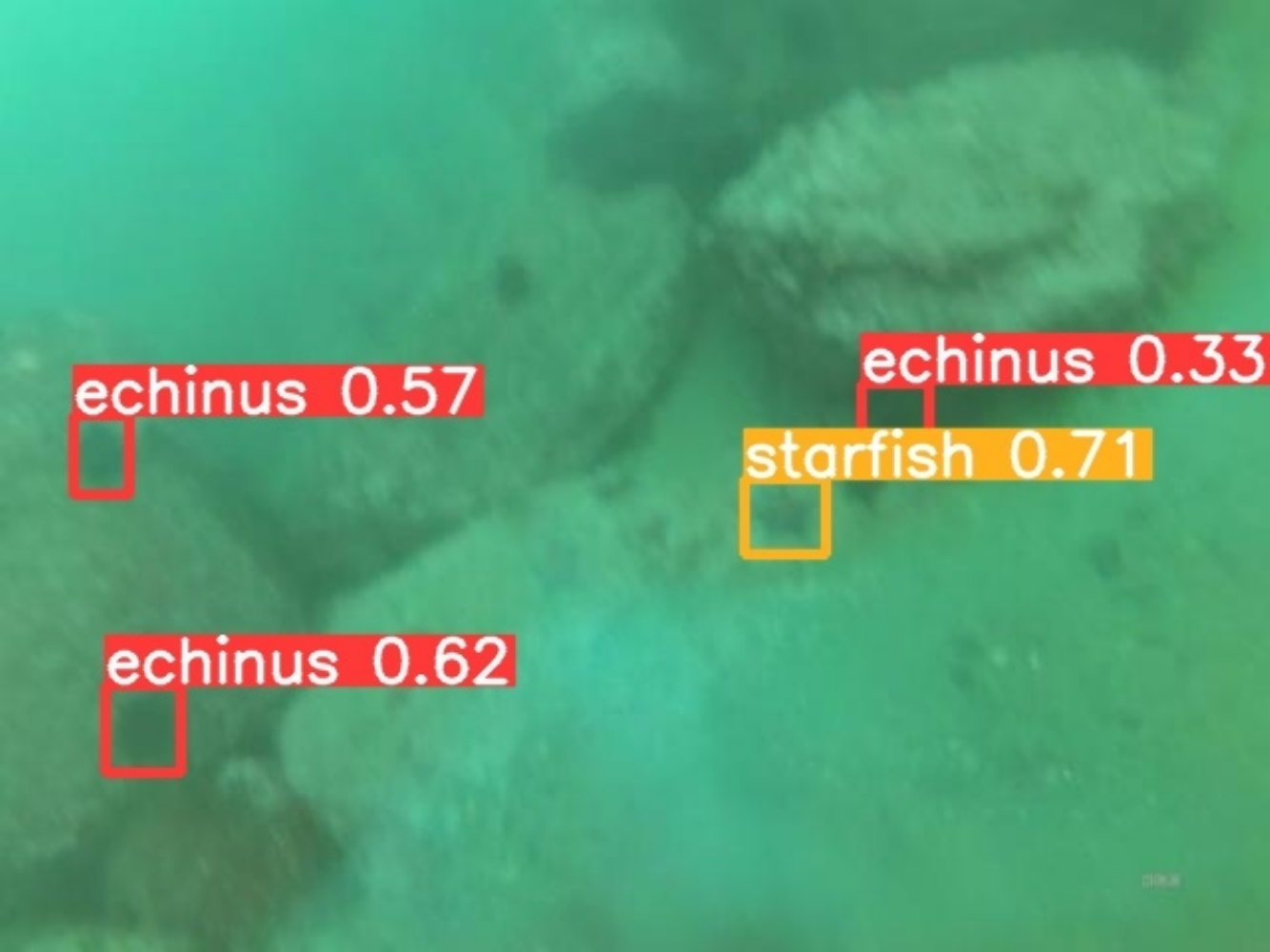} &
	\includegraphics[width = 0.24\linewidth]{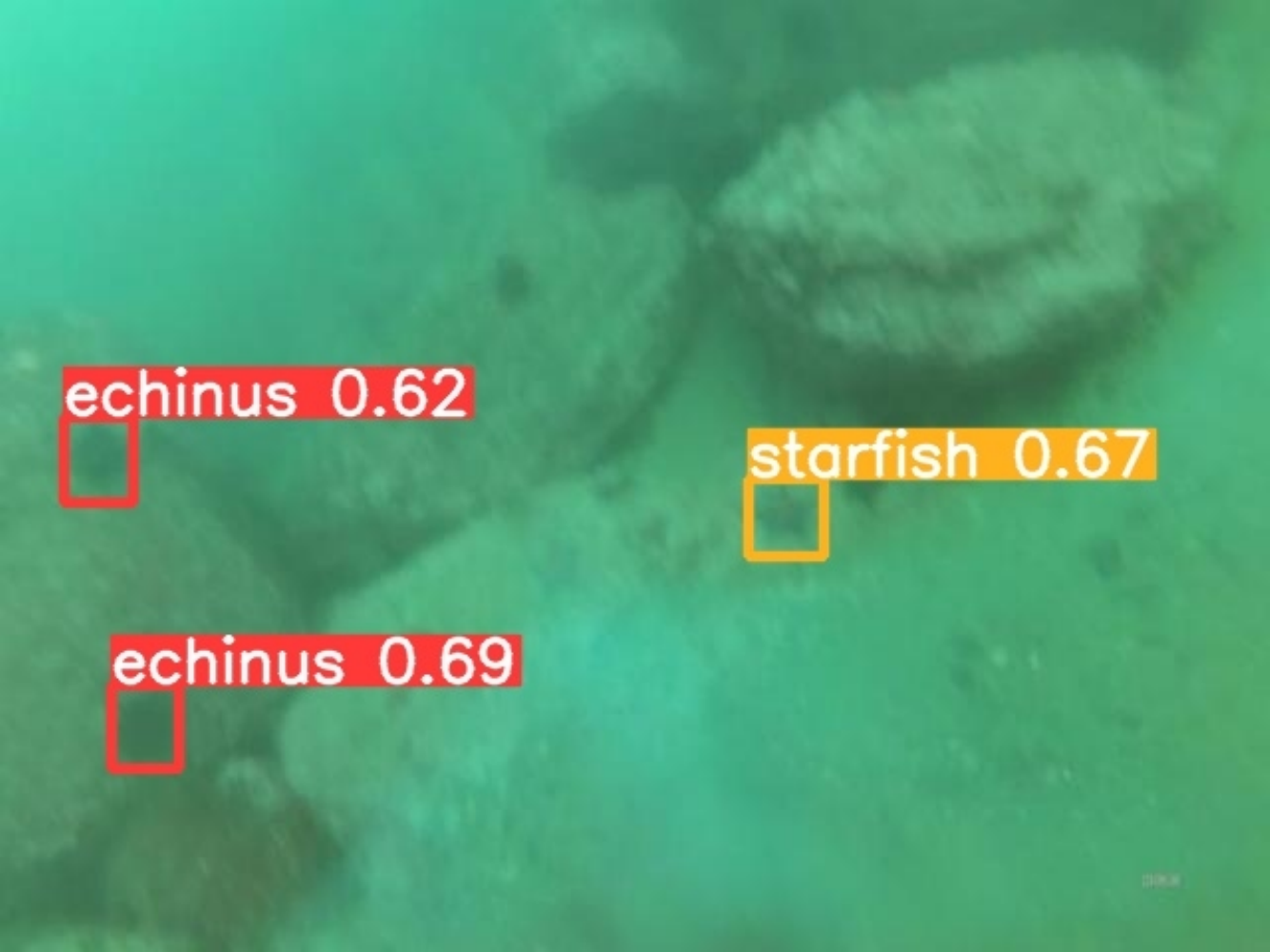} &
	\includegraphics[width = 0.24\linewidth]{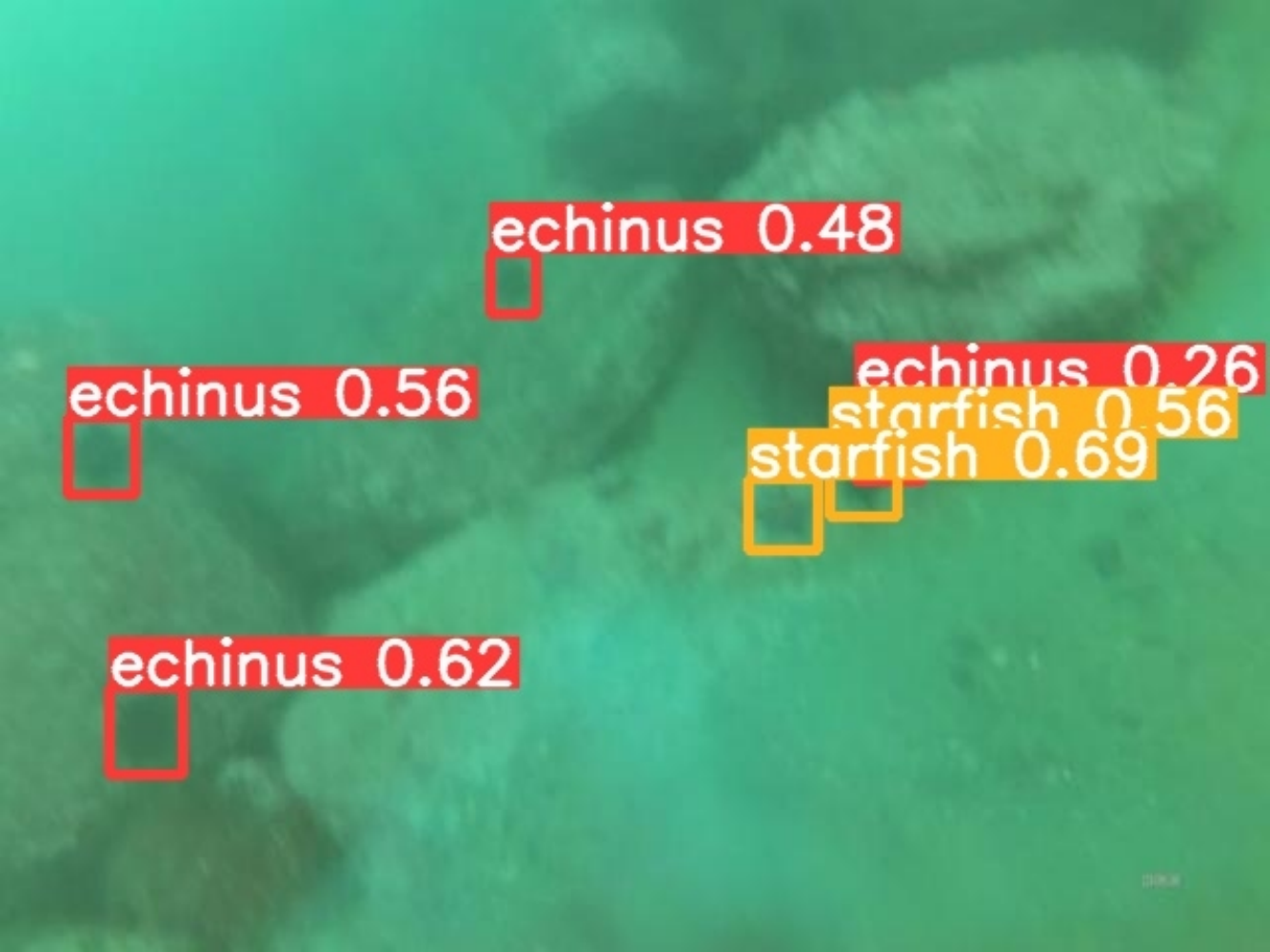} &
	\includegraphics[width = 0.24\linewidth]{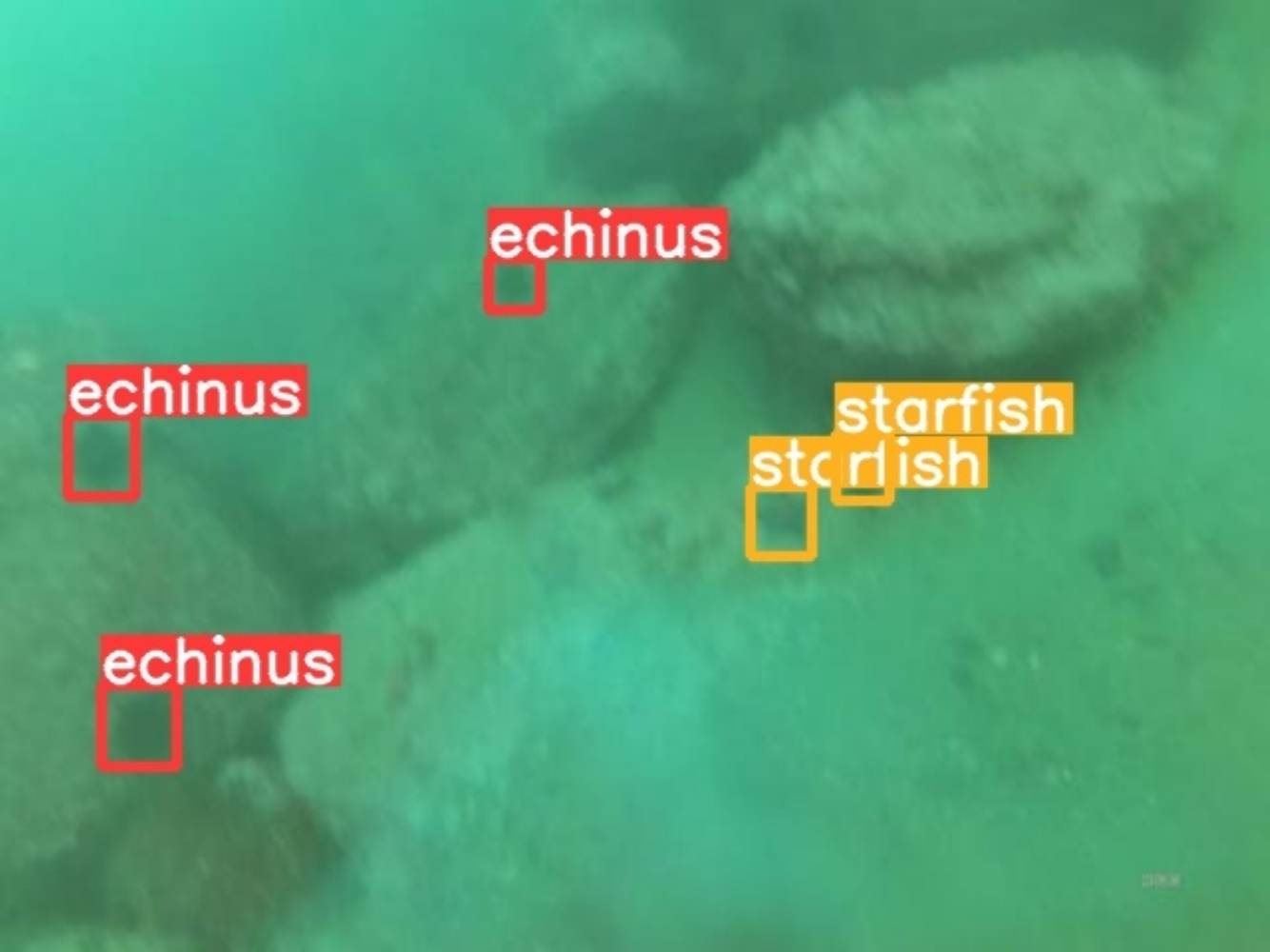} 
	\vspace{-2pt} \\
	\includegraphics[width = 0.24\linewidth]{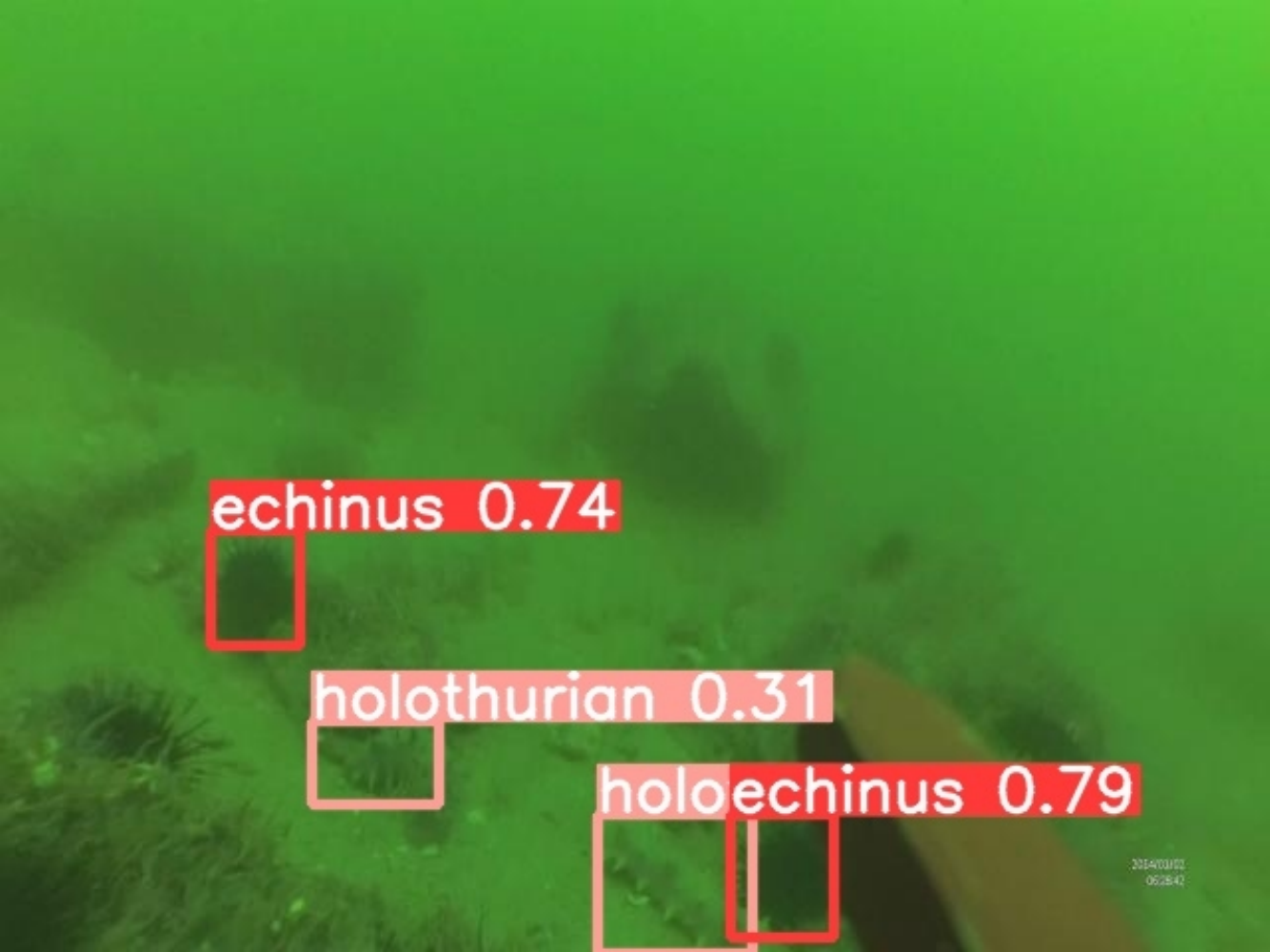} &
	\includegraphics[width = 0.24\linewidth]{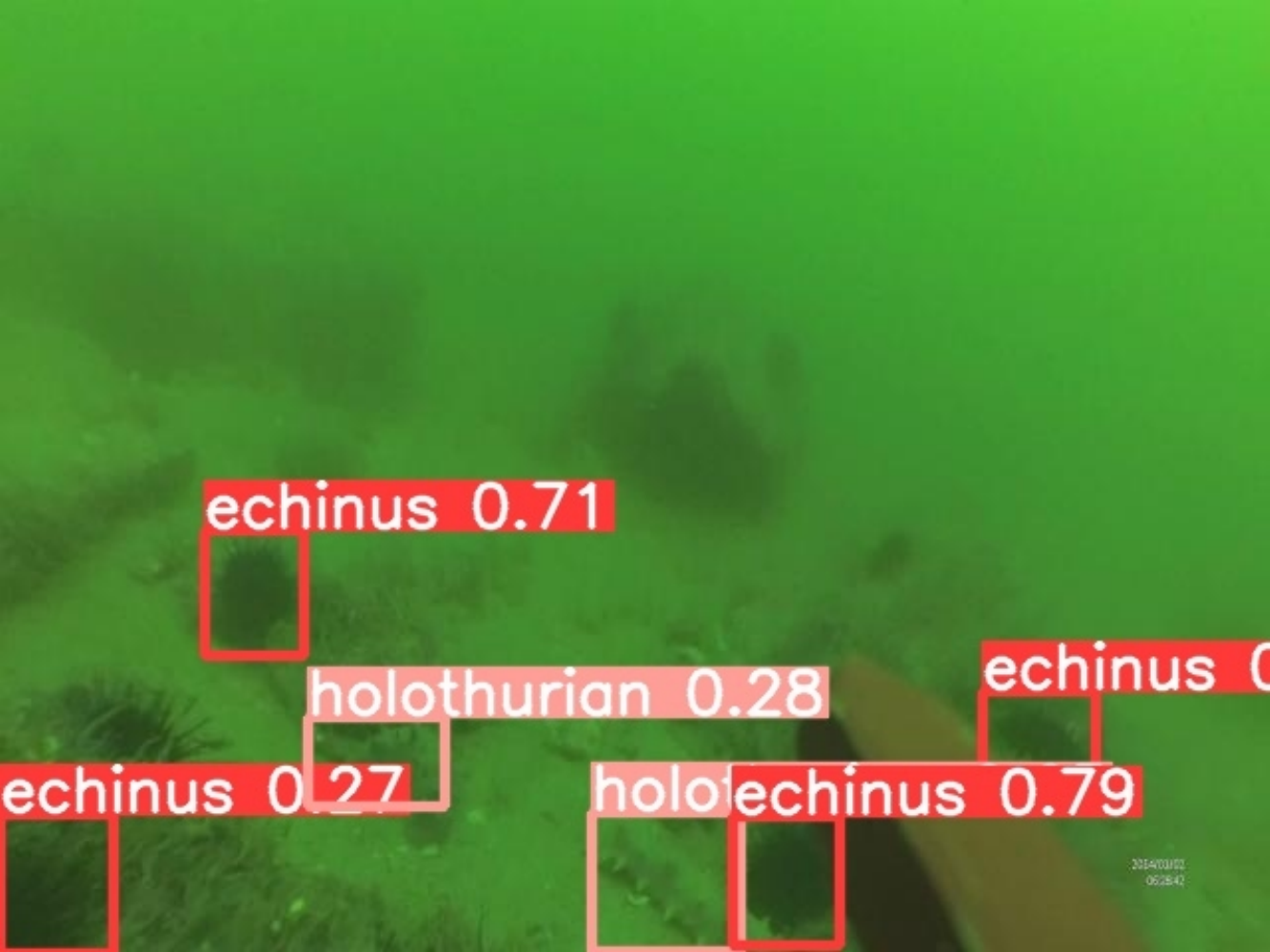} &
	\includegraphics[width = 0.24\linewidth]{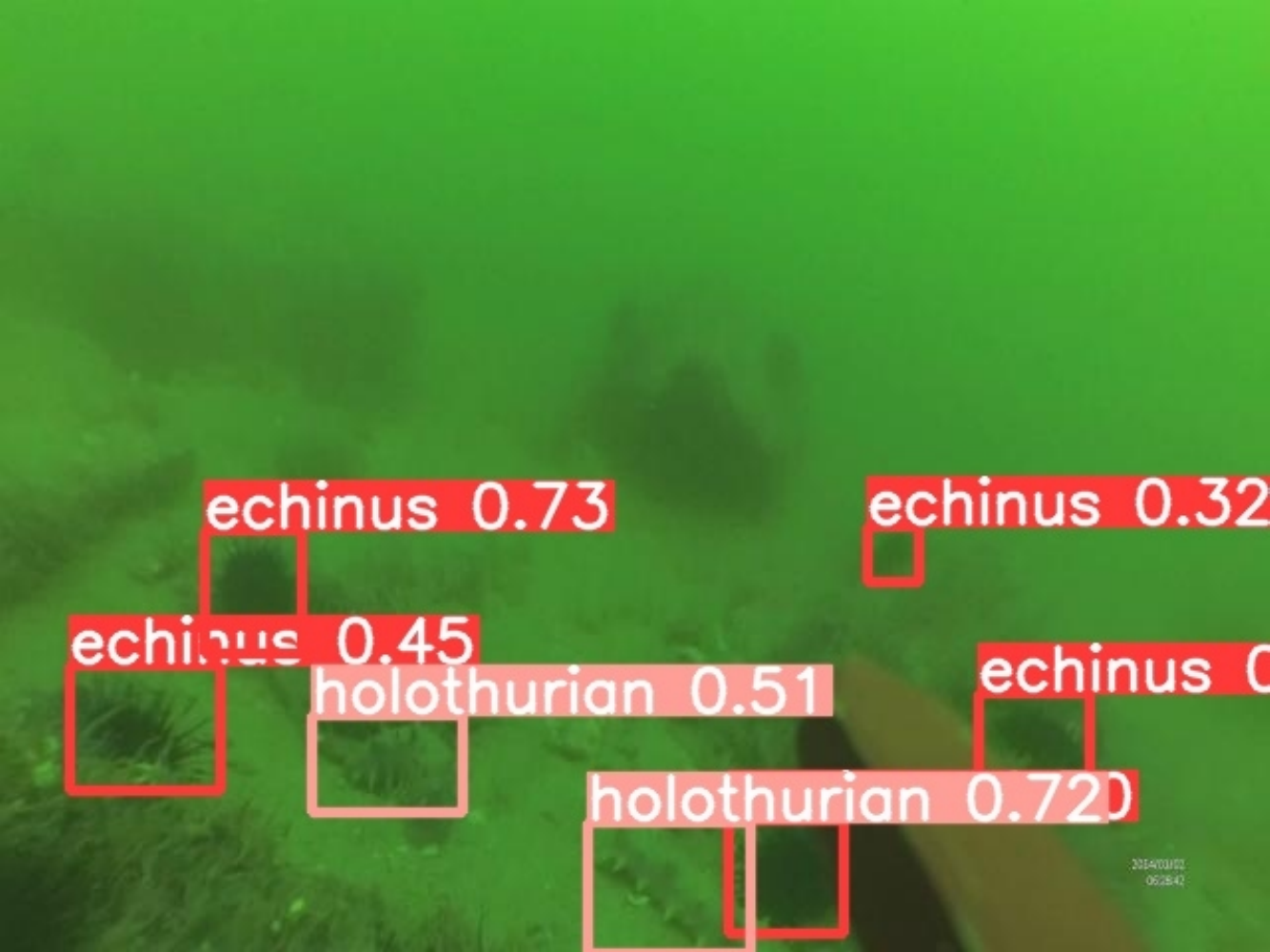} &
	\includegraphics[width = 0.24\linewidth]{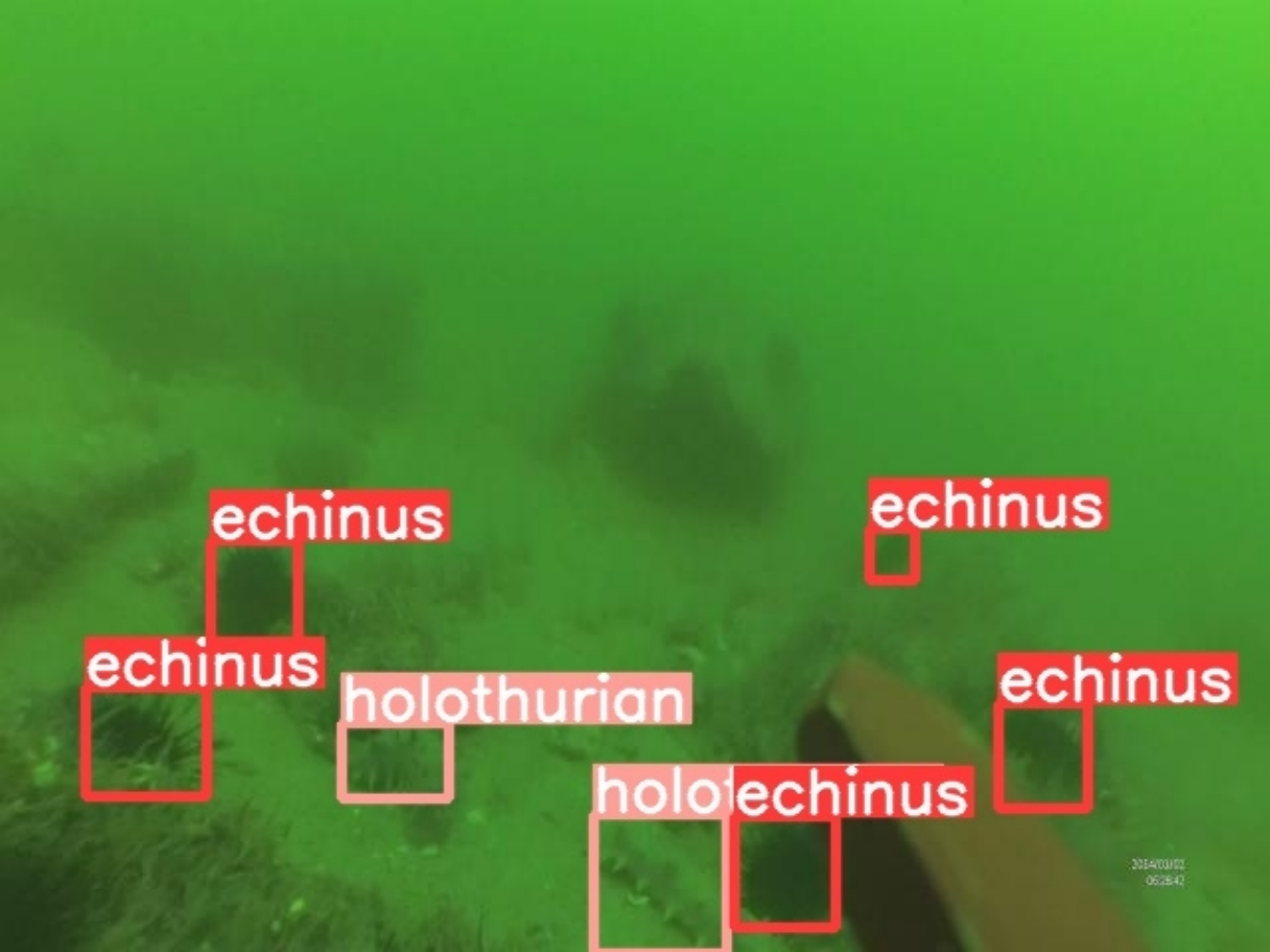} 
	\vspace{-2pt} \\
	\includegraphics[width = 0.24\linewidth]{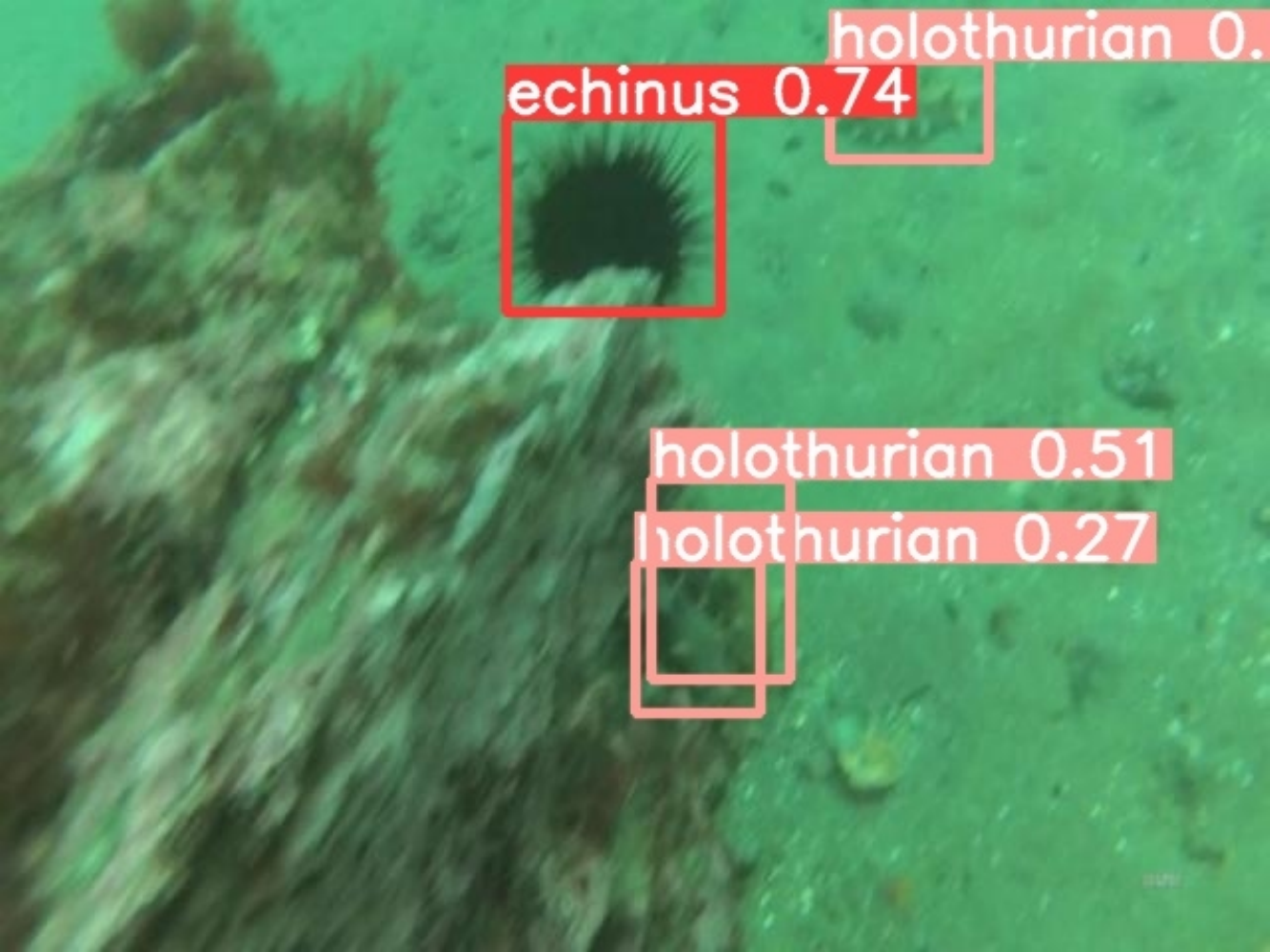} &
	\includegraphics[width = 0.24\linewidth]{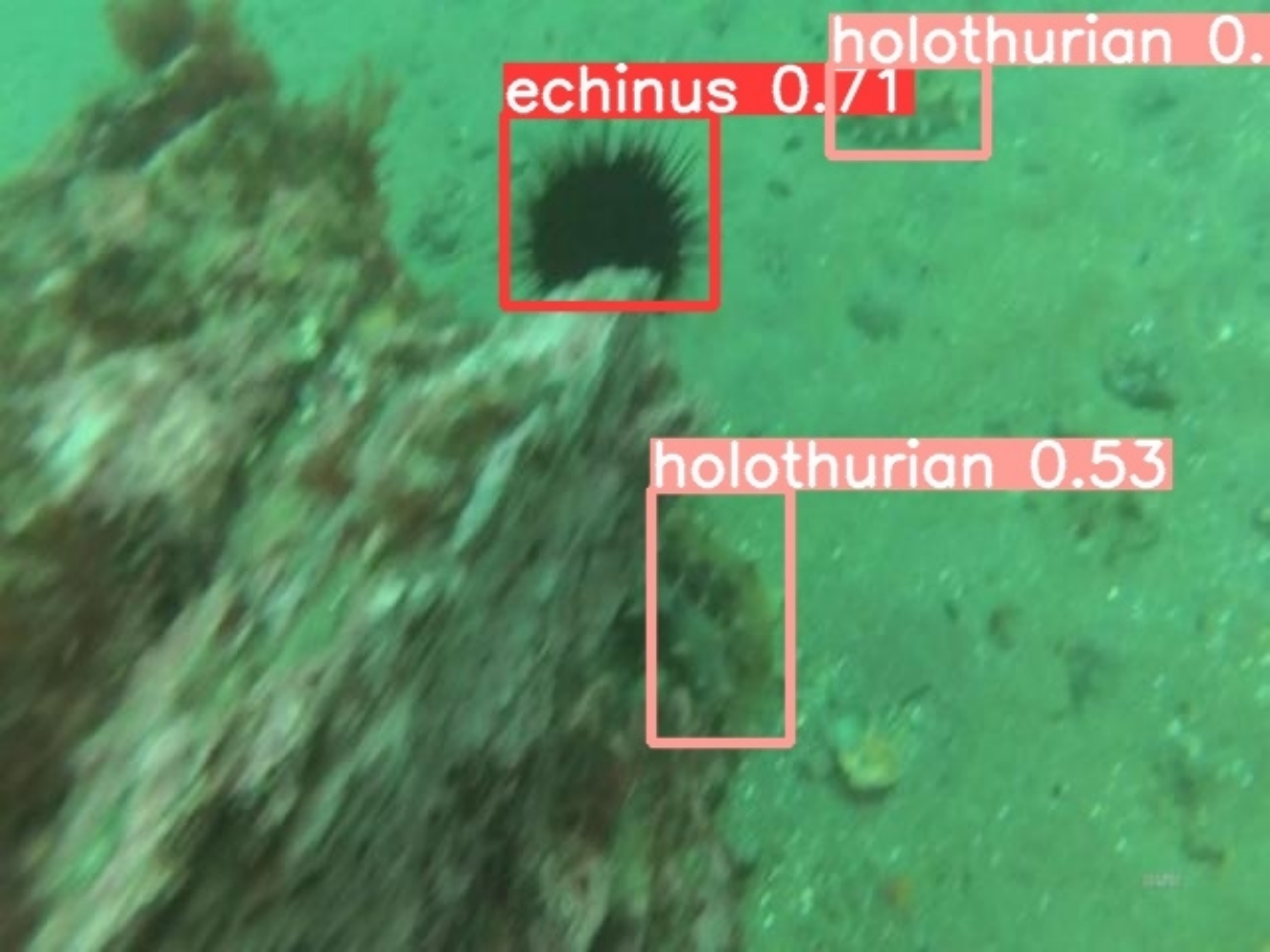} &
	\includegraphics[width = 0.24\linewidth]{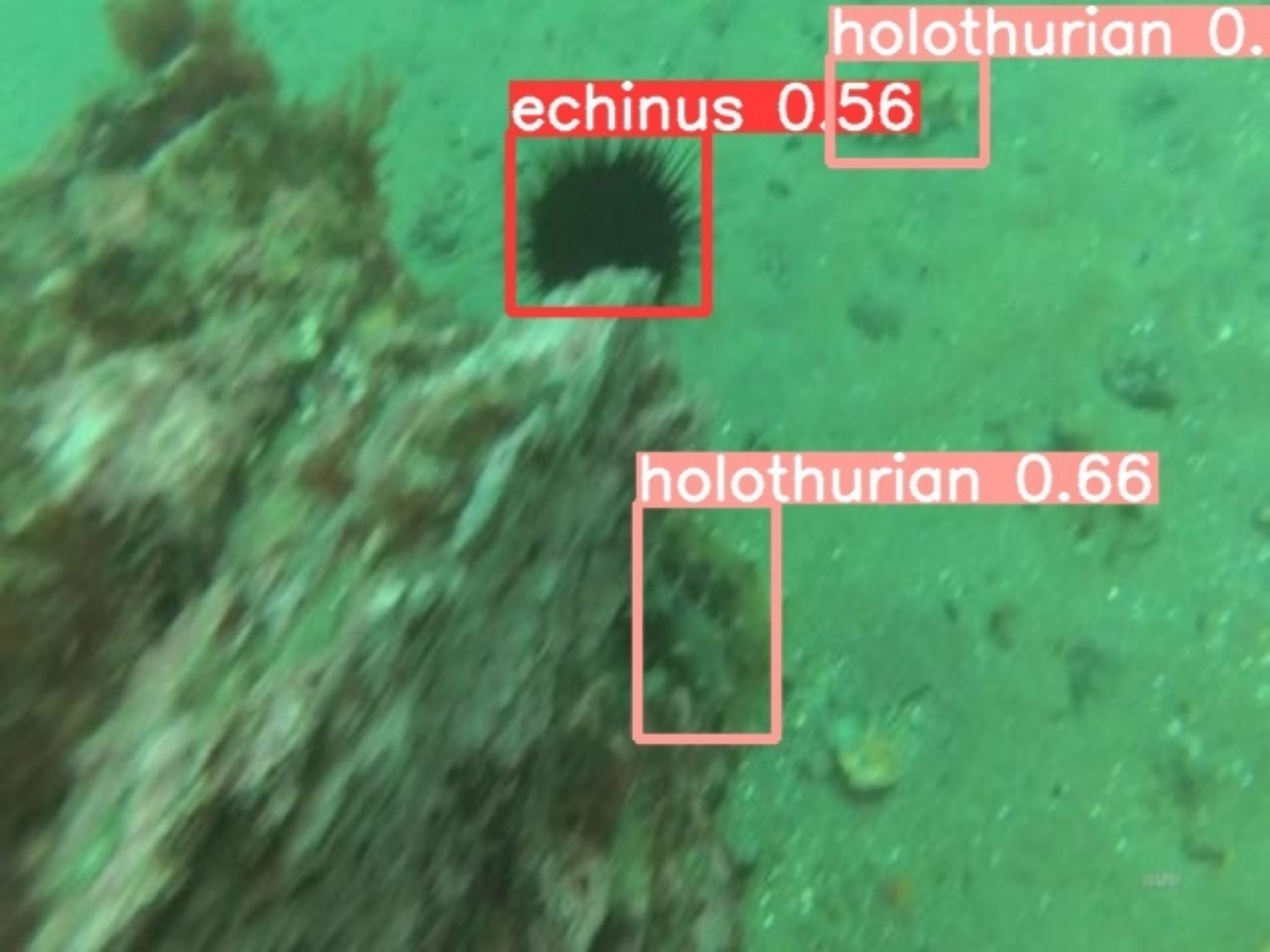} &
	\includegraphics[width = 0.24\linewidth]{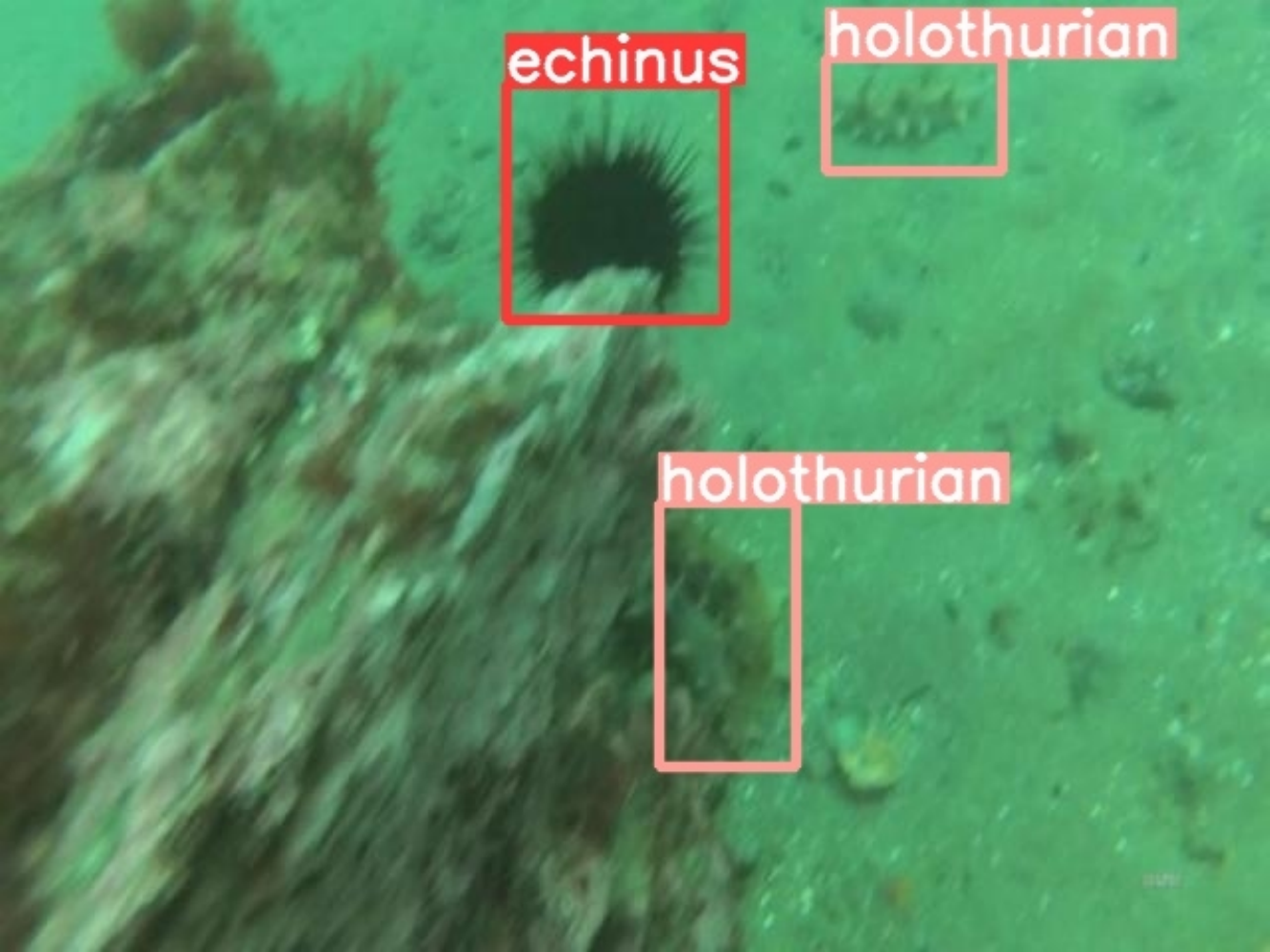} 
	\vspace{-2pt} \\
	\footnotesize{(a) ANN-Res18} &
	\footnotesize{(b) EMS-Res18} &
	\footnotesize{(b) SU-Block} &
	\footnotesize{(d) Ground Truth}
	\end{tabular}
	\caption{\textbf{Detection results using various residual blocks on \textsc{URPC2019}.} Ground Truth shows the original dataset labels.}
	\label{Fig:11}
\end{figure} 

\cref{Fig:11} presents example detection results, highlighting that the SU-Block model detects more objects. These results suggest that SU-Block can match or even surpass the performance of ANN residual blocks of the same scale in underwater object detection. Combined with the low-power consumption characteristics of SNNs, a fully SNN structure may offer advantages over hybrid ANN-SNN architectures.

\begin{figure}
	\centering
	\begin{tabular}{ccc}
	\includegraphics[width = 0.3\linewidth]{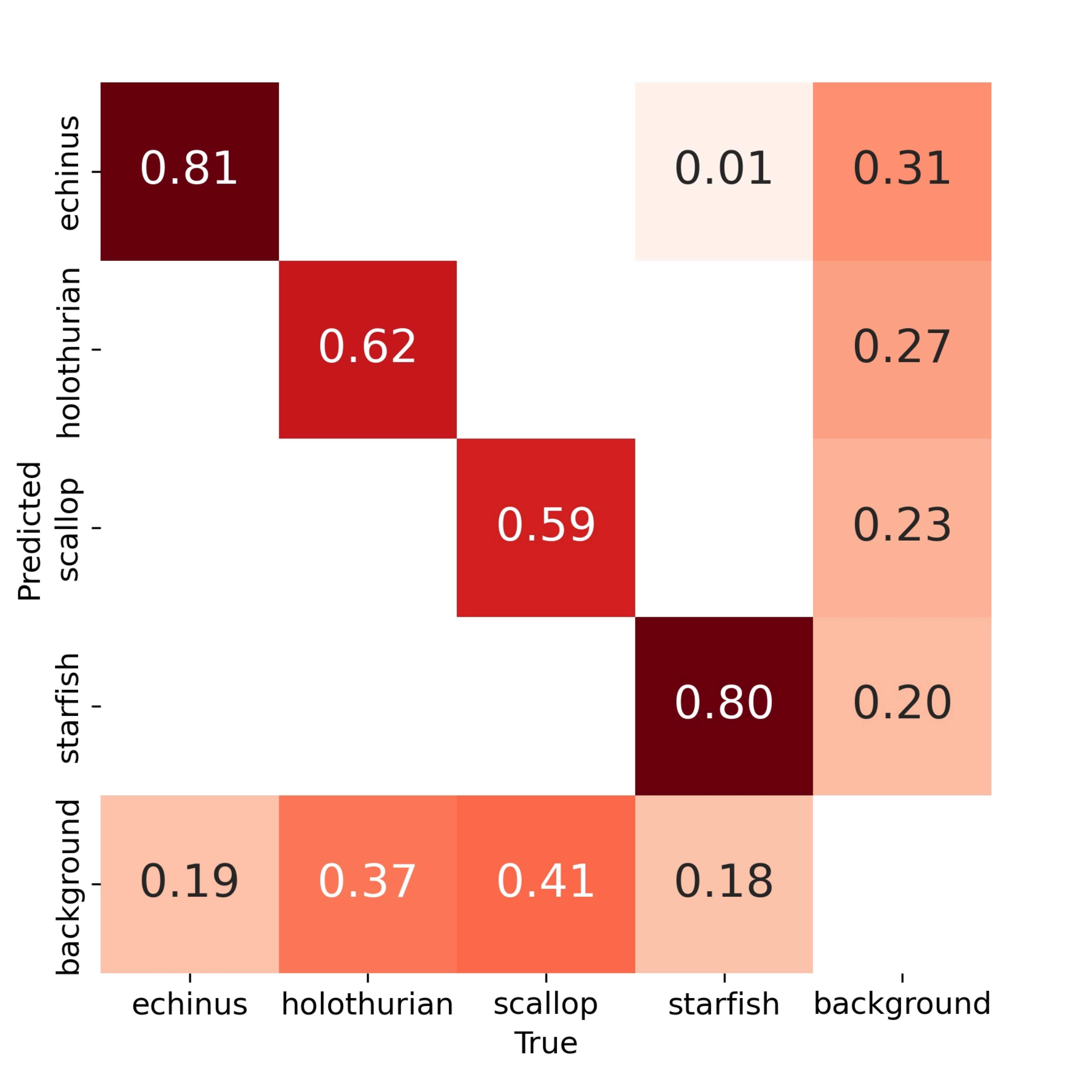} &
	\includegraphics[width = 0.3\linewidth]{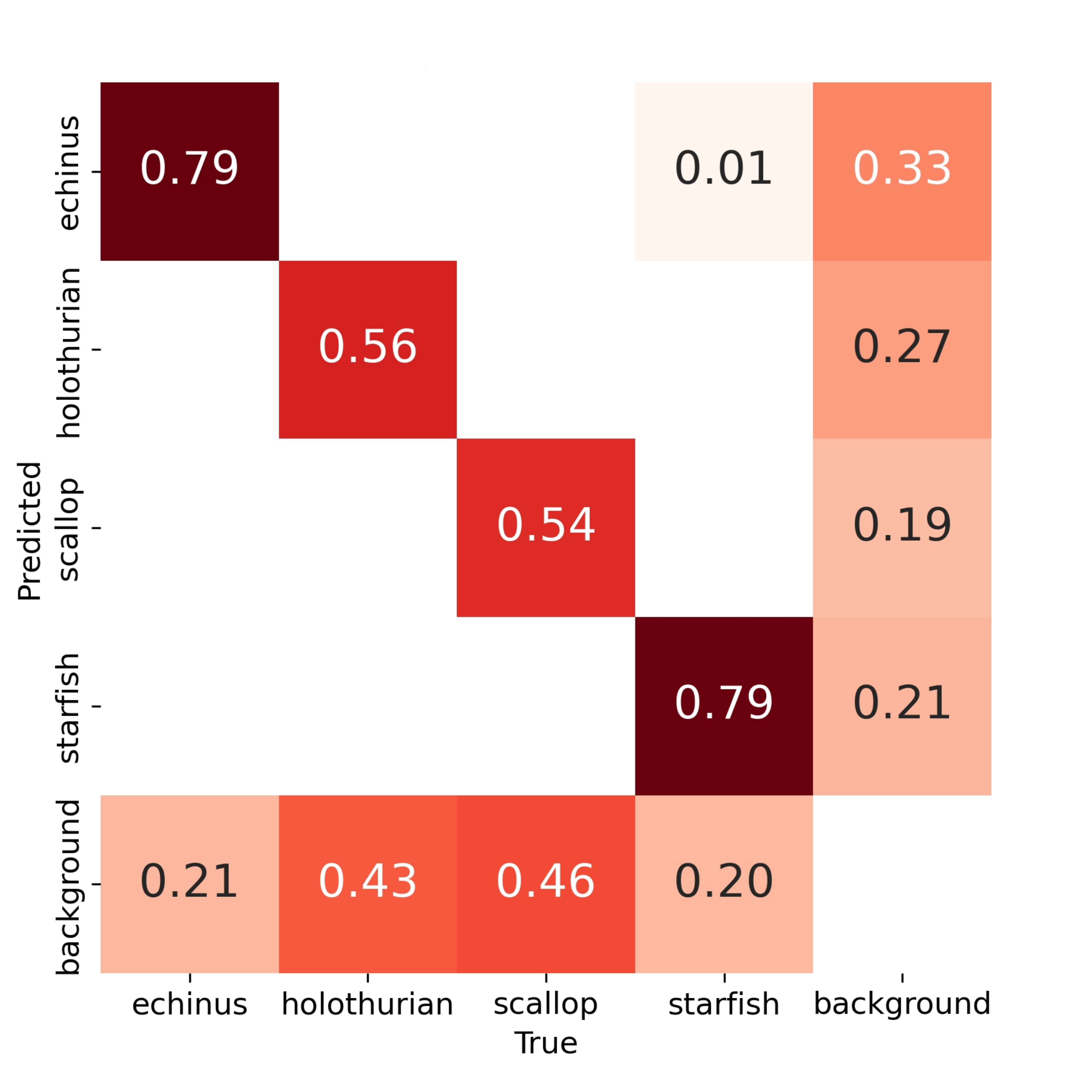} &
	\includegraphics[width = 0.3\linewidth]{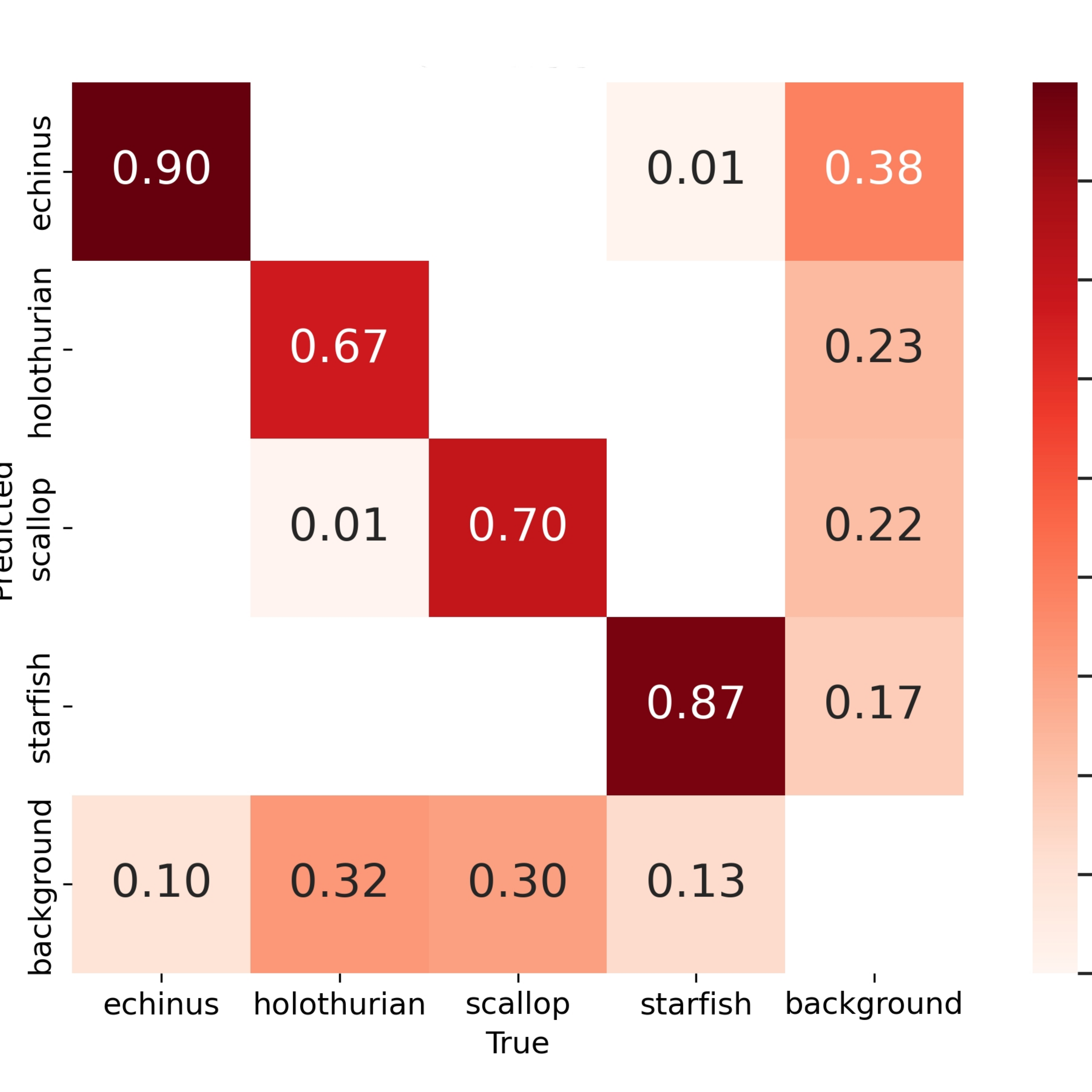} \\
	\footnotesize{(a) ANN-Res18} &
	\footnotesize{(b) EMS-Res18} &
	\footnotesize{(c) SU-Block}
	\end{tabular}
	\caption{\textbf{Confusion matrices for models employing different residual blocks on \textsc{URPC2019}.} The horizontal axis represents actual objects, and the vertical axis represents predicted objects of the model.}
	\label{Fig:12}
\end{figure} 

\subsubsection{Confusion Matrix and Precision-Recall Curve}

The confusion matrix in \cref{Fig:12} illustrates the model's classification performance. For starfish and echinus, all three models achieve high accuracy; however, holothurian and scallop detections exhibit lower accuracy, likely because their lighter colors blend with the water background, making them harder to distinguish. Notably, the SU-Block model demonstrates the highest accuracy across all object classes, particularly outperforming ANN-Res18 and EMS-Res18 by 12\% and 16\%, respectively, in scallop detection. This indicates that SU-Block has superior feature extraction capabilities, enabling it to detect less conspicuous objects.

\begin{figure}
	\centering
	\begin{tabular}{ccc}
	\includegraphics[width = 0.3\linewidth]{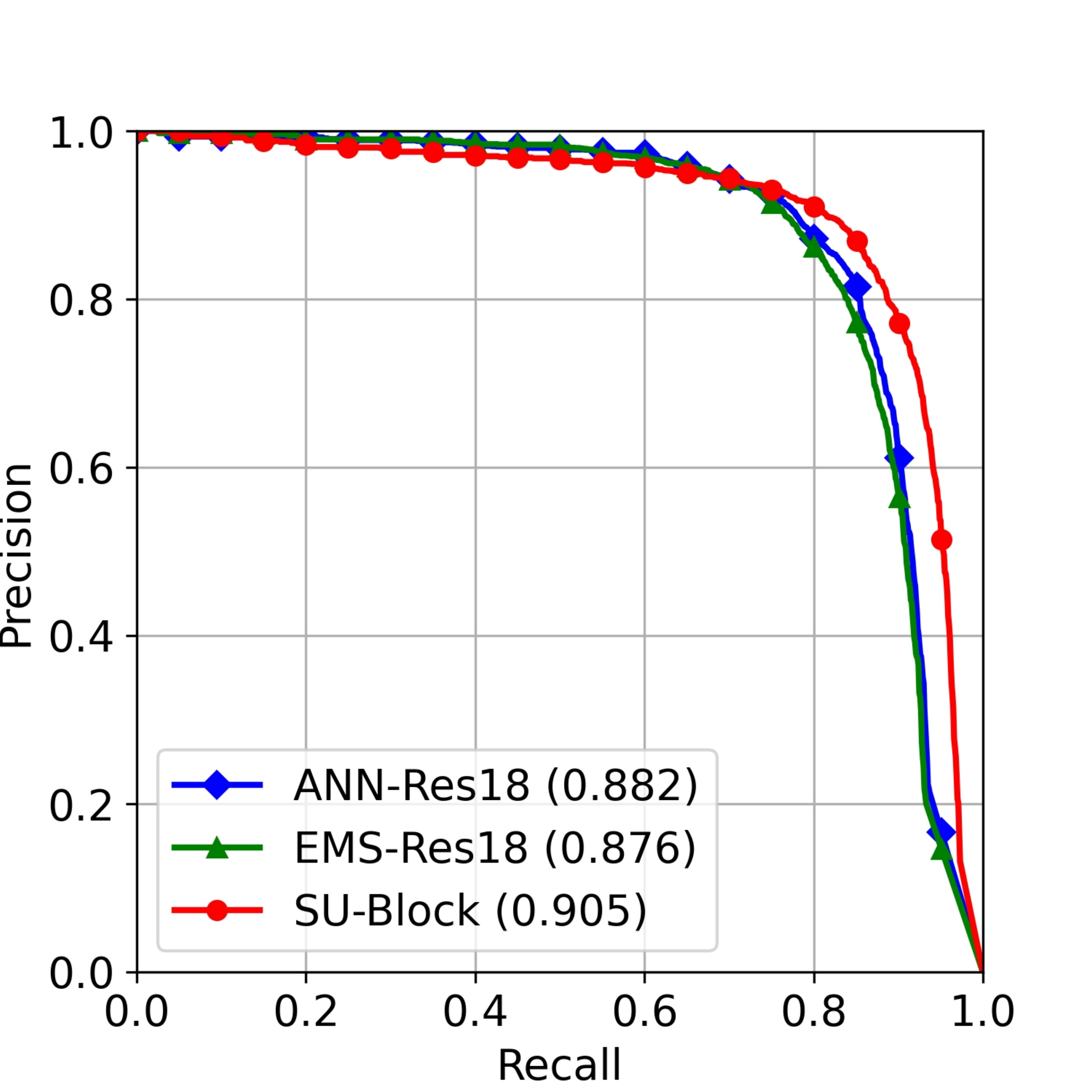} &
	\includegraphics[width = 0.3\linewidth]{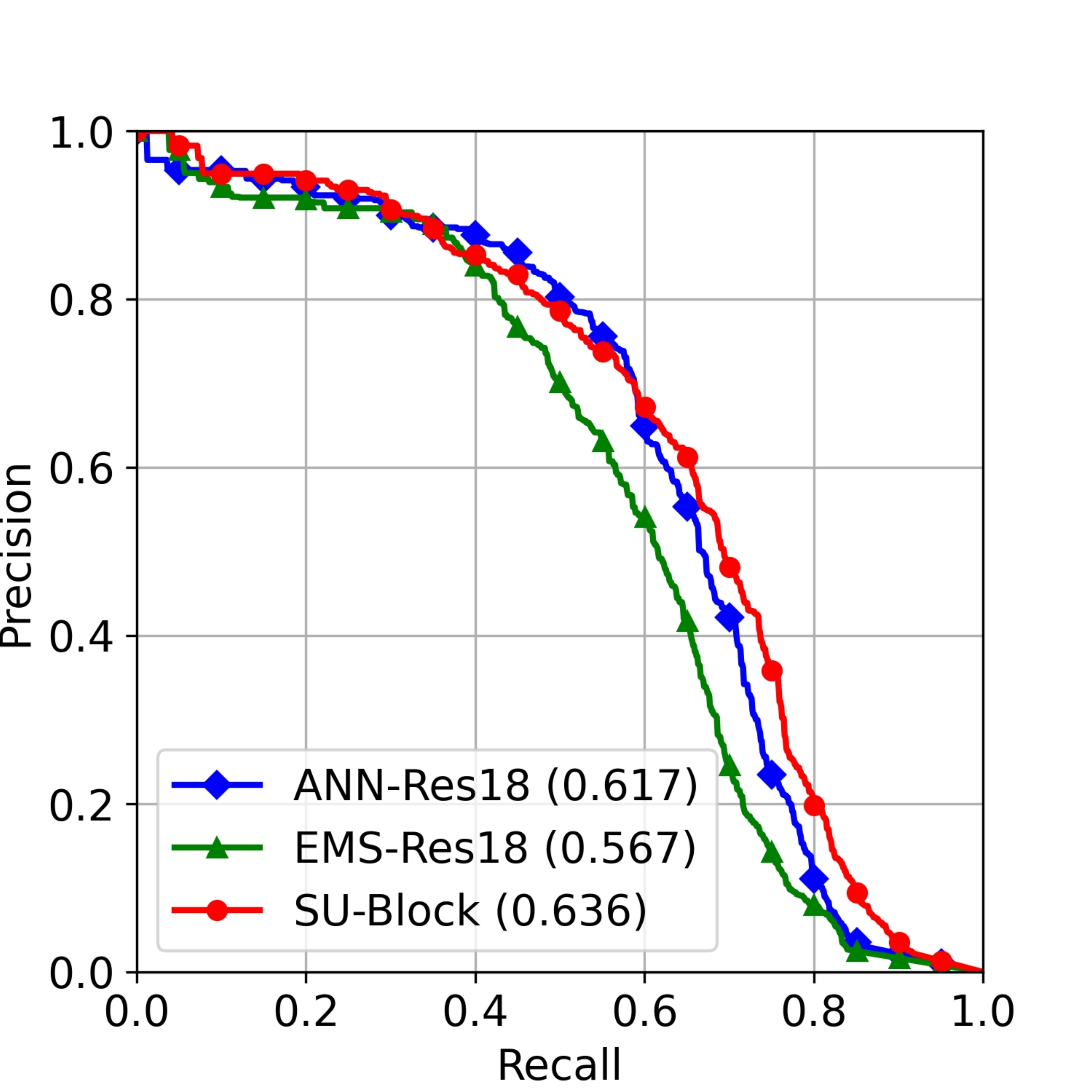} &
	\includegraphics[width = 0.3\linewidth]{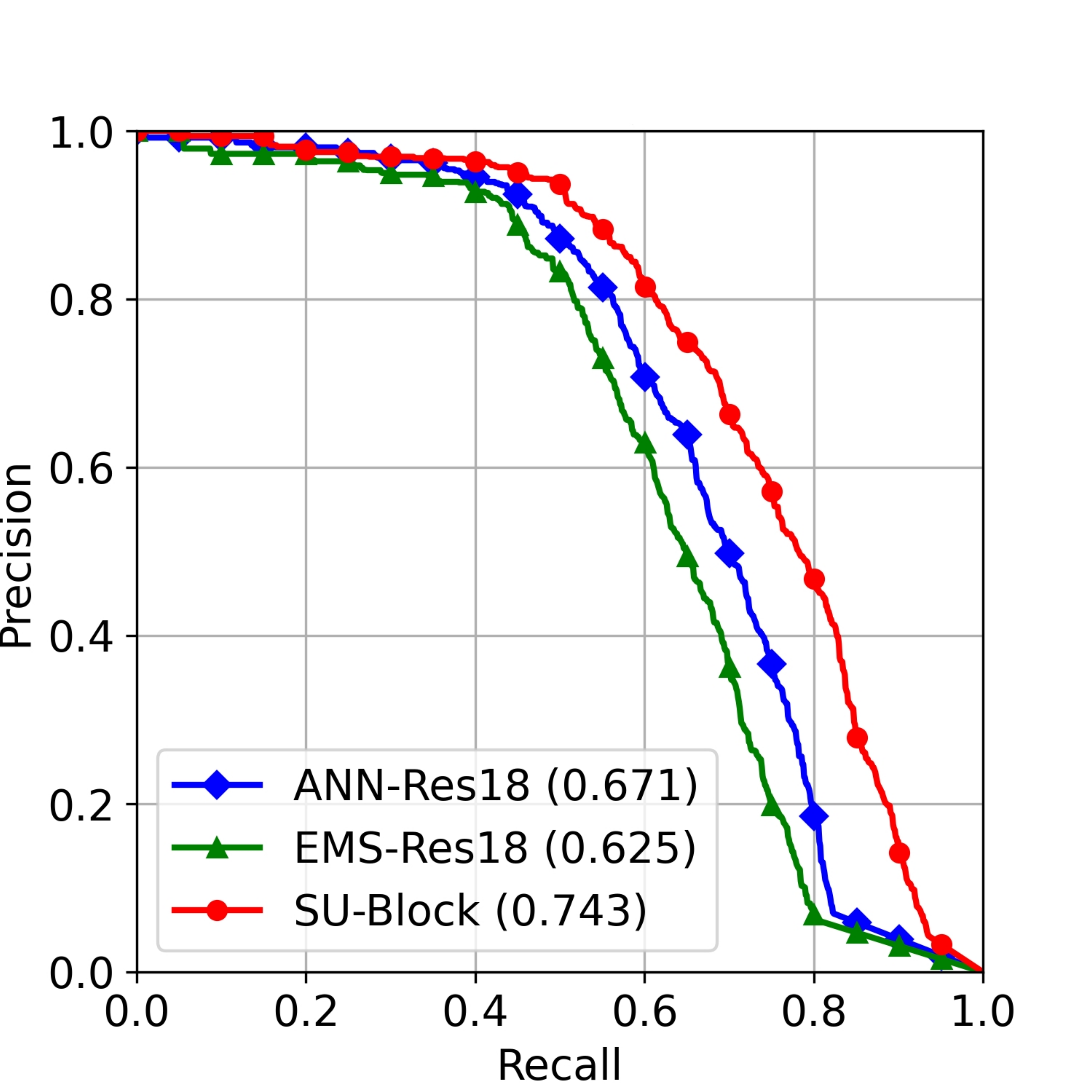} \\
	\footnotesize{(a) echinus} &
	\footnotesize{(b) holothurain} &
	\footnotesize{(c) scallop}
	\end{tabular}
	\begin{tabular}{cc}
	\includegraphics[width = 0.3\linewidth]{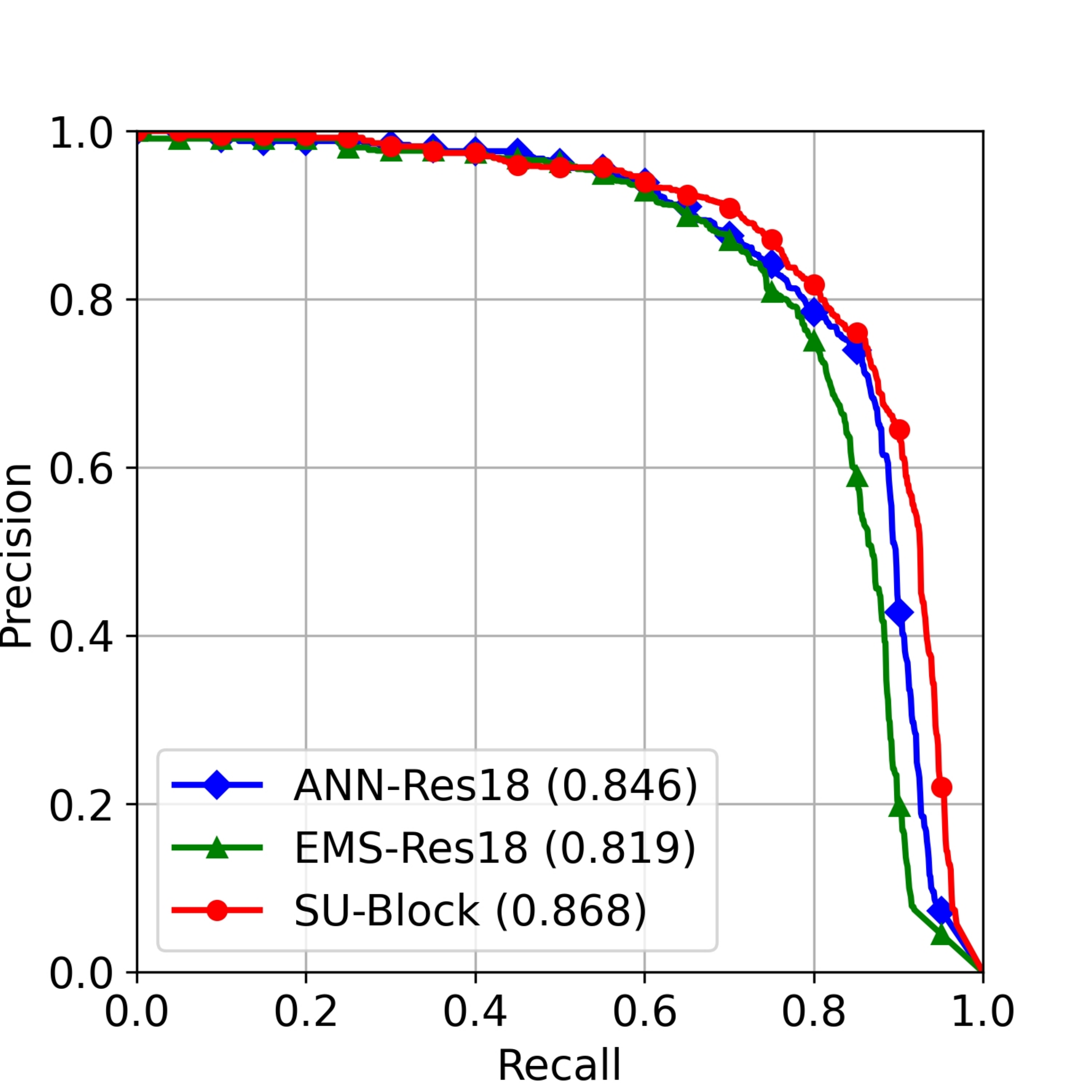} &
	\includegraphics[width = 0.3\linewidth]{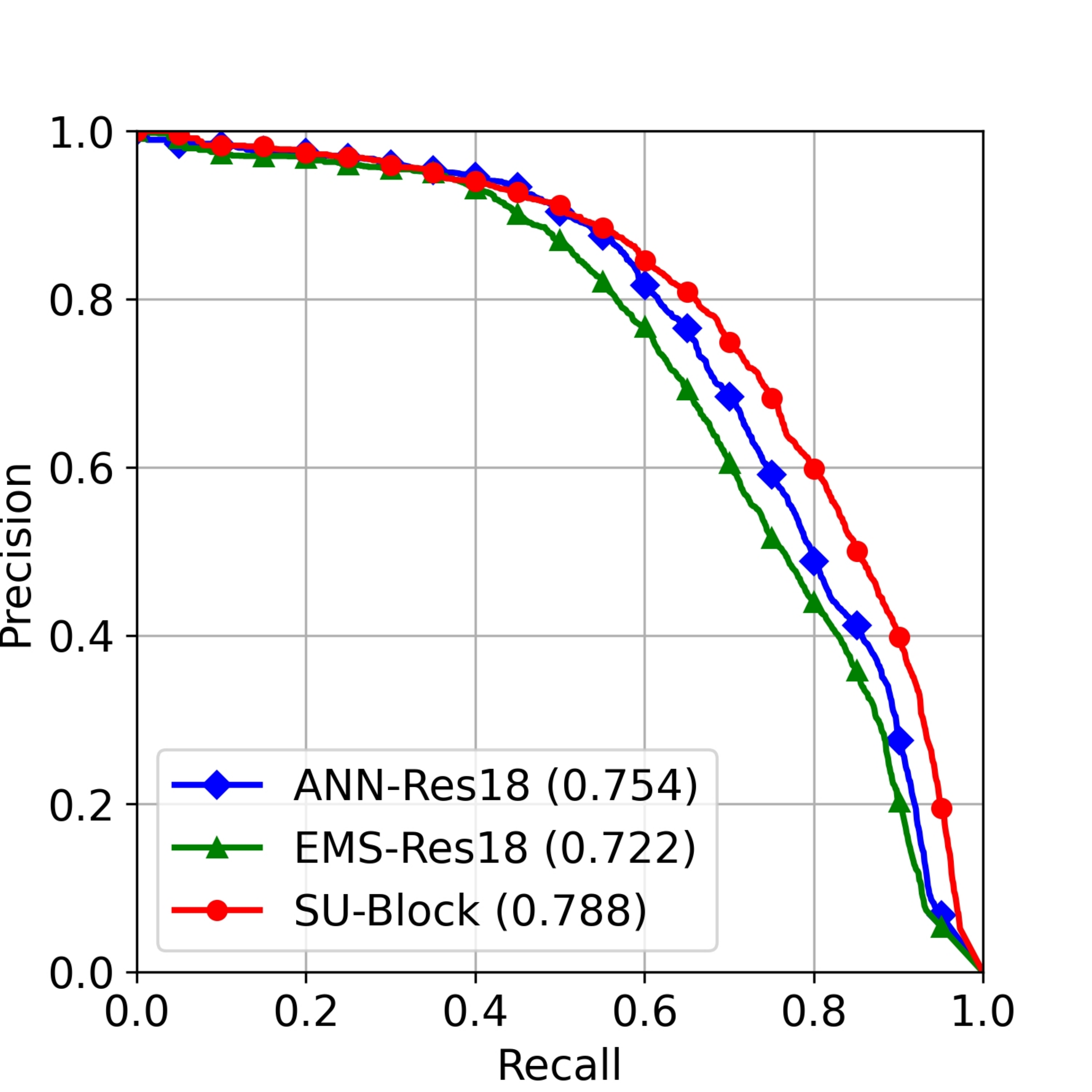} \\
	\footnotesize{(d) starfish} & 
	\footnotesize{(e) overall}
	\end{tabular}
	\caption{\textbf{Precision-Recall curves for models employing different residual blocks on \textsc{URPC2019}.} The values in the legends represent mAP$_{0.5}$.}
	\label{Fig:13}
\end{figure} 

The Precision-Recall curves in \cref{Fig:13} further highlight the superior performance of the SU-Block model, especially in scallop detection, with an overall mAP$_{0.5}$ of 78.8\%, compared to 75.4\% for ANN-Res18 and 72.2\% for EMS-Block.

\begin{figure}
	\centering
	\begin{tabular}{cc}
 	\setlength{\tabcolsep}{10pt}
	\includegraphics[width = 0.4\linewidth]{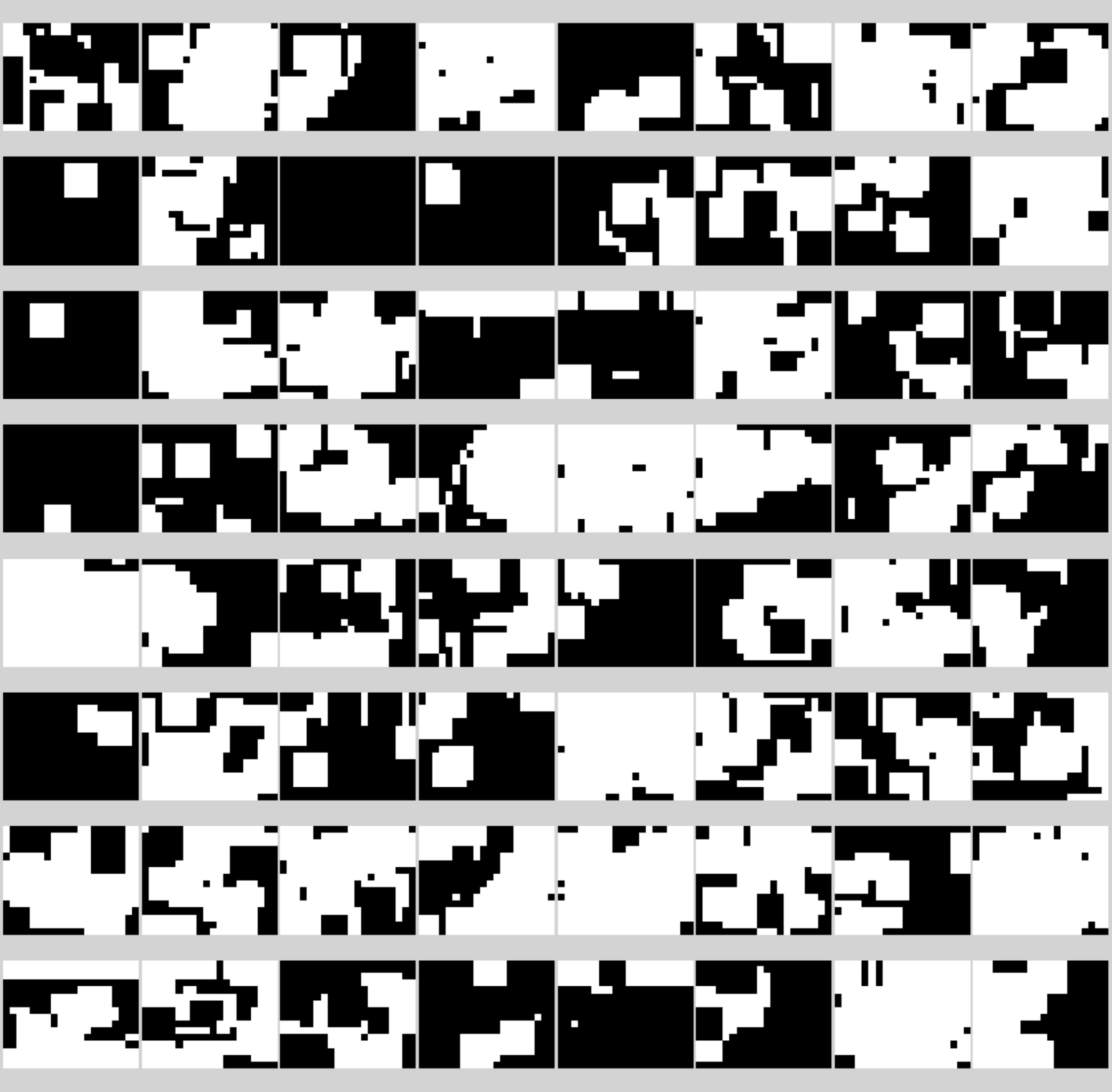} &
	\includegraphics[width = 0.4\linewidth]{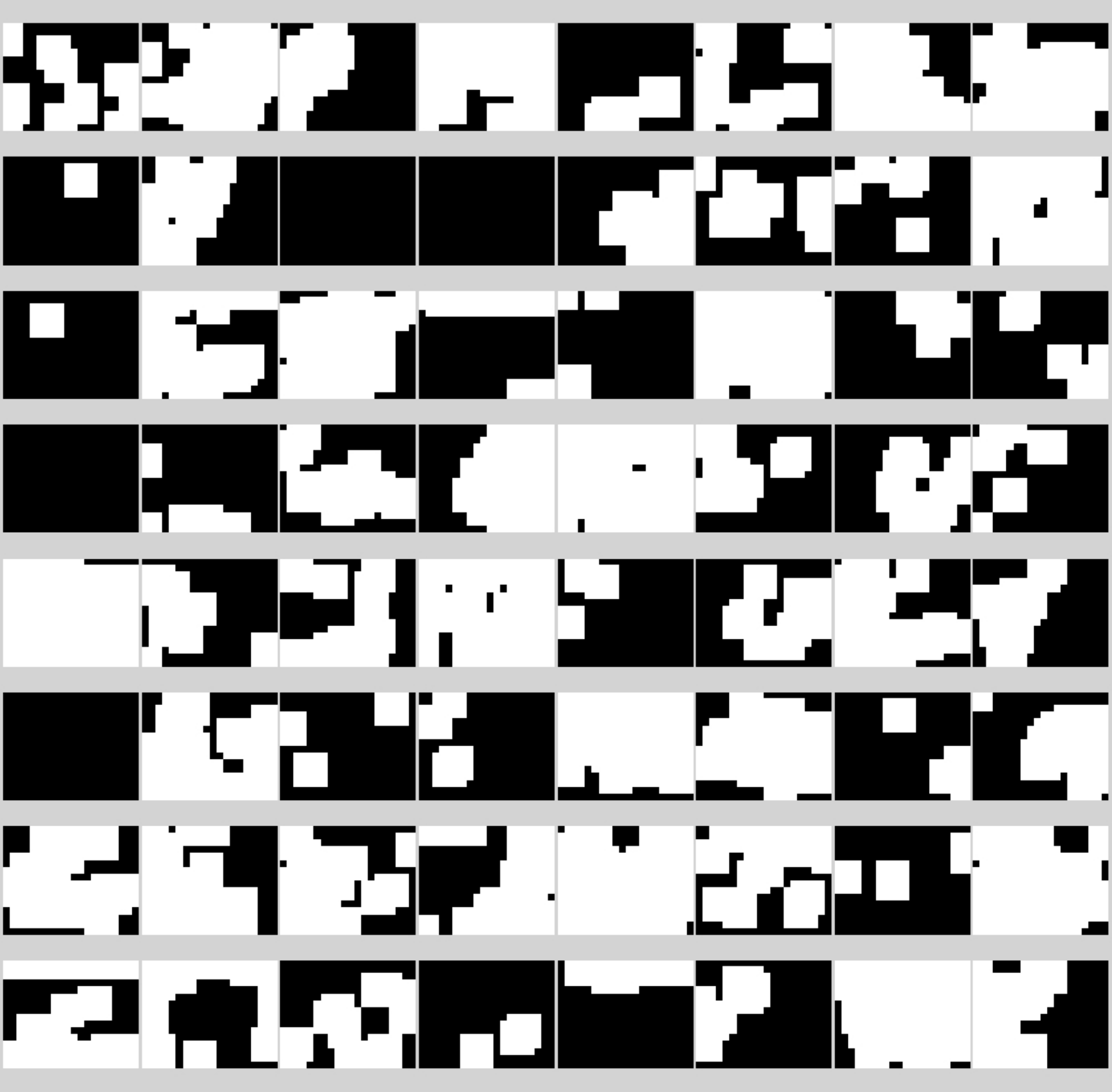} \\
	\footnotesize{(a) Conv-Maxpool-IF} &
	\footnotesize{(b) Conv-IF-Maxpool}
	\end{tabular}
	\caption{\textbf{Pooled spike feature maps at time step $T = 1$, where black indicates a value of 0 and white indicates a value of 1.}}
	\label{Fig:14}
\end{figure} 

\subsubsection{Sequence of Pooling and Activation}

The sequence of max-pooling and activation operations significantly affects model performance in SNN-based object detection. We conducted experiments comparing two scenarios. As shown in \cref{Fig:14}, when pooling is applied before activation in the SpikeSPP, the resulting feature maps preserve more detail. 

\begin{table}
	\centering
 	\setlength{\tabcolsep}{10pt}
	\scriptsize
	\begin{tabular}{l|cc|cc}
	\toprule[1.1pt]
	\multirow{2}[2]{*}{Sequence} & \multicolumn{2}{c|}{mAP$_{0.5}$} & \multicolumn{2}{c}{mAP$_{0.5:0.95}$} \\
	\cmidrule(lr){2-3} \cmidrule(lr){4-5}
	& \textsc{URPC} & \textsc{UDD} & \textsc{URPC} & \textsc{UDD} \\
	\midrule
	Conv-IF-MaxPool & 0.775 & 0.548 & 0.416 & 0.253 \\
	\rowcolor{gray!20}
	Conv-MaxPool-IF & \textbf{0.788} & \textbf{0.582} & \textbf{0.429} & \textbf{0.266} \\
	\bottomrule[1.1pt]
	\end{tabular}
	\caption{\textbf{Effects of two different sequences on detection accuracy in the SpikeSPP module of SU-YOLO.}}
	\label{tab:6}
\end{table}

This observation is corroborated by the quantitative results in \cref{tab:6}: placing the pooling layer before the IF neurons improves mAP$_{0.5}$ by 1.3\% on \textsc{URPC2019} and by 3.4\% on \textsc{UDD}, thereby confirming our analysis.

\begin{table}
	\centering
	\setlength{\tabcolsep}{10pt}
	\scriptsize
	\begin{tabular}{cc|cccccc}
	\toprule[1.1pt]
	\multicolumn{2}{c|}{Time Step $T$} & 1 & 2 & 3 & \cellcolor{gray!20}4 & 5 & 6 \\
	\midrule
	\multirow{2}{*}{\textsc{URPC}} & mAP$_{0.5}$ & 0.700 & 0.757 & 0.776 & \cellcolor{gray!20}\textbf{0.788} & 0.787 & 0.742 \\
	& mAP$_{0.5:0.95}$ & 0.358 & 0.401 & 0.416 & \cellcolor{gray!20}\textbf{0.429} & 0.428 & 0.406 \\
	\midrule
	\multirow{2}{*}{\textsc{UDD}} & mAP$_{0.5}$ & 0.488 & 0.532 & 0.537 & \cellcolor{gray!20}\textbf{0.582} & 0.578 & 0.567 \\
	& mAP$_{0.5:0.95}$ & 0.219 & 0.238 & 0.250 & \cellcolor{gray!20}0.266 & \textbf{0.277} & 0.257 \\
	\bottomrule[1.1pt]
	\end{tabular}
	\caption{\textbf{mAP of the model as the time step $T$ varies from 1 to 6.}}
	\label{tab:7}
\end{table}

\subsubsection{Length of Time Steps}

The choice of time steps in SNNs directly impacts both performance and computational load. As presented in \cref{tab:7}, SU-YOLO achieves optimal accuracy when the number of time steps, $T$, is set to 4. Increasing $T$ beyond 4 does not further improve performance and may slightly reduce it. Therefore, we adopt 4 time steps to achieve the best balance between detection accuracy and power consumption.

\section{Conclusion}

In this study, we proposed SU-YOLO, a lightweight SNN model for underwater object detection designed for resource-constrained platforms. To address the challenge of noise in underwater images, we developed a spike-based image denoising method that proved highly effective in our experiments. Additionally, we introduced SeBN, a novel batch normalization method tailored to the unique characteristics of SNNs, which outperformed existing normalization techniques. The carefully designed modules within SU-YOLO significantly enhance detection capabilities in underwater environments, and experimental results demonstrate that our model achieves state-of-the-art performance among current SNN approaches for underwater object detection.

However, our conclusions regarding energy efficiency remain theoretical, as they are based on estimates rather than actual hardware measurements. In future work, we aim to further enhance SU-YOLO’s performance and implement it on physical hardware systems to achieve more reliable detection results. This advancement will bolster the practical applications of SNNs in engineering and contribute to the progress of underwater object detection technology. Furthermore, we plan to explore the generalization of the SU-YOLO architecture for broader object detection tasks by extending its applicability to additional datasets and applications.

\section*{Acknowledgments}
This work was supported in part by the National Natural Science Foundation of China under Grants 62271361, and the Hubei Provincial Key Research and Development Program under Grant 2024BAB039.

\bibliographystyle{elsarticle-num}
\bibliography{SU-YOLO}

\begin{thebibliography}{10}
\expandafter\ifx\csname url\endcsname\relax
  \def\url#1{\texttt{#1}}\fi
\expandafter\ifx\csname urlprefix\endcsname\relax\def\urlprefix{URL }\fi
\expandafter\ifx\csname href\endcsname\relax
  \def\href#1#2{#2} \def\path#1{#1}\fi

\bibitem{Ju2025UnderwaterSN}
Y.~Ju, L.~Li, X.~Zhong, Y.~Rao, Y.~Liu, J.~Dong, A.~C. Kot, Underwater surface normal reconstruction via cross-grained photometric stereo transformer, IEEE J. Ocean. Eng. 50~(1) (2025) 192--203.

\bibitem{Pan2024OptimizationAA}
W.~Pan, J.~Chen, B.~J. Lv, L.~Peng, Optimization and application of improved yolov9s-ui for underwater object detection, Appl. Sci. (2024).

\bibitem{Wang2020YOLONU}
L.~Wang, X.~Ye, H.~Xing, Z.~Wang, P.~Li, Yolo nano underwater: A fast and compact object detector for embedded device, Glob. Oceans (2020) 1--4.

\bibitem{Wang2022ULOAU}
L.~Wang, X.~Ye, S.~Wang, P.~Li, Ulo: An underwater light-weight object detector for edge computing, Mach. (2022).

\bibitem{Maass1996NetworksOS}
W.~Maass, Networks of spiking neurons: The third generation of neural network models, Neural Networks 10~(9) (1997) 1659--1671.

\bibitem{mm/ZhongHLHDY024}
X.~Zhong, S.~Hu, W.~Liu, W.~Huang, J.~Ding, Z.~Yu, T.~Huang, Towards low-latency event-based visual recognition with hybrid step-wise distillation spiking neural networks, in: Proc. {ACM} Int. Conf. Multimedia, 2024, pp. 9828--9836.

\bibitem{tamd/YouZLWHYH24}
H.~You, X.~Zhong, W.~Liu, Q.~Wei, W.~Huang, Z.~Yu, T.~Huang, Converting artificial neural networks to ultralow-latency spiking neural networks for action recognition, {IEEE} Trans. Cogn. Dev. Syst. 16~(4) (2024) 1533--1545.

\bibitem{Kim2019SpikingYOLOSN}
S.~J. Kim, S.~Park, B.~Na, S.~Yoon, Spiking-yolo: Spiking neural network for energy-efficient object detection, in: Proc. {AAAI} Conf. Artif. Intell., 2020, pp. 11270--11277.

\bibitem{Su2023DeepDS}
Q.~Su, Y.~Chou, Y.~Hu, J.~Li, S.~Mei, Z.~Zhang, G.~Li, Deep directly-trained spiking neural networks for object detection, in: Proc. {IEEE/CVF} Int. Conf. Comput. Vis., 2023, pp. 6532--6542.

\bibitem{Luo2024IntegerValuedTA}
X.~Luo, M.~Yao, Y.~Chou, B.~Xu, G.~Li, Integer-valued training and spike-driven inference spiking neural network for high-performance and energy-efficient object detection, arXiv:2407.20708 (2024).

\bibitem{Redmon2015YouOL}
J.~Redmon, S.~K. Divvala, R.~B. Girshick, A.~Farhadi, You only look once: Unified, real-time object detection, in: Proc. {IEEE/CVF} Int. Conf. Comput. Vis., 2016, pp. 779--788.

\bibitem{Wang2024YOLOv9LW}
C.~Wang, I.~Yeh, H.~M. Liao, Yolov9: Learning what you want to learn using programmable gradient information, arXiv:2402.13616 (2024).

\bibitem{He2015DeepRL}
K.~He, X.~Zhang, S.~Ren, J.~Sun, Deep residual learning for image recognition, in: Proc. {IEEE/CVF} Conf. Comput. Vis. Pattern Recognit., 2016, pp. 770--778.

\bibitem{Wang2019CSPNetAN}
C.~Wang, H.~M. Liao, Y.~Wu, P.~Chen, J.~Hsieh, I.~Yeh, Cspnet: {A} new backbone that can enhance learning capability of {CNN}, in: Proc. {IEEE/CVF} Int. Conf. Comput. Vis. Workshops, 2020, pp. 1571--1580.

\bibitem{Wu2018DirectTF}
Y.~Wu, L.~Deng, G.~Li, J.~Zhu, Y.~Xie, L.~Shi, Direct training for spiking neural networks: Faster, larger, better, in: Proc. {AAAI} Conf. Artif. Intell., 2019, pp. 1311--1318.

\bibitem{Zheng2020GoingDW}
H.~Zheng, Y.~Wu, L.~Deng, Y.~Hu, G.~Li, Going deeper with directly-trained larger spiking neural networks, in: Proc. {AAAI} Conf. Artif. Intell., 2021, pp. 11062--11070.

\bibitem{Kim2020RevisitingBN}
Y.~Kim, P.~Panda, Revisiting batch normalization for training low-latency deep spiking neural networks from scratch, Front. Neurosci. 15 (2020).

\bibitem{Duan2022TemporalEB}
C.~Duan, J.~Ding, S.~Chen, Z.~Yu, T.~Huang, Temporal effective batch normalization in spiking neural networks, in: Adv. Neural Inf. Process. Syst., 2022.

\bibitem{Ren2015FasterRT}
S.~Ren, K.~He, R.~B. Girshick, J.~Sun, Faster {R-CNN:} towards real-time object detection with region proposal networks, {IEEE} Trans. Pattern Anal. Mach. Intell. 39~(6) (2017) 1137--1149.

\bibitem{Rueckauer2017ConversionOC}
B.~Rueckauer, I.~Lungu, Y.~Hu, M.~Pfeiffer, S.~Liu, Conversion of continuous-valued deep networks to efficient event-driven networks for image classification, Front. Neurosci. 11 (2017).

\bibitem{Chakraborty2021AFS}
B.~Chakraborty, X.~She, S.~Mukhopadhyay, A fully spiking hybrid neural network for energy-efficient object detection, {IEEE} Trans. Image Process. 30 (2021) 9014--9029.

\bibitem{Li2022SpikeCF}
Y.~Li, X.~He, Y.~Dong, Q.~Kong, Y.~Zeng, Spike calibration: Fast and accurate conversion of spiking neural network for object detection and segmentation, arXiv:2207.02702 (2022).

\bibitem{Hu2023FastSNNFS}
Y.~Hu, Q.~Zheng, X.~Jiang, G.~Pan, Fast-snn: Fast spiking neural network by converting quantized {ANN}, {IEEE} Trans. Pattern Anal. Mach. Intell. 45~(12) (2023) 14546--14562.

\bibitem{Tavanaei2018DeepLI}
A.~Tavanaei, M.~Ghodrati, S.~R. Kheradpisheh, T.~Masquelier, A.~Maida, Deep learning in spiking neural networks, Neural Networks 111 (2019) 47--63.

\bibitem{Lee2016TrainingDS}
J.~Lee, T.~Delbr{\"{u}}ck, M.~Pfeiffer, Training deep spiking neural networks using backpropagation, Front. Neurosci. 10 (2016).

\bibitem{Neftci2019SurrogateGL}
E.~O. Neftci, H.~Mostafa, F.~Zenke, Surrogate gradient learning in spiking neural networks: Bringing the power of gradient-based optimization to spiking neural networks, {IEEE} Signal Process. Mag. 36~(6) (2019) 51--63.

\bibitem{Wu2017SpatioTemporalBF}
Y.~Wu, L.~Deng, G.~Li, J.~Zhu, L.~Shi, Spatio-temporal backpropagation for training high-performance spiking neural networks, Front. Neurosci. 12 (2017).

\bibitem{Lee2019EnablingSB}
C.~Lee, S.~S. Sarwar, K.~Roy, Enabling spike-based backpropagation in state-of-the-art deep neural network architectures, arXiv:1903.06379 (2019).

\bibitem{Hu2021AdvancingSN}
Y.~Hu, Y.~Wu, L.~Deng, G.~Li, Advancing residual learning towards powerful deep spiking neural networks, IEEE Trans. Neural Networks Learn. Syst. (2021).

\bibitem{Fang2021DeepRL}
W.~Fang, Z.~Yu, Y.~Chen, T.~Huang, T.~Masquelier, Y.~Tian, Deep residual learning in spiking neural networks, in: Adv. Neural Inf. Process. Syst., 2021, pp. 21056--21069.

\bibitem{song2020research}
S.~Song, J.~Zhu, Research on underwater biological object recognition based on mask r-cnn and transfer learning, Comput. Appl. Res 37 (2020) 386--391.

\bibitem{He2017MaskR}
K.~He, G.~Gkioxari, P.~Doll{\'{a}}r, R.~B. Girshick, Mask {R-CNN}, {IEEE} Trans. Pattern Anal. Mach. Intell. 42~(2) (2020) 386--397.

\bibitem{Zhang2021LightweightUO}
M.~Zhang, S.~Xu, W.~Song, Q.~He, Q.~Wei, Lightweight underwater object detection based on {YOLO} v4 and multi-scale attentional feature fusion, Remote. Sens. 13~(22) (2021) 4706.

\bibitem{Bochkovskiy2020YOLOv4OS}
A.~Bochkovskiy, C.~Wang, H.~M. Liao, Yolov4: Optimal speed and accuracy of object detection, arXiv:2004.10934 (2020).

\bibitem{Zhang2023AnIY}
J.~Zhang, J.~Zhang, K.~Zhou, Y.~Zhang, H.~Chen, X.~Yan, An improved yolov5-based underwater object-detection framework, Sensors 23~(7) (2023) 3693.

\bibitem{Wang2022YOLOv7TB}
C.~Wang, A.~Bochkovskiy, H.~M. Liao, Yolov7: Trainable bag-of-freebies sets new state-of-the-art for real-time object detectors, in: Proc. {IEEE/CVF} Conf. Comput. Vis. Pattern Recognit., 2023, pp. 7464--7475.

\bibitem{Varghese2024YOLOv8AN}
R.~Varghese, M.~Sambath, Yolov8: A novel object detection algorithm with enhanced performance and robustness, Proc. Int. Conf. Adv. Data Eng. Intell. Comput. Syst. (2024) 1--6.

\bibitem{Liu2023UnderwaterOD}
K.~Liu, L.~Peng, S.~Tang, Underwater object detection using {TC-YOLO} with attention mechanisms, Sensors 23~(5) (2023) 2567.

\bibitem{Vaswani2017AttentionIA}
A.~Vaswani, N.~Shazeer, N.~Parmar, J.~Uszkoreit, L.~Jones, A.~N. Gomez, L.~Kaiser, I.~Polosukhin, Attention is all you need, in: Adv. Neural Inf. Process. Syst., 2017, pp. 5998--6008.

\bibitem{Li2021EnhancingUI}
X.~Li, G.~Hou, K.~Li, Enhancing underwater image via adaptive color and contrast enhancement, and denoising, Eng. Appl. Artif. Intell. 111 (2021) 104759.

\bibitem{You2023ResearchOI}
N.~You, L.~Han, D.~Zhu, W.~Song, Research on image denoising in edge detection based on wavelet transform, Appl. Sci. (2023).

\bibitem{Tian2023ACT}
C.~Tian, M.~Zheng, W.~Zuo, S.~Zhang, Y.~Zhang, C.~Lin, A cross transformer for image denoising, Inf. Fusion 102 (2024) 102043.

\bibitem{Ju2024TowardsMS}
Y.~Ju, J.~Xiao, C.~Zhang, H.~Xie, A.~Luo, H.~Zhou, J.~Dong, A.~C. Kot, Towards marine snow removal with fusing fourier information, Inf. Fusion 117 (2025) 102810.

\bibitem{Ioffe2015BatchNA}
S.~Ioffe, C.~Szegedy, Batch normalization: Accelerating deep network training by reducing internal covariate shift, in: Proc. Int. Conf. Mach. Learn., 2015, pp. 448--456.

\bibitem{Fang2023SpikingJellyAO}
W.~Fang, Y.~Chen, J.~Ding, Z.~Yu, T.~Masquelier, D.~Chen, L.~Huang, H.~Zhou, G.~Li, Y.~Tian, Spikingjelly: An open-source machine learning infrastructure platform for spike-based intelligence, Sci. Adv. 9 (2023).

\bibitem{Horowitz201411CE}
M.~Horowitz, 1.1 computing's energy problem (and what we can do about it), in: Proc. IEEE Int. Solid-State Circuits Conf., 2014, pp. 10--14.

\bibitem{He2014SpatialPP}
K.~He, X.~Zhang, S.~Ren, J.~Sun, Spatial pyramid pooling in deep convolutional networks for visual recognition, {IEEE} Trans. Pattern Anal. Mach. Intell. 37~(9) (2015) 1904--1916.

\bibitem{urpc2019-nrbk1_dataset}
underwater fish, Urpc2019 dataset, \url{ https://universe.roboflow.com/underwater-fish-f6cri/urpc2019-nrbk1} (2023).

\bibitem{Liu2020AND}
C.~Liu, Z.~Wang, S.~Wang, T.~Tang, Y.~Tao, C.~Yang, H.~Li, X.~Liu, X.~Fan, A new dataset, poisson {GAN} and aquanet for underwater object grabbing, {IEEE} Trans. Circuits Syst. Video Technol. 32~(5) (2022) 2831--2844.

\bibitem{Everingham2014ThePV}
M.~Everingham, S.~M.~A. Eslami, L.~V. Gool, C.~K.~I. Williams, J.~M. Winn, A.~Zisserman, The pascal visual object classes challenge: {A} retrospective, Int. J. Comput. Vis. 111~(1) (2015) 98--136.

\bibitem{Duan2019CenterNetKT}
K.~Duan, S.~Bai, L.~Xie, H.~Qi, Q.~Huang, Q.~Tian, Centernet: Keypoint triplets for object detection, in: Proc. {IEEE/CVF} Int. Conf. Comput. Vis., 2019, pp. 6568--6577.

\bibitem{Tan2019EfficientDetSA}
M.~Tan, R.~Pang, Q.~V. Le, Efficientdet: Scalable and efficient object detection, in: Proc. {IEEE/CVF} Conf. Comput. Vis. Pattern Recognit., 2020, pp. 10778--10787.

\bibitem{glenn_jocher_2021_5563715}
G.~J. et. al., {ultralytics/yolov5: v6.0 - YOLOv5n 'Nano' models, Roboflow integration, TensorFlow export, OpenCV DNN support} (2021).
\newblock \href {https://doi.org/10.5281/zenodo.5563715} {\path{doi:10.5281/zenodo.5563715}}.

\bibitem{Li2023YOLOv6VA}
C.~Li, L.~Li, Y.~Geng, H.~Jiang, M.~Cheng, B.~Zhang, Z.~Ke, X.~Xu, X.~Chu, Yolov6 v3.0: A full-scale reloading, ArXiv:2301.05586 (2023).

\bibitem{Wang2024YOLOv10RE}
A.~Wang, H.~Chen, L.~Liu, K.~Chen, Z.~Lin, J.~Han, G.~Ding, Yolov10: Real-time end-to-end object detection, arXiv:2405.14458 (2024).

\bibitem{Shao2021AnIA}
C.~Shao, P.~Kaur, R.~Kumar, An improved adaptive weighted mean filtering approach for metallographic image processing*, J. Intell. Syst. 30~(1) (2021) 470--478.

\bibitem{Yu2019MemristorBL}
Y.~Yu, N.~Yang, C.~Yang, N.~Tashi, Memristor bridge-based low pass filter for image processing, J. Syst. Eng. Electron. (2019).

\end{thebibliography}

\newpage

\section*{Author Biography}

\noindent
\begin{minipage}{\linewidth}
	\begin{minipage}[t]{25mm}
	\vspace{0pt}
	\includegraphics[width=1in, height=1.25in]{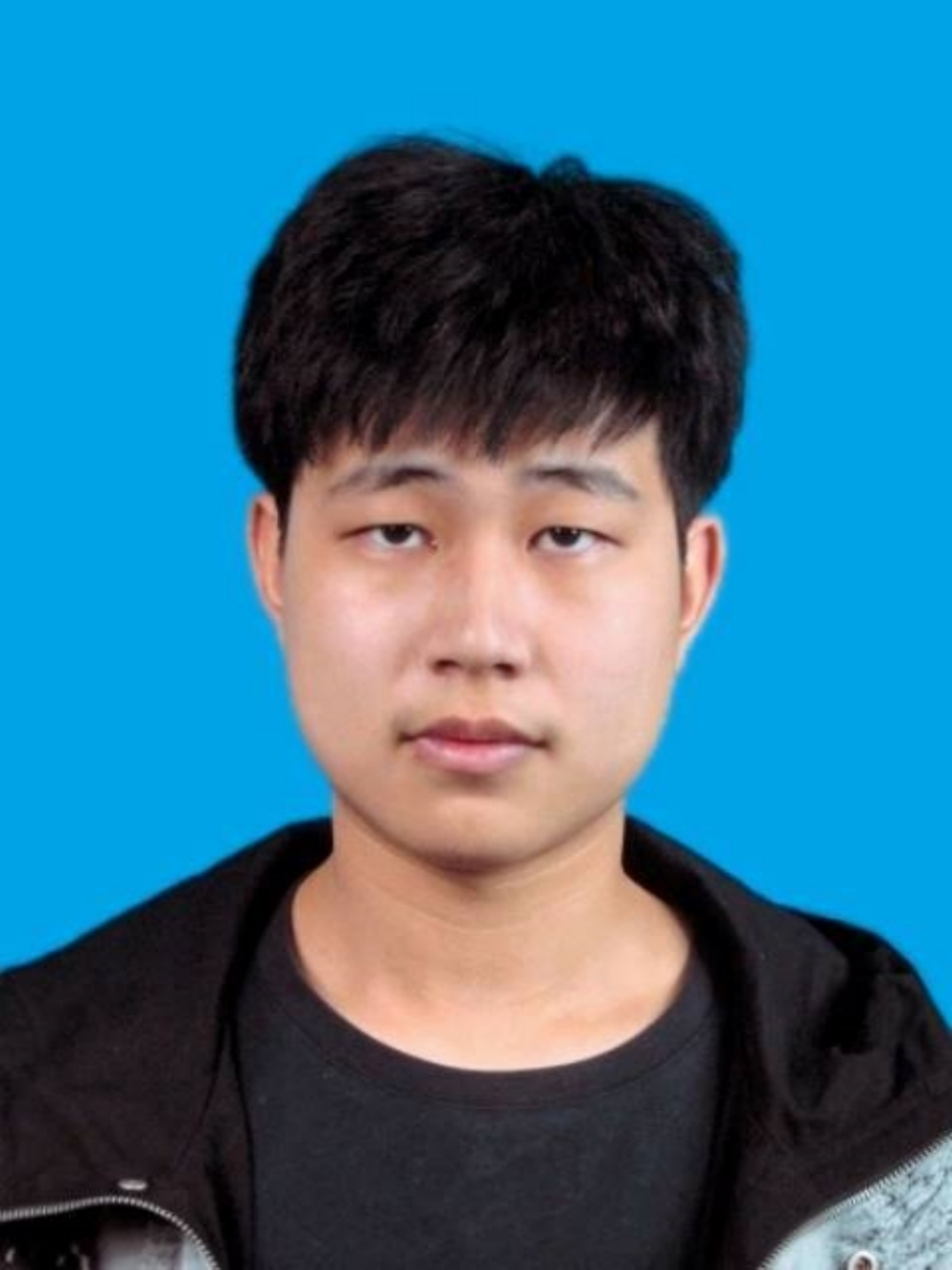}
	\end{minipage}
	\hspace{0.5em}
	\begin{minipage}[t]{\dimexpr \linewidth - 25mm - 0.5em}
	\vspace{0pt}
	\textbf{Chenyang~Li} is a M.S. student in computer technology at China Three Gorges University. He received his B.S. degree in water conservancy and hydropower engineering from China Three Gorges University in 2021. His research interests include computer vision and neuromorphic computing.
	\end{minipage}
\end{minipage}

\vspace{1em}

\noindent
\begin{minipage}{\linewidth}
	\begin{minipage}[t]{25mm}
	\vspace{0pt}
	\includegraphics[width=1in, height=1.25in]{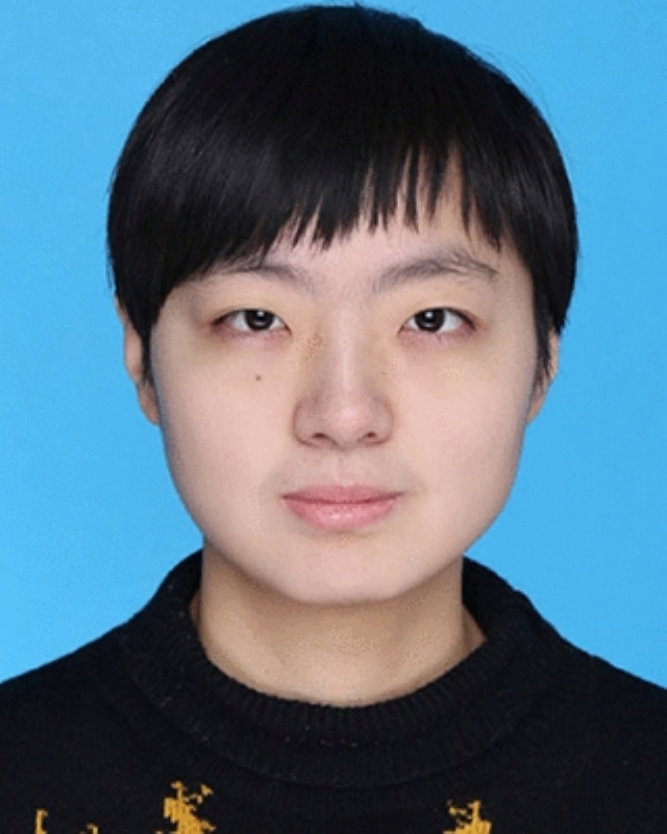}
	\end{minipage}
	\hspace{0.5em}
	\begin{minipage}[t]{\dimexpr \linewidth - 25mm - 0.5em}
	\vspace{0pt}
	\textbf{Wenxuan~Liu} received her Ph.D. from Wuhan University of Technology in 2024. Now she is a postdoc at Peking University. Her research interests include neuromorphic computing, spike vision, action recognition, and multimedia content analysis.
	\end{minipage}
\end{minipage}

\vspace{1em}

\noindent
\begin{minipage}{\linewidth}
	\begin{minipage}[t]{25mm}
	\vspace{0pt}
	\includegraphics[width=1in, height=1.25in]{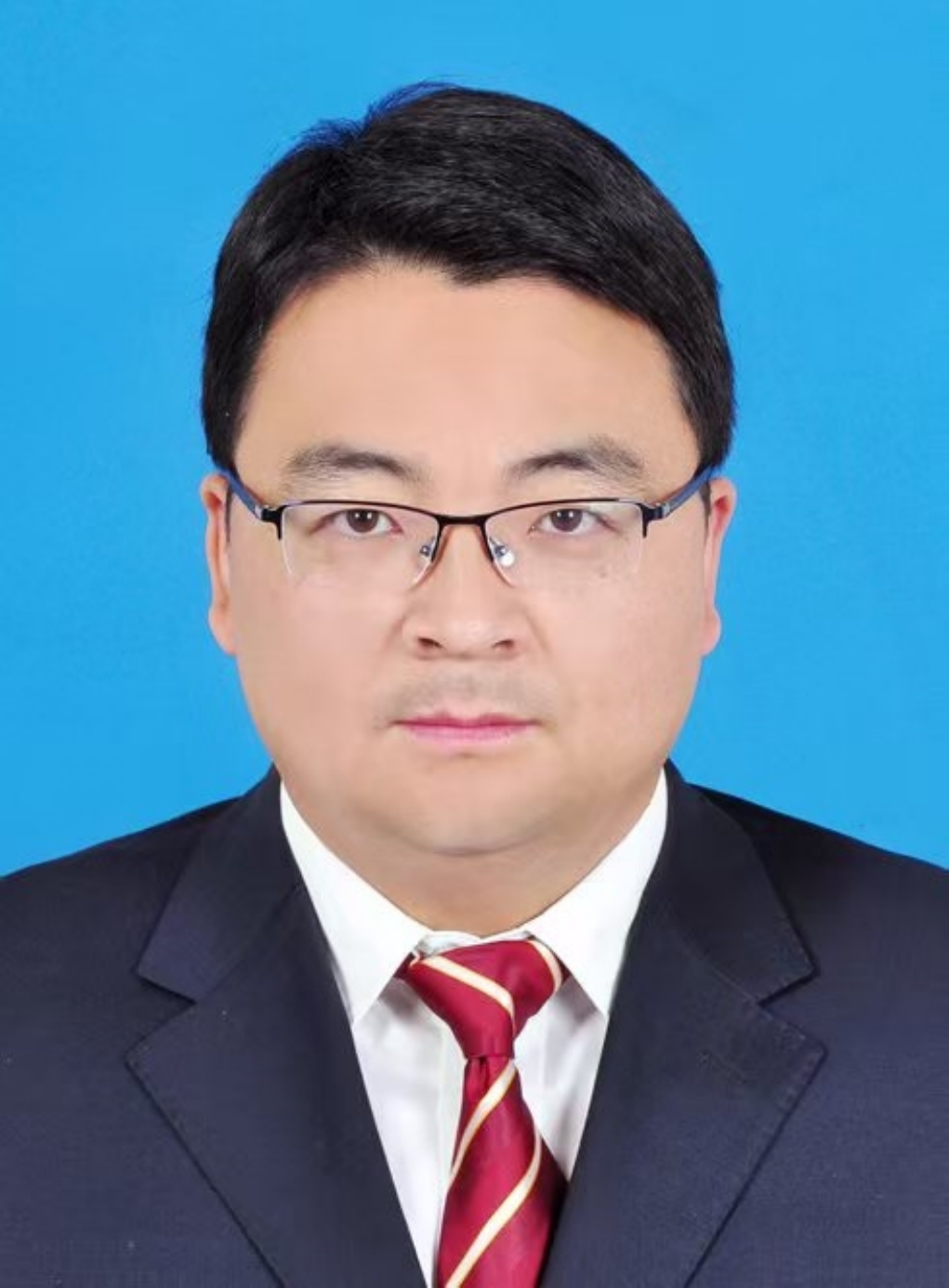}
	\end{minipage}
	\hspace{0.5em}
	\begin{minipage}[t]{\dimexpr \linewidth - 25mm - 0.5em}
	\vspace{0pt}
	\textbf{Guoqiang~Gong} received his B.S. from Lanzhou University in 1998 and Ph.D. from Tongji University in 2010. He is currently a Professor at China Three Gorges University. His research interests include intelligent signal processing and computer vision.
	\end{minipage}
\end{minipage}

\vspace{1em}

\noindent
\begin{minipage}{\linewidth}
	\begin{minipage}[t]{25mm}
	\vspace{0pt}
	\includegraphics[width=1in, height=1.25in]{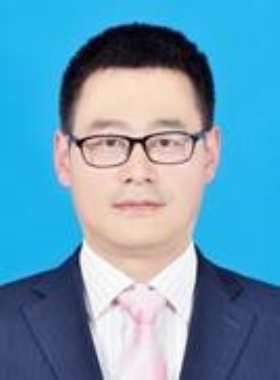}
	\end{minipage}
	\hspace{0.5em}
	\begin{minipage}[t]{\dimexpr \linewidth - 25mm - 0.5em}
	\vspace{0pt}
	\textbf{Xiaobo~Ding} received his B.S. from Beijing Electronic Science and Technology Institute in 1995 and Ph.D. from Harbin Engineering University in 2015. He is currently a Associate Professor at China Three Gorges University. His research interests include IoT technology and system virtualization.
	\end{minipage}
\end{minipage}

\vspace{1em}

\noindent
\begin{minipage}{\linewidth}
	\begin{minipage}[t]{25mm}
	\vspace{0pt}
	\includegraphics[width=1in, height=1.25in]{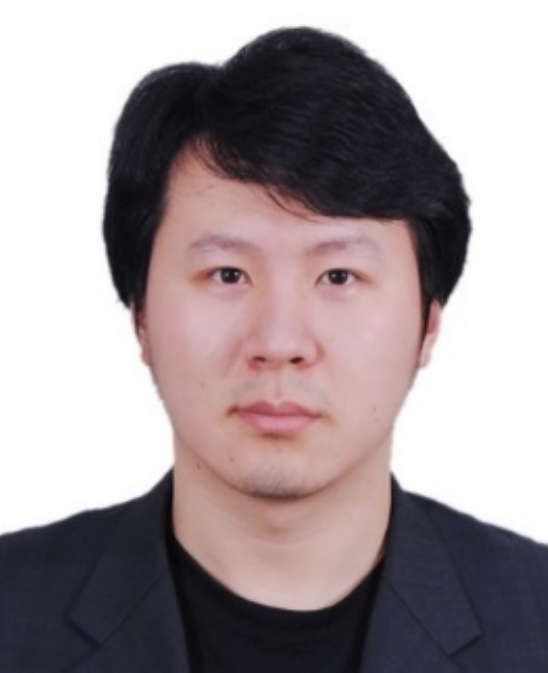}
	\end{minipage}
	\hspace{0.5em}
	\begin{minipage}[t]{\dimexpr \linewidth - 25mm - 0.5em}
	\vspace{0pt}
	\textbf{Xian~Zhong} received his B.S. degree in Computer Science from Wuhan University, China, in 2007, and his Ph.D. degree in Computer Science from Huazhong University of Science and Technology, China, in 2013. He is currently a Professor in the School of Computer Science and Artificial Intelligence at Wuhan University of Technology, China. His research interests include multimodal information processing, intelligent transportation systems, and neuromorphic computing. 
	\end{minipage}
\end{minipage}

\newpage

\section*{Figure Captions}

\cref{Fig:1}: \textbf{Comparison of detection performance and energy consumption across various models on \textsc{URPC2019}.} The blue dots represent ANN models, while the green and red dots indicate SNN models.

\cref{Fig:2}: \textbf{Diagram of SU-YOLO and the structure of its modules.} In the main structural diagram, narrower graphics represent feature maps with reduced dimensions.

\cref{Fig:3}: \textbf{Schematic diagram illustrating the processing flow and effects of the SpikeDenoiser module.}

\cref{Fig:4}: \textbf{Feature maps extracted from backbone layers when using ResNet or CSPNet in SNN.}

\cref{Fig:5}: \textbf{Impact of pooling and activation sequence on output.}

\cref{Fig:6}: \textbf{Calculation and fusion methods of SeBN.} In the SeBN computation diagram, cubes represent the feature maps at each time step, where $B$ denotes the batch size, $C$ indicates the number of channels, $H$ and $W$ are the height and width of the feature maps, and $T$ represents the total number of time steps. The symbols $\mu_k$ and $\sigma_k^2$ denote the mean and variance of the $k$-th feature map.

\cref{Fig:7}: \textbf{Process of batch-scale fusion of SeBN.} Here, $\theta_t$ represents all parameters of the SeBN layer at time step $t$, $\phi$ represents parameters shared by the convolutional layer across all time steps, and $\phi_t$ denotes parameters of the convolutional layer at time step $t$.

\cref{Fig:8}: \textbf{Running process of SU-YOLO varies with time steps.}

\cref{Fig:9}: \textbf{Denoising effect of the SpikeDenoiser.}

\cref{Fig:10}: \textbf{Data distribution after a element summation residual block with different BN methods on \textsc{URPC2019}.} $\mu$ and $\sigma^2$ are the mean and variance of data.

\cref{Fig:11}: \textbf{Detection results using various residual blocks on \textsc{URPC2019}.} Ground Truth shows the original dataset labels.

\cref{Fig:12}: \textbf{Confusion matrices for models employing different residual blocks on \textsc{URPC2019}.} The horizontal axis represents actual objects, and the vertical axis represents predicted objects of the model.

\cref{Fig:13}: \textbf{Precision-Recall curves for models employing different residual blocks on \textsc{URPC2019}.} The values in the legends represent mAP$_{0.5}$.

\cref{Fig:14}: \textbf{Pooled spike feature maps at time step $T = 1$, where black indicates a value of 0 and white indicates a value of 1.}

\newpage

\section*{Graphical Abstract}

\begin{figure}[h]
	\centering
	\includegraphics[width = \linewidth]{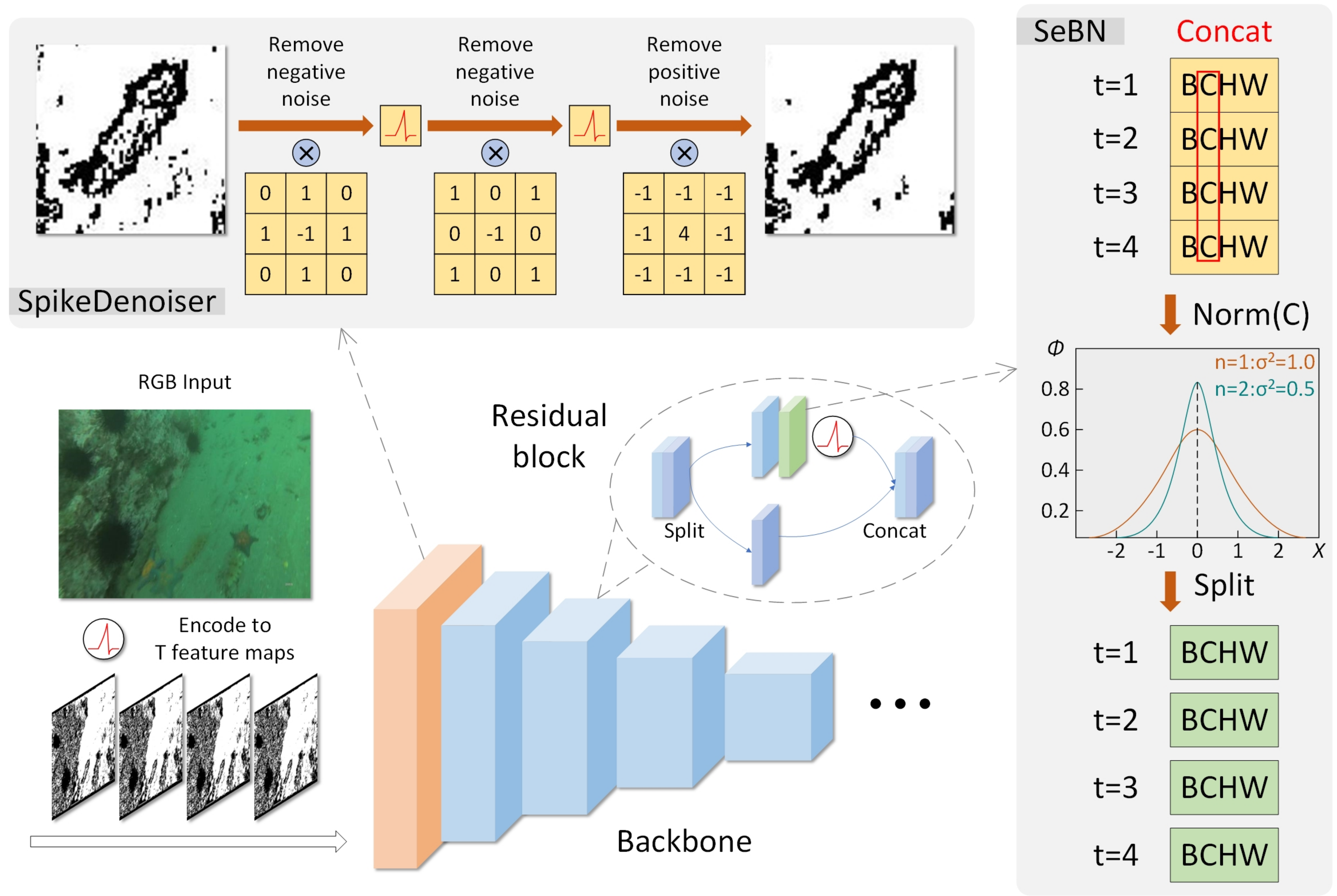}
\end{figure}





\end{document}